\documentclass[10pt,twocolumn,letterpaper]{article}
\usepackage[pagenumbers]{cvpr}
\usepackage{balance}
\usepackage{placeins}
\makeatletter
\patchcmd{\@BAlancecol}{\unvbox1\vfil}
  {\dimen2=\dp1\unvbox1\kern-\dimen2\vfil}{}
  {\PackageError{GGF}{Column balancing patch failed}{Check balance.sty.}}
\makeatother
\usepackage{array}
\usepackage{microtype}
\definecolor{cvprblue}{rgb}{0.21,0.49,0.74}
\usepackage[pagebackref,breaklinks,colorlinks,allcolors=cvprblue]{hyperref}
\pdftrailerid{}
\hypersetup{pdfauthor={Waldemar Chang and Andrii Krutsylo},pdftitle={Guided Group Featurizers (GGF): Improving Block-Sparse Featurizers with Object-Guided Learning},pdfsubject={Research preprint},pdfcreator={},pdfproducer={}}
\AtEndDocument{\pdfinfo{/Creator () /Producer ()}}

\newcommand{\ind}{\mathbf{1}}
\newcommand{\Lrec}{\mathcal{L}_{\mathrm{rec}}}
\newcommand{\Lcap}{\mathcal{L}_{\mathrm{contrast}}}
\newcommand{\Ldet}{\mathcal{L}_{\mathrm{det}}}

\newcommand{\paperimage}[2][\linewidth]{\IfFileExists{figures/#2}{\includegraphics[width=#1,height=3.4in,keepaspectratio]{figures/#2}}{\fbox{\parbox{.9\linewidth}{Figure companion file: \texttt{\detokenize{#2}}. Compile the complete project ZIP.}}}}

\newcounter{agfalgorithm}
\newcounter{agfline}
\newlength{\agfindent}
\newcommand{\agfstep}[2][0]{%
  \par\noindent\stepcounter{agfline}%
  \begingroup
  \hangindent=\dimexpr2em+#1\agfindent\relax\hangafter=1
  \makebox[1.4em][r]{\scriptsize\theagfline:}\hspace{.6em}%
  \hspace*{#1\agfindent}#2\par
  \endgroup}

\newenvironment{agfalgorithm}[1][GGF pipeline]{\begin{figure}[htbp]
\refstepcounter{agfalgorithm}
\hrule\smallskip{\small\textbf{Algorithm \theagfalgorithm: #1}}\par\smallskip\hrule\smallskip\small
\begingroup\raggedright\setlength{\parindent}{0pt}%
\setlength{\parskip}{0pt}\setcounter{agfline}{0}}
{\endgroup\smallskip\hrule\end{figure}}
\newenvironment{agfresulttable}{\begin{table}[htbp]\centering\small\setlength{\tabcolsep}{3pt}}{\end{table}}

\makeatletter
\patchcmd{\@maketitle}{\vspace*{12pt}}{\vspace*{-6pt}}{}{\PackageError{GGF}{Preprint title spacing patch failed}{Check the title block.}}
\makeatother
\title{Guided Group Featurizers (GGF): Improving Block-Sparse Featurizers with Object-Guided Learning}
\author{Waldemar Chang \qquad Andrii Krutsylo\\
\raisebox{-2pt}{\includegraphics[height=12pt]{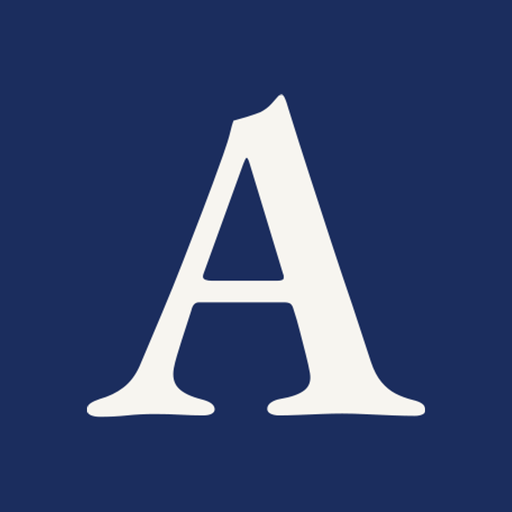}}\hspace{4pt}Algoverse AI Research\\
{\small\url{https://github.com/MechInterpreter/Guided-Group-Featurizers}}}
\begin{document}
\maketitle
\begin{abstract}
Block-sparse featurizers (BSFs) use groups of learned directions to
capture the multidimensional structure of visual concepts. However,
grouping alone does not ensure that information about a concept is
consolidated within a single group.
We introduce Guided Group Featurizers (GGF), a novel object-guided extension
that uses object annotations for initialization and training while
preserving the BSF architecture. We assess whether each group's strongest response regions share an
object type, part or material across images. Blinded GPT reviews give GGF
groups an average semantic coherence score of $93.4\pm1.6$\%, versus
$83.4\pm2.5$\% for SLiCS and $67.9\pm3.1$\% for supervised Vanilla BSF.
GGF also leads in a separate blinded Gemini review.
GGF's consolidation advantage extends to unguided categories: one selected
group per category covers $66.1\pm1.9$\% of target-containing crops, versus
$59.2\pm1.8$\% for supervised Vanilla BSF, at approximately 5\%
false-positive rates for both. Across COCO and VOC, GGF also improves single-group object Capture and
localization over Vanilla BSFs, SLiCS and Concept-SAE. Its groups produce more
focused and reproducible object responses while supporting recognition
and retaining associations under rotation, scale and brightness changes. GGF also
uses fewer learned parameters than Concept-SAE and shorter recorded
VOC training loops than both supervised alternatives. These findings
support object guidance for organizing more interpretable BSF groups and
show that response concentration extends to categories excluded from guidance.
\end{abstract}

\section{Introduction}\label{sec:introduction}
Sparse autoencoders (SAEs) reconstruct neural activations as sparse
combinations of inspectable learned directions \citep{gao2024scaling}.
Accurate reconstruction, however, does not explain how those components
organize meaningful concepts.

\begin{figure}[htbp]\centering
\includegraphics[width=.90\linewidth]{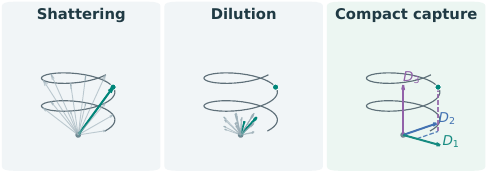}
\caption{\textbf{Shattering, dilution, and compact capture.}
The helix represents a concept manifold; arrows denote dictionary directions.
Shattering uses local directions, dilution uses changing, overlapping subsets,
and compact capture reuses a small shared set across the manifold.
Schematic adapted from \citet[Fig.~3]{bhalla2026manifolds}.}
\label{fig:tiling-schematic}
\end{figure}

This distinction matters when a concept has multidimensional structure.
Its activations may trace a curve or surface, forming a \emph{concept
manifold}. SAEs can \emph{tile} these manifolds
\citep{bhalla2026manifolds}: different features cover local regions
(shattering), or changing, overlapping subsets provide redundant
representations (dilution). Compact capture instead uses a small shared
set of decoder directions, with coefficients that vary across manifold
points. Reconstruction can succeed in each case, while the organization
across examples differs (Figure~\ref{fig:tiling-schematic}).

\begin{figure*}[t]\centering
\includegraphics[width=\textwidth]{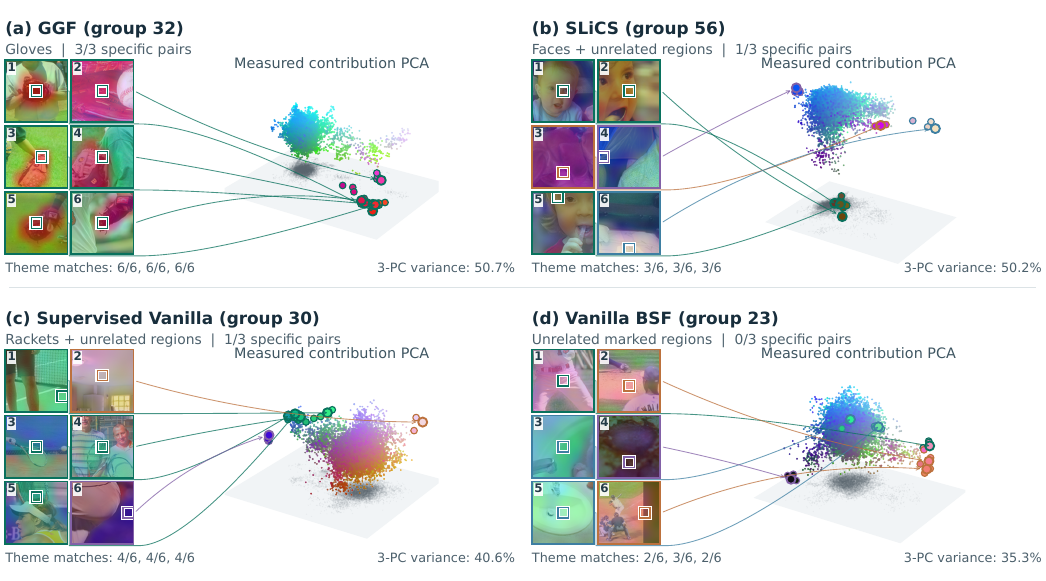}
\caption{\textbf{Coherent and mixed responses in selected native groups.}
Numbered crops show the strongest marked patches from six images for
one native group per 75-group, seed-21 model.
\emph{Specific pairs} counts majority-confirmed pairs sharing an object
type, part, material or other specific meaning, out of three fixed pairs.
\emph{Theme matches} gives the view count (out of six) fitting each of
three blinded AI reviewers' themes.
Clouds project decoded patch contributions using principal component
analysis (PCA); \emph{3-PC variance} is reference variance retained by
the first three components. RGB encodes PCA coordinates; borders,
arrows and outlines mark visible content. Arrows locate shown patches;
outlined points include all four marks per image. PCA bases are fitted
separately (Appendix~\ref{app:coherence-pca}). These selected illustrations
complement aggregate comparisons of every group
(Table~\ref{tab:matched75-coherence}).}
\label{fig:response-geometry}\end{figure*}

Block-sparse featurizers (BSFs) make this richer structure an architectural
possibility by replacing individual directions with groups of directions
\citep{fel2026}. Sparsity selects groups together, while input-dependent coefficients
describe positions within each active group's subspace. This represents
multidimensional structure as one unit, improving manifold recovery
and activation description length.

Providing that capacity does not ensure that training consolidates a
concept within it. In their analysis of Grassmannian BSFs,
\citet{jerpelea2026} find that synthetic factors can still split across
blocks, including pairs that capture complementary parts of the same
factor. This gap between representational capacity and learned organization
motivates asking whether explicit guidance can help groups capture
more of a concept's information.

Object annotations provide a practical way to investigate this question.
They connect annotated regions to groups during training. Objects serve
as concept proxies: a model's concepts can also
describe properties shared across object categories. We ask: \emph{can
object guidance help groups consolidate concepts, or annotated objects
used as concept proxies, more effectively?} Here, consolidation means
making substantial object-related information accessible within an
individual group. We separately test reconstructed regional information
and the concentration of category responses within small group subsets.

Figure~\ref{fig:response-geometry} illustrates the semantic coherence of a
group's responses across images.
The selected GGF group responds to baseball gloves across six images;
all three fixed pairs confirm that specific meaning. The selected SLiCS,
supervised Vanilla and Vanilla groups mix unrelated marked regions,
confirming only one, one and zero pairs, respectively. The exhaustive 75-group comparison gives GGF the highest aggregate
specific-pair coherence among the four matched methods (Section~\ref{sec:matched75-coherence},
Table~\ref{tab:matched75-coherence}).

\paragraph{Contributions.}
Our contributions are as follows:
\begin{itemize}
\item \textbf{Object guidance for more interpretable BSF groups.}\nopagebreak[4]
We introduce GGF, a novel object-guided extension that makes object-related
responses more semantically coherent and information more accessible within
individual BSF groups. Object-informed \emph{Initialization} and adaptive
\emph{Guided Learning} preserve the native architecture and additive
activation reconstruction.
\item \textbf{Evidence of coherent and concentrated object responses.}\nopagebreak[4]
We provide exhaustive coherence and concentration comparisons of matched
75-group models and show concentration benefits for categories excluded
from guidance. COCO and VOC comparisons, recognition tests, response
consistency across images and appearance changes, and visualizations
establish the scope of these interpretability gains.
\item \textbf{Explaining and validating consolidation.}\nopagebreak[4]
We explain which training components improve single-group reconstruction
of object-versus-background feature contrast, measured by Capture, and
which improve localization. Tests of group subsets and contribution
removal assess information accessibility and effects on predictions.
Geometric diagnostics show how groups use multiple directions, while
recovery of known synthetic factors directly tests consolidation.
\item \textbf{Practical capacity and resource tradeoffs.}\nopagebreak[4]
We quantify tradeoffs among object alignment, reconstruction, group
capacity, activity, training duration and resources for training and
inference. These comparisons make GGF's practical costs explicit,
including fewer learned parameters than Concept-SAE and shorter recorded
VOC training loops than SLiCS and Concept-SAE on the tested hardware.
\end{itemize}

\section{Related Work}\label{sec:related}
\paragraph{From sparse directions to concept geometry.}
Manifold analyses examine whether sparse learned directions
\citep{gao2024scaling} capture concepts compactly or tile local regions
\citep{bhalla2026manifolds}. BSFs address this mismatch through structured
group sparsity, connecting concept geometry to reconstruction and
description length \citep{fel2026,goodfire2026}. Yet related information
can remain distributed across blocks, as \citet{jerpelea2026} show for
Grassmannian BSFs. This motivates studying how training organizes the
capacity that groups provide.

\paragraph{Supervision for concept representations.}
Annotations offer another way to influence that organization. Task-driven
dictionary learning establishes supervision as part of sparse
representation learning \citep{mairal2012}, while SpLiCE connects
representations to word-derived concept directions \citep{bhalla2024}.
For multidimensional concepts, SLiCS learns concept-specific groups of
atoms with nonnegative, input-dependent coefficients \citep{li2025}.
Concept-SAE instead learns representations supervised for concept
existence and location, combining concept tokens with free tokens to
reconstruct activations \citep{ding2026conceptsae}. These methods connect supervision to the content of a concept unit.
Other approaches supervise class-aligned scalar features \citep{stepien2026classae}, centralize concepts for diffusion
unlearning \citep{cassano2026saemnesia}, and anchor directions or
subspaces \citep{fraser2026anchoring}. GGF instead guides consolidation
within the native BSF architecture.

\paragraph{From spatial alignment to group organization.}
Group evaluation connects spatial responses to represented information.
Network Dissection measures alignment with annotated regions \citep{bau2017}, and LUCID-SAE evaluates
the spatial grounding of sparse concepts \citep{gu2026}. Foreground/background
prototype pooling \citep{wang2019panet} and support decontamination
\citep{xing2026} further motivate comparing object information with
surrounding evidence. We connect spatial alignment to the object-related activation contrast
reconstructed by a group. Measuring recovery through group subsets and
removing selected contributions from fixed readouts then reveals how
concentrated that information is and how the readouts use it. Profiles of
object responses and individual direction contributions show whether these
associations repeat across images and agree within groups.

\section{Methodology}\label{sec:method}
\begin{figure*}[t]\centering
\includegraphics[width=\textwidth]{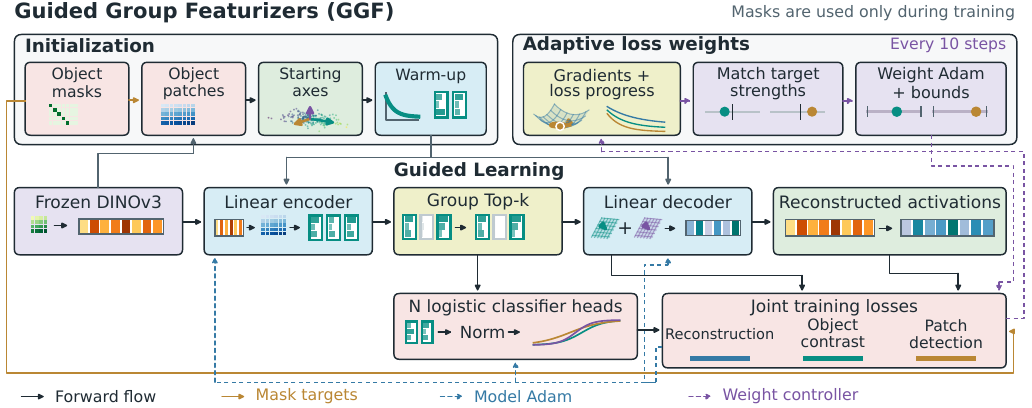}
\caption{GGF training workflow. Upper left: masks initialize axes before
reconstruction warm-up. Guided Learning encodes, selects and decodes groups
from left to right. For $N$ supervised categories, $N$ logistic heads map
group norms to detection loss, alongside reconstruction and object contrast.
Model Adam updates the representation and heads; the upper-right controller
adjusts loss weights.}
\label{fig:pipeline}\end{figure*}

GGF uses Vanilla BSF's encoder, group gate, decoder and fixed groups
\citep{fel2026,goodfire2026}. We choose it over Grassmannian and Group
Lasso for its initial COCO Capture lead, supported by synthetic recovery
and reconstruction (Appendix~\ref{app:screen256}). Figure~\ref{fig:pipeline} follows two stages: Initialization establishes a
reconstruction model, then Guided Learning adds object guidance.
The supplementary code and evidence bundle accompanies this manuscript.

\subsection{Features and object evidence}\label{sec:masks}
We use DINOv3 ViT-B/16 \citep{simeoni2025} as our frozen backbone,
following \citet{fel2026}, who trained BSFs on DINOv3's final-layer
patch activations. Each image supplies 196 vectors of dimension $d=768$. We resize the shorter side to 256 and
center-crop to $224\times224$, applying the same geometry to masks.
Training-derived positional means remove average spatial patterns, and
a shared scale normalizes magnitude (Appendix~\ref{app:featuredetails}).
All methods share these cached inputs. Because attention combines
information across the image, regional features can also reflect context
beyond their own pixels.

Write $x_{ip}\in\mathbb R^d$ for the normalized vector at patch $p$ of
image $i$. Let $\mathcal C$ be the supervised category set; $\ind[\cdot]$
is an indicator and $\|\cdot\|_2$ the Euclidean norm.
For object instance $o$ of category $c$ in image $i$, let $f_{op}$ be its
coverage of patch $p$, $q_{icp}$ the category-union coverage, and $u_{ip}$
the crowd-mask coverage. Write $v_{ip}=\ind[u_{ip}\leq0.1]$ for usable
patches and $b_{icp}=v_{ip}\ind[q_{icp}\leq0.1]$ for comparison patches.
Foreground/background prototypes motivate regional pooling
\citep{wang2019panet}. Here we define an activation contrast by subtracting
the comparison mean from the coverage-weighted object mean:
\begin{equation}
\begin{split}
w_{op}&=\frac{f_{op}v_{ip}}{\sum_{p'}f_{op'}v_{ip'}}
-\frac{b_{icp}}{\sum_{p'}b_{icp'}},\\
\delta x_o&=\sum_p w_{op}x_{ip}.
\end{split}\label{eq:contrast}
\end{equation}
A contrast requires at least two usable foreground patch-equivalents
and four comparison patches. Since comparison regions can contain other
objects, subtraction reduces shared components without isolating a pure
concept or removing all context. Images without eligible instances still
contribute to reconstruction.

\subsection{Initialization}\label{sec:groups}
\paragraph{Object patches and starting axes.}
Initialization places strong foreground directions together so guidance
starts from object-associated groups (Figure~\ref{fig:pipeline}). A dictionary contains $G$ groups of
$r$ directions, with decoder $D_g\in\mathbb R^{d\times r}$, encoder
$E_g\in\mathbb R^{r\times d}$, and bias $b_g\in\mathbb R^r$ for group $g$.
Initialization uses training patches with at least 50\% category coverage
and no more than 10\% crowd coverage. For each retained category, the
leading $r$ eigenvectors of the uncentered foreground second moment
initialize one decoder group, retaining high-energy foreground structure.
The encoder starts as the decoder transpose with zero bias, while random
reserve groups can model other activation structure. Keeping a shared
origin lets the axes reflect both category means and variation. Training
can then change a group's orientation or the categories it represents
(Appendix~\ref{app:initdetails}).

\paragraph{Warm-up.}
Reconstruction-only warm-up adapts the starting directions to the full
activation distribution before applying object losses. The resulting
encoder and decoder initialize \emph{Guided Learning}, where selected
groups receive object-specific guidance.

\subsection{Guided Learning}\label{sec:loss}
\paragraph{Forward pass.}
For a normalized patch vector $x$, the encoder produces coefficients
$a_g(x)$. Group Top-$k$ retains $k$ groups with largest coefficient norms;
$K(x)$ is their index set, $z_g(x)$ their gated codes, and $r_g(x)$ their
decoded contributions. The reconstruction is $\widehat x$
(Figure~\ref{fig:pipeline}):
\begin{equation}
\begin{split}
a_g(x)&=E_gx+b_g,\quad
K(x)=\operatorname{TopK}_{k}\{\|a_h(x)\|_2\}_{h=1}^G,\\
z_g(x)&=a_g(x)\ind[g\in K(x)],\\
r_g(x)&=D_gz_g(x),\qquad\widehat x=\sum_g r_g(x),
\end{split}\label{eq:architecture}
\end{equation}
Top-$k$ retains at most $kr$ scalar coefficients per patch.

\paragraph{Fixed readouts and logistic classifiers.}
After warm-up, training data select each category's contrast readout
$g_c$ and detector readout $g_c^{\rm det}$ independently because the
tasks can favor different groups (Section~\ref{sec:metrics}). Readouts
may share groups or differ from initialized slots. Fixing these
assignments gives guidance stable targets without evaluation-set selection.

For $N=|\mathcal C|$ supervised categories, each head maps its group norm
$h_{icp}=\|z_{g_c^{\rm det}}(x_{ip})\|_2$ to logit $\ell_{icp}$ and
probability $\pi_{icp}$. It learns slope parameter $\alpha_c$ and intercept
$\beta_c$, with fixed training-derived center $m_c$ and scale $t_c\geq10^{-3}$:
\begin{equation}
\begin{split}
\ell_{icp}&=\operatorname{softplus}(\alpha_c)
\frac{h_{icp}-m_c}{t_c}+\beta_c,\\
\pi_{icp}&=\operatorname{sigmoid}(\ell_{icp}),
\end{split}\label{eq:logit}\end{equation}
The positive slope calibrates activity while preserving norm ordering.
The heads add $2N$ parameters (150 for COCO, 38 for VOC). Evaluation
uses the raw norms, so the heads affect training but do not enter
reconstruction or Top-$k$ selection.

\paragraph{Reconstruction, contrast, and detection.}
With $B$ a clean patch batch and $\epsilon$ standard normal noise, denoising
\citep{vincent2010} trains tolerance to feature perturbations using
$\widetilde x=x+0.1\operatorname{std}(B)\epsilon$. At update $u$,
weights $\lambda_C(u)$ and $\lambda_D(u)$ scale contrast loss $\Lcap$ and
detection loss $\Ldet$; $\Lrec$ is reconstruction loss and $\mathcal L$
the total. The expectation averages training patches and noise:
\begin{equation}
\begin{split}
\mathcal L&=\Lrec+\lambda_C(u)\Lcap+\lambda_D(u)\Ldet,\\
\Lrec&=\mathbb E\|x-\widehat x(\widetilde x)\|_2^2/d.
\end{split}\label{eq:total}\end{equation}
This noise is used only for reconstruction; object contrast and patch
detection are computed from clean inputs.

\paragraph{Preserving regional contrast.}
Using clean inputs, the chosen group's regional contribution is
$\delta r_{og}=\sum_p w_{op}r_g(x_{ip})$.
For batch categories $\mathcal C_B$ with eligible instances
$\mathcal O_{Bc}$, the object loss is
\begin{equation}
\Lcap=\frac{1}{|\mathcal C_B|}\sum_{c\in\mathcal C_B}
\frac{1}{|\mathcal O_{Bc}|}\sum_{o\in\mathcal O_{Bc}}
\frac{\|\delta x_o-\delta r_{o g_c}\|_2^2}
{\max(\|\delta x_o\|_2^2,10^{-6})}.
\label{eq:caploss}\end{equation}
Normalizing each target and averaging categories equally prevents larger
objects or frequent categories from dominating this objective.

\paragraph{Localizing category evidence.}
A group can reconstruct a regional contrast without responding
selectively at the object location. Detection therefore uses the
classifier probabilities to reward activity on object patches. Usable
patches with coverage at least 0.5 are positive and those with coverage
at most 0.1 are negative. Intermediate patches are ignored.
Let $\mathcal C_B^{\rm det}$ contain batch categories with usable detection
patches. Write $\mathcal A_{c,t}$ for their patch pool with sign $t$
($+1$ foreground, $-1$ background), and $T_c$ for its nonempty signs:
\begin{equation}
\begin{split}
\Ldet={}&\frac1{|\mathcal C_B^{\rm det}|}
\sum_{c\in\mathcal C_B^{\rm det}}\frac1{|T_c|}
\sum_{t\in T_c}\frac1{|\mathcal A_{c,t}|}\\
&\cdot\sum_{(i,p)\in\mathcal A_{c,t}}
\operatorname{softplus}(-t\ell_{icp}).
\end{split}\label{eq:detloss}\end{equation}
Averaging the positive and negative pools separately prevents the more
numerous background patches from dominating. A task contributes zero
when its eligible pool is empty.

\paragraph{Model Adam and adaptive loss weights.}
The three losses differ in scale and learning speed, so the model and
its loss weights are updated separately. Adam \citep{kingma2015adam} updates the encoder, decoder and heads,
followed by unit normalization of decoder directions
(Figure~\ref{fig:pipeline}). A separate
controller compares encoder-gradient norms and relative loss progress.
Every ten steps, its own Adam optimizer adjusts the two bounded object
weights while keeping the reconstruction weight at one. Gradient norms
and targets are detached, so these updates require no second-order
differentiation and take effect at the next model step. This GradNorm-inspired refinement
\citep{chen2018gradnorm} and its exact bounds appear in
Appendix~\ref{app:controller}. Constant-weight controls distinguish object
guidance from the controller's smaller contribution.
Algorithm~\ref{alg:agf-main} summarizes the full workflow.

\section{Experimental design}\label{sec:data}
\paragraph{Data and evaluation.}
We use COCO train2017 \citep{lin2014}, retaining 75 categories.
The shared allocation has 2,000 training images and three 400-image
cohorts for selection, confirmation and readout calibration. The 75-group
study calibrates on confirmation images and scores 1,500 reserved audit
images (Appendix~\ref{app:allocation}). For cross-dataset evaluation,
we fit and develop models on VOC2012 training images and test on official
validation images \citep{everingham2010voc,voc2012}. VOC normalization,
initialization and readouts use its own training data
(Appendix~\ref{app:voc-results}).

\paragraph{Group count and category splits.}
We use 75 groups per method for interpretability: SLiCS requires one
dictionary per category, and matching that count makes exhaustive review
practical for 75 categories. The methods can still differ in parameters,
activity and learned meanings. Other counts answer complementary questions:
256 groups support the BSF screen, while 1,024 and 4,096 groups support
reconstruction, capacity and training-component comparisons. These BSF
control arms match group count within each comparison
(Appendices~\ref{app:screen256} and \ref{app:pilotlatest}).

To test transfer beyond guidance, two complementary folds guide training
on 37 or 38 categories while retaining 75 groups. Excluded annotations
enter neither initialization, guided training nor model selection, although
their objects may occur in training images. Their readouts are selected
after fitting on separate calibration data. Shared Vanilla counts once
per run. Native SLiCS with 37 or 38 dictionaries is capacity-unmatched;
Concept-SAE with 111 tokens remains exploratory
(Appendix~\ref{app:matched75}).

\paragraph{Comparisons and controls.}
Supervised Vanilla shares GGF's object losses and classifier heads,
testing what initialization and adaptive weighting add
(Appendix~\ref{app:current-supervised-control}). Mask-initialized and
fixed-guidance arms separate these components
(Appendix~\ref{app:components-current}). We adapt SLiCS to patch-level training and retain Concept-SAE's native
training stages \citep{li2025,ding2026conceptsae}. Both use COCO-selected
learning rates on VOC (Appendices~\ref{app:supervised-five}, \ref{app:concept-sae-protocol}).
Further studies vary capacity, activity, loss weights and continuation
(Appendices~\ref{app:supervision-overview}, \ref{app:capacity-overview}),
while Manifold Zoo tests recovery of known generating factors
(Appendix~\ref{app:manifoldzoo}).

\paragraph{Primary interpretability measures.}
We assess coherence \citep{ghorbani2019ace,fel2023craft,bagci2026elude}
using six strongest-response views for every group and three fixed pairs.
Reviewers judge specific shared meaning (e.g., object type, part or material)
and visual structure separately for each pair. A whole-group theme must
fit at least five views, with specificity rated separately. We score majority
votes across three fresh blinded GPT-6.1 Sol contexts at high reasoning
effort \citep{openai2026gpt61sol}. All planned pairs and groups count
toward their respective totals, including those with unclear judgments.
Separately, Gemini 3.5 Flash-Lite \citep{google2026gemini35flashlite}
applies the same criteria to the same blinded cards once per task and
group (Appendix~\ref{app:cross-model}).

Response concentration asks how broadly a small subset detects an object
category at a measured false-positive rate. Calibration ranks all 75 groups by balanced accuracy for each category
and sets thresholds at the 95th percentile of
target-negative responses. We freeze the top 1, 2, 4 or 8 groups before
audit scoring. A crop is detected if any selected group exceeds its
threshold. Coverage is the fraction of known target-present crops
detected; the false-positive rate is the fraction of known target-absent
crops detected \citep{fawcett2006roc}. We average these rates equally over the 73 supported
categories. Unknown presence labels are excluded, with support retained
for all 75 targets. Mathematical definitions are in
Appendix~\ref{app:matched75}.

\paragraph{Supporting alignment and reconstruction measures.}\label{sec:metrics}
Object contrast Capture measures how much regional object-versus-background
information one group reconstructs, adapting block recovery
\citep{fel2026} through foreground/background pooling \citep{wang2019panet}.
Patch AP measures how well native responses rank object patches above
comparison patches \citep{everingham2015retrospective}; reconstruction VE
is one minus normalized MSE relative to the evaluated split's global
mean \citep{gao2024scaling}. Definitions, readout conventions and the
32-patch benchmark versus full-grid audit distinction appear in
Appendix~\ref{app:metric-definitions}.

\paragraph{Complementary organization tests.}
We also examine whether object associations repeat across image sets,
remain stable under appearance changes and depend on selected groups.
These tests use category-response profiles, training-selected subsets,
contribution-PCA maps \citep{fel2026} and fixed linear-probe removals
\citep{alain2016probes,gao2024scaling}; their protocols are given in
Appendix~\ref{app:revision-protocol}.

\paragraph{Uncertainty and cost.}
The 75-group study uses seven training orders (seeds 21--27)
for coherence, concentration, Capture, Patch AP and applicable
reconstruction metrics.
The original COCO GGF--SLiCS benchmark reports five of the seven orders
(21--25) using its 32-patch AP and joint-reconstruction protocol.
Most other comparisons, including paired capacity and continuation, use
three runs. Screens and pilots use one or two
(Appendices~\ref{app:evaluation-overview}--\ref{app:capacity-overview}).
For $n$ orders, we report sample SD ($n-1$ denominator) with shared
initialization and images, training-run Student-$t$ intervals and
conditional image-bootstrap intervals \citep{everingham2015retrospective}.
Appendix~\ref{app:compute-reporting} details fitting time, latency,
memory and partial-FLOP coverage.

\section{Results}\label{sec:results}
We begin with semantic coherence and response concentration, then
examine the object information these groups retain.

\begin{figure*}[t]\centering
\includegraphics[width=\textwidth]{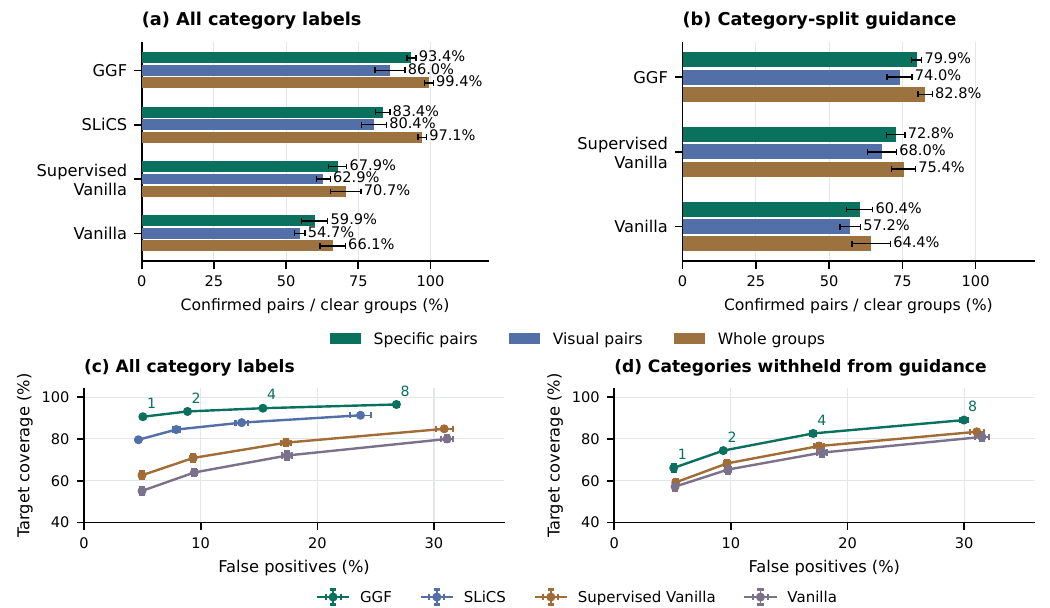}
\caption{\textbf{Coherence and response concentration at 75 groups.}
(a,b) Three blinded AI reviews assess specific-pair meaning, visual
structure and whole-group themes ($\geq5/6$ views). Category-split scores
include all groups. (c,d) Calibrated subsets detect category presence
across 73 supported targets; labels give GGF's subset sizes. Panel (d)
withholds categories from guidance. Means and sample SD (both axes for
curves) summarize variation across seven training orders with shared
initialization and images. Protocols are in Appendix~\ref{app:matched75}.}
\label{fig:matched75-coherence}\label{fig:matched75-concentration}\end{figure*}

\subsection{Coherent meanings at a common group count}\label{sec:matched75-coherence}
With all category labels, the GPT majority gives GGF $93.4\pm1.6$\% specific-pair
coherence, versus $83.4\pm2.5$\% for SLiCS, $67.9\pm3.1$\% for
supervised Vanilla and $59.9\pm4.6$\% for Vanilla
(Figure~\ref{fig:matched75-coherence}(a),
Table~\ref{tab:matched75-coherence}). The whole-group test gives
$99.4\pm1.5$\% for GGF and $97.1\pm1.4$\% for SLiCS, placing
both near ceiling on the broader theme criterion. Parts, textures and
cross-category motifs are valid meanings, so intended category names
do not determine the judgment. Figure~\ref{fig:response-geometry}
shows specific-pair successes and failures.

With category-split guidance, GGF's specific-pair score is $79.9\pm1.7$\%, versus $72.8\pm3.1$\% for
supervised Vanilla, while their whole-group scores are $82.8\pm2.4$\%
and $75.4\pm4.1$\% (Figure~\ref{fig:matched75-coherence}(b)). These
averages cover all groups in each fit. Some folds tie or reverse.

Gemini scores are lower overall, but GGF still leads on all three
coherence measures in both settings. It leads on specific-pair coherence
in all seven orders, averaging 9.5 percentage points above SLiCS with all
labels and 4.7 above supervised Vanilla with split guidance
(Appendix~\ref{app:cross-model}).
\begin{table}[t]\centering\small
\setlength{\tabcolsep}{3pt}
\caption{Blinded coherence at 75 groups: mean $\pm$ sample SD (\%) across seven training orders.
Pair meaning, visual structure and five-of-six whole-group themes are
separate AI-majority endpoints. Every planned pair/group is retained.}
\label{tab:matched75-coherence}
\begin{tabular}{@{}lrrr@{}}\toprule
Method & Meaning & Visual & Whole\\\midrule
\multicolumn{4}{@{}l}{\textit{All category labels}}\\
GGF & $93.4\pm1.6$ & $86.0\pm5.2$ & $99.4\pm1.5$\\
SLiCS & $83.4\pm2.5$ & $80.4\pm4.3$ & $97.1\pm1.4$\\
Supervised Vanilla & $67.9\pm3.1$ & $62.9\pm2.4$ & $70.7\pm5.3$\\
Vanilla BSF & $59.9\pm4.6$ & $54.7\pm1.8$ & $66.1\pm4.4$\\
\midrule
\multicolumn{4}{@{}l}{\textit{Category-split guidance}}\\
GGF & $79.9\pm1.7$ & $74.0\pm4.3$ & $82.8\pm2.4$\\
Supervised Vanilla & $72.8\pm3.1$ & $68.0\pm5.0$ & $75.4\pm4.1$\\
Vanilla BSF & $60.4\pm4.4$ & $57.2\pm3.5$ & $64.4\pm6.5$\\
\bottomrule\end{tabular}\end{table}

\subsection{Concentration beyond the guidance categories}\label{sec:matched75-concentration}
The strongest-response examples show what a group responds to, but not
how broadly or selectively it does so. We therefore select small group
subsets using 400 calibration crops, freeze them and evaluate category
presence on 1,500 audit crops (Appendix~\ref{app:matched75}). In the
full-category condition, one GGF group covers $90.62\pm0.10$\% of
target-present crops, compared with $79.60\pm0.79$\% for SLiCS. Their
false-positive rates are $5.05\pm0.07$\% and $4.66\pm0.11$\%,
respectively, so SLiCS has slightly fewer false positives. Adding groups
increases both coverage and false positives
(Figure~\ref{fig:matched75-concentration}(c,d)).

This coverage advantage extends to categories whose annotations never
entered initialization or guided training. One GGF group covers
$66.13\pm1.87$\% of their target-present crops, versus $59.20\pm1.76$\%
for supervised Vanilla and $57.15\pm1.99$\% for Vanilla. The corresponding
false-positive rates are $5.09\pm0.14$\%, $5.27\pm0.23$\% and
$5.22\pm0.10$\%. Excluded objects can still occur in training images,
and their labels calibrate readouts after fitting. The coverage advantage
thus reflects calibrated category transfer within the existing image pool.

\subsection{Object alignment and reconstruction}\label{sec:released-comparison}
These more interpretable responses also retain object-region information. At 4,096 groups, GGF achieves $42.77\pm0.10$\%
Capture and $74.76\pm0.14$\% Patch AP, versus $21.37\pm0.84$\% and
$51.66\pm1.55$\% for Vanilla. Capture improves for all 75 categories
in every run, and mean Patch AP improves for 72 categories. Paired image
bootstrap intervals support gains in both metrics. The improvement has
a reconstruction cost: GGF explains $76.94\pm0.27$\% of activation
variance, versus $80.35\pm0.39$\% for Vanilla
(Appendix~\ref{app:pilotlatest}). Extending guided training at 1,024
groups retains this alignment--reconstruction tradeoff
(Appendix~\ref{app:capacitylatest}).

\label{sec:supervised-five}
The supervised comparisons place these gains alongside alternative
representations. On COCO and VOC, GGF exceeds SLiCS in Capture, Patch AP
and reconstruction VE in every run. The VOC gains are 21.31, 16.33 and
22.32 percentage points, respectively
(Figure~\ref{fig:released-main}; Appendix~\ref{app:supervised-five}).
GGF also exceeds Concept-SAE on all three metrics, with VOC gains of
42.69, 27.31 and 14.80 points
(Appendices~\ref{app:concept-sae-protocol} and \ref{app:voc-results}).

On withheld categories in the 75-group study, GGF gains 2.63
Capture points over supervised Vanilla. Full-grid Patch AP is
$22.92\pm1.17$\% for GGF and $22.19\pm0.69$\% for supervised
Vanilla, with GGF ahead in 5 of seven orders. Vanilla BSF has the
highest reconstruction VE in both guidance settings
(Tables~\ref{tab:matched75-AP} and~\ref{tab:matched75-reconstruction}).

Separate analyses compare the 1,024-group BSFs with native supervised
alternatives. GGF's intended
category accounts for $77.1\pm0.4$\% of positive category-response profile
weight, versus $63.9\pm1.6$\% for supervised Vanilla. Its profiles are the most reproducible across two image halves (Appendix~\ref{app:category-fingerprints}). Across six categories, controlled rotation, scale and brightness changes
give
GGF $80.92\pm0.33$\% worst-condition recognition AP, close to
supervised Vanilla's $80.52\pm1.22$\%. Recognition is clearly higher than Concept-SAE's. Each transformation also gives GGF higher regional Capture than both
BSF controls (Appendix~\ref{app:controlled-robustness}).
Both reuse existing images in configurations separate from the 75-group
study.

\subsection{What initialization and guidance contribute}\label{sec:ablation}
Supervised Vanilla shares GGF's object objectives and classifier heads,
allowing us to distinguish their benefit from that of initialization and
adaptive weights. At 1,024 groups on COCO, these shared components recover
most of the Capture gain. GGF adds 0.73 points, with a 95\% interval of
$[-0.05,1.52]$, and 10.91 Patch AP points, with an interval of
$[10.70,11.11]$, while reconstruction VE decreases by 1.35 points.
On VOC, GGF gains 1.72 Capture and 1.36 Patch AP points at a 0.37-point
VE cost. Both object metrics improve in every run, although the intervals
support an additional gain only for Capture
(Appendices~\ref{app:current-supervised-control}, \ref{app:voc-results}).

The component comparison in Table~\ref{tab:components-main} explains this
pattern further. Initialization primarily improves localization, while
adding object losses supplies most of the Capture gain. Fixed guidance
retains 97.6\% of the full Capture improvement over Vanilla, and adaptive
weighting adds a clear improvement only for Capture
(Appendix~\ref{app:components-current}).
\begin{table}[t]\centering\small
\caption{Component ablation: mean $\pm$ sample SD (\%). All arms use 4,096 groups, 12 directions, Top-32 and 10,640 updates. Fixed guidance uses training-derived weights.}\label{tab:components-main}
\begin{tabular}{@{}lrrr@{}}\toprule
Arm & Capture & Patch AP & VE\\\midrule
Vanilla & $21.37\pm0.84$ & $51.66\pm1.55$ & $80.35\pm0.39$\\
Mask + rec. & $23.56\pm0.05$ & $68.55\pm0.17$ & $79.56\pm0.23$\\
Fixed guidance & $42.25\pm0.12$ & $74.63\pm0.19$ & $77.33\pm0.20$\\
GGF & $42.77\pm0.10$ & $74.76\pm0.14$ & $76.94\pm0.27$\\
\bottomrule\end{tabular}\end{table}

\subsection{Capacity and known-factor recovery}\label{sec:widthbalance}
Increasing capacity offers an additional way to improve recovery. With
equal additional updates, expanding groups from eight to twelve
directions adds 0.82 Capture and 1.60 VE points, at the cost of 50\% more
parameters and active coefficients (Appendix~\ref{app:capacitylatest}).

\label{sec:synthetic}
We also test consolidation where the generating factors are known.
On Manifold Zoo, GGF improves recovery and reconstruction over the
archived BSFs, and every assigned group has factor-recovery score
$R^2>0.90$ in every run
(Appendix~\ref{app:manifoldzoo}, Table~\ref{tab:synthetic-seeds}).

\section{Discussion}\label{sec:limitations}
BSFs give concepts room to occupy several directions within one group.
GGF uses that capacity to improve interpretability by making responses
more semantically coherent and concentrating object-related information
within individual groups.

The selected bird and doughnut readouts make the semantic benefit concrete
(Figures~\ref{fig:blinded-birds} and~\ref{fig:blinded-doughnuts};
Appendix~\ref{app:matched-blinded-examples}). GGF repeatedly marks feathers
or doughnut surfaces, while SLiCS combines hair with feathers or different
topped foods. The Vanilla variants also include fingers, sky or unrelated
regions. GGF has three confirmed specific pairs in both cases, compared
with one for each alternative. Its advantage is a description that remains
specific across varied examples, allowing the same group to retain a
consistent meaning (Appendix~\ref{app:matched75}).

The consolidation tests quantify a complementary benefit: information
available within individual groups. GGF's higher Capture indicates more
regional information. Removing its contributions also weakens fixed readouts
(Appendices~\ref{app:planned-consolidation}, \ref{app:voc-group-organization}).
The 75-group coverage curves show that small GGF subsets detect objects
in more crops at the measured false-positive cost
(Appendix~\ref{app:matched75}). Within the existing image pool, this
advantage extends to categories excluded from guidance even where Patch
AP is close to supervised Vanilla (Table~\ref{tab:matched75-AP}).
Useful information can still remain in other groups.

That organization should also persist as an object's appearance changes. In the six-category controlled 2D
study, GGF and supervised Vanilla recognize objects similarly under
rotation, resizing and brightness changes, but GGF retains greater mean
regional Capture (Figure~\ref{fig:category-fingerprints}(b,c);
Appendix~\ref{app:controlled-robustness}). GGF's category preferences also
reproduce more consistently across image halves in the guided-category
development study (Table~\ref{tab:fingerprint-groups}). Jointly
rotating a BSF's encoder and decoder can change direction profiles while
preserving group norms and contributions \citep[Sec.~3.5]{fel2026}. Whole-group focus
and reproducibility therefore assess object association more directly
(Appendix~\ref{app:category-fingerprints}).

The component studies trace most Capture gains to object losses.
Initialization mainly improves localization, and adaptive weighting adds
a smaller Capture benefit (Appendix~\ref{app:components-current}). Recovery
of known synthetic factors supplies a direct consolidation check
(Appendix~\ref{app:manifoldzoo}). Capacity studies place these gains
alongside reconstruction, size and activity costs
(Appendices~\ref{app:capacitylatest}, \ref{app:capacity}).

These improvements remain efficient to train.\label{sec:resource-costs}
At 1,024 groups, GGF uses $9.3\times$ fewer learned parameters on COCO
and $4.9\times$ fewer on VOC than Concept-SAE, excluding the frozen
backbone and probes (Appendix~\ref{app:concept-sae-protocol};
Table~\ref{tab:compute-main}). On VOC, the 19-group comparison gives a
$18.33\times$ mean training-loop speedup over SLiCS on A100. At 1,024
groups on T4, GGF is $1.66\times$ faster than Concept-SAE, with 3.4\%
overhead over supervised Vanilla
(Appendices~\ref{app:fit-costs}, \ref{app:execution-profiles}). This makes
coherent, concentrated groups practical within the same BSF architecture.

\section{Conclusion}\label{sec:conclusion}
GGF shows how object guidance can make block-sparse visual representations
easier to interpret. Its groups more often retain a recognizable meaning
across images, and responses to annotated categories are accessible
through fewer groups. This concentration advantage also extends to
categories excluded from guidance, suggesting that the resulting
organization can be useful beyond the supervision vocabulary.
Reconstruction, localization and appearance analyses provide complementary
evidence about the visual information these groups retain. GGF makes object guidance a practical means of learning more coherent,
interpretable groups that consolidate object-related information.

\clearpage
{\small
\bibliographystyle{ieeenat_fullname}
\nocite{goodfire2026}
\bibliography{references}
}
\section*{Supplementary material}
\renewcommand{\theHfigure}{appendix.\arabic{figure}}
\renewcommand{\theHtable}{appendix.\arabic{table}}
\renewcommand{\theHequation}{appendix.\arabic{equation}}
\setcounter{figure}{0}
\setcounter{table}{0}
\setcounter{equation}{0}
\renewcommand{\thefigure}{S\arabic{figure}}
\renewcommand{\thetable}{S\arabic{table}}
\renewcommand{\theequation}{S\arabic{equation}}
\appendix
\raggedbottom
\captionsetup{font=footnotesize}
\setlength{\floatsep}{8pt plus 1pt minus 1pt}
\setlength{\textfloatsep}{12pt plus 1pt minus 1pt}
\setlength{\intextsep}{8pt plus 1pt minus 1pt}
\setlength{\dblfloatsep}{8pt plus 1pt minus 1pt}
\setlength{\dbltextfloatsep}{12pt plus 1pt minus 1pt}
\setlength{\abovecaptionskip}{6pt}
\setlength{\belowcaptionskip}{0pt}
\makeatletter
\AtBeginEnvironment{tabular}{%
  \ifdim\f@size pt>8.5pt\fontsize{8.5}{10}\selectfont\fi
  \ifdim\tabcolsep>2.5pt\setlength{\tabcolsep}{2.5pt}\fi
  \renewcommand{\arraystretch}{.98}}
\makeatother

\setcounter{topnumber}{6}
\setcounter{bottomnumber}{4}
\setcounter{totalnumber}{12}
\setcounter{dbltopnumber}{6}
\renewcommand{\topfraction}{.95}
\renewcommand{\bottomfraction}{.90}
\renewcommand{\dbltopfraction}{.95}
\renewcommand{\textfraction}{.05}
\renewcommand{\floatpagefraction}{.95}
\renewcommand{\dblfloatpagefraction}{.95}

\makeatletter
\setlength{\@fptop}{0pt}
\setlength{\@fpsep}{8pt}
\setlength{\@fpbot}{0pt plus 1fil}
\setlength{\@dblfptop}{0pt}
\setlength{\@dblfpsep}{8pt}
\setlength{\@dblfpbot}{0pt plus 1fil}
\makeatother
\section*{Appendix contents}
\begingroup\small
\newcommand{\appentry}[3]{\noindent\hyperref[#1]{\textbf{#2.} #3}\nobreak\dotfill\pageref{#1}\par\smallskip}
\appentry{app:implementation-overview}{A}{Implementation and training}
\appentry{app:evaluation-overview}{B}{Data and evaluation protocol}
\appentry{app:comparisons-overview}{C}{Model comparisons}
\appentry{app:supervision-overview}{D}{Components and loss-weight studies}
\appentry{app:capacity-overview}{E}{Capacity and continuation studies}
\appentry{app:geometry-overview}{F}{Response geometry and synthetic recovery}
\appentry{app:revision-protocol}{G}{Group organization and functional diagnostics}
\appentry{app:compute-reporting}{H}{Fitting costs, execution profiles and FLOP coverage}
\appentry{app:qualitative-current}{I}{Qualitative comparisons}
\endgroup
\medskip

\section{Implementation and training}\label{app:implementation-overview}

\subsection{Feature construction and objective details}\label{app:featuredetails}\label{detail:sec:masks}\label{detail:sec:loss}

\paragraph{Normalization and masks.}
For training-image indices $\mathcal I_{\rm tr}$, let
$N_{\rm tr}=|\mathcal I_{\rm tr}|$ and let $a_{ip}\in\mathbb R^d$
be DINOv3's patch vector at position $p\in\{1,\ldots,196\}$, with $d=768$.
The training positional mean is $\mu_p$. The positive scale $s_{\rm norm}$
makes average squared training coordinates approximately one and is
unrelated to random seeds. With $\|\cdot\|_2$ denoting Euclidean length:
\begin{equation}
\begin{aligned}
\mu_p&=\frac{1}{N_{\rm tr}}\sum_{i\in\mathcal I_{\rm tr}}a_{ip},
\quad x_{ip}=s_{\rm norm}(a_{ip}-\mu_p),\\
s_{\rm norm}&=\sqrt{\frac{d}{\frac{1}{196N_{\rm tr}}
\sum_{i\in\mathcal I_{\rm tr}}\sum_{p=1}^{196}\|a_{ip}-\mu_p\|_2^2}}.
\end{aligned}\label{eq:norm}
\end{equation}
All splits reuse training-only means and scale, without coordinate rescaling
or whitening. The frozen backbone is \path{facebook/dinov3-vitb16-pretrain-lvd1689m}.
Its final patch tokens retain attention context. We discard the CLS token and
four registers. Images use bicubic resizing and masks use nearest-neighbor
resizing with identical crop geometry. Mask thresholds and object contrasts follow Section~\ref{sec:masks}.
Appendix~\ref{app:allocation} specifies stored mask precision and split allocation.

\paragraph{Loss averaging.}
The reconstruction loss $\Lrec$ compares predictions from noisy inputs with
clean targets. Let $B$ be a batch of clean $d$-dimensional patch vectors and
$|B|$ its number of patches. For each $x\in B$, let $\widetilde{x}$ be the
noisy input, $\widehat{x}(\widetilde{x})$ its reconstruction, and $\epsilon$
a vector of independent standard normal draws. The scalar
$\operatorname{std}(B)$ is the standard deviation over all batch entries:
\begin{equation}
\widetilde{x}=x+0.1\operatorname{std}(B)\epsilon,\qquad
\Lrec=\frac{1}{|B|d}\sum_{x\in B}\|x-\widehat{x}(\widetilde{x})\|_2^2.
\label{eq:rec}
\end{equation}
The averaging gives patches and coordinates equal weight. Reconstruction
and contrast update both encoder and decoder. Detection updates the encoder
and the two-parameter heads, with no decoder gradient. Object losses use
clean inputs. Their averaging, head initialization and fixed training
normalization are specified in Equations~\ref{eq:logit},
\ref{eq:caploss}, and~\ref{eq:detloss}.
Training uses all 196 patches. Detection audits rank raw group norms on
the deterministic sample described in Appendix~\ref{app:allocation}.

\paragraph{Protocol choices and empirical support.}
Mask/support cutoffs, normalization, noise and warm-up length are study
conventions, not literature-prescribed constants. Component/loss controls
(Appendix~\ref{app:supervision-overview}), capacity/continuation studies
(Appendix~\ref{app:capacity-overview}) and mask-threshold sensitivity
(Appendix~\ref{app:secondary}) test selected choices, not every constant.

\subsection{Mask-derived starting directions}\label{app:initdetails}

The fresh configuration uses the leading $r=12$ directions. The original
three-direction study uses $r=3$. For category $c$, let
$\mathcal F_c$ contain normalized training patch vectors
covered at least 0.5 by the union of that category's masks and at most 0.1 by
crowd masks. Crowd annotations mark regions without separate instance masks.
For $x\in\mathcal F_c$, $x^\top$ denotes transpose. We compute the second-moment
matrix $S_c=\sum_{x\in\mathcal F_c}xx^\top$ in float64. Its $r$ leading
unit eigenvectors retain maximal foreground energy under rank-$r$ orthogonal
projection and initialize one decoder group. In the original three-direction runs, we obtain these
directions from the saved singular value decomposition (SVD) table, without subtracting a category mean after shared normalization.
This is the sense in which the calculation is uncentered.
Each direction is oriented so positive projections contain at least as much
squared energy as negative ones. GGF coefficients remain signed. We reuse the
verified float32 table to avoid differences from recomputing the SVD.

In the original 1,024-group study, groups 0--74 receive the category
directions in order. Groups 75--1023 retain
same-seed random directions. The encoder starts as the decoder transpose,
with zero bias. Across seeds, the 75 SVD groups are identical. The 949 random
groups, noise, and data order vary. The runs therefore share one SVD dictionary rather than independently
estimating it. After warm-up, we search all 1,024 groups for each category's
contrast and detector readouts, the groups used to score it. Initial category
slots are not enforced: categories may share groups, and contrast and detection
may choose different groups.

\subsection{Main GGF training algorithm}\label{app:mainalgorithm}
Let $X$ denote normalized training patch activations, $Y$ their aligned
masks, $\mathcal V$ the development split, and $s$ a random seed. The
dictionary has $G$ groups of width $r$, with $k$ retained per patch.
$T$ stores fixed training-selected category readouts. $w_C,w_D$ are
the bounded object-loss weights. Development data record scores and
eligibility. They do not supply the controller's gradients or targets.
\begin{agfalgorithm}[GGF: Initialization and Guided Learning]
\label{alg:agf-main}
\textbf{Input:} $X,Y,\mathcal V,s,G,r,k$ and the fixed training schedule.
\agfstep{Initialize one category group from each category's TRAIN foreground second moment and randomly initialize reserve groups.}
\agfstep{Run reconstruction warm-up and choose $T$ using TRAIN masks and activations.}
\agfstep{Initialize logistic classifier heads and bounded loss weights $w_C,w_D$.}
\agfstep{\textbf{for} each scheduled guided update \textbf{do}}
\agfstep[1]{Sample training images and compute reconstruction on noisy inputs and object contrast/detection on clean inputs.}
\agfstep[1]{Retain the task weights for this model update. Every ten updates, use encoder gradient norms and loss ratios to adjust only the two log weights with detached targets.}
\agfstep[1]{Update encoder, decoder and logistic classifier heads with the retained weights and apply controller changes to the next step.}
\agfstep[1]{Unit-normalize decoder directions and periodically save model, optimizer, controller and sampling states.}
\agfstep{\textbf{end for}}
\agfstep{Freeze the scheduled endpoint and fixed readouts before evaluation.}
\agfstep{\textbf{return} GGF and its training records. Group membership never changes.}
\end{agfalgorithm}

\subsection{Bounded gradient-based weight adjustment}\label{app:controller}
The controller is inspired by GradNorm~\citep{chen2018gradnorm}, with a fixed
reconstruction anchor and bounded object weights rather than canonical
GradNorm's jointly normalized task weights. Let $j\in\{R,C,D\}$ index
reconstruction, contrast and detection. Let $L_j(u)$ be the corresponding
unweighted loss at update $u$, and $L_j^0$ its mean over the first 20 batches
whose three losses are finite and greater than $10^{-12}$. Let $\theta_E$
collect the shared encoder matrix and bias, and $n_j(u)$ denote the
Euclidean norm of that task's gradient on $\theta_E$. Write $w_R=1$,
$w_C=\lambda_C$ and $w_D=\lambda_D$. Define the normalized loss ratios
$\rho_j$, weighted gradient strengths $s_j$, and target strengths $a_j$:
\begin{equation}
\begin{split}
n_j&=\|\nabla_{\theta_E}L_j\|_2,\qquad
\rho_j=\frac{L_j/L_j^0}{\frac13\sum_{h\in\{R,C,D\}}L_h/L_h^0},\\
s_j&=w_jn_j,\qquad
a_j=\left(\frac13\sum_{h\in\{R,C,D\}}s_h\right)\rho_j^{1.5}.
\end{split}\label{eq:balance-target}
\end{equation}
A larger $\rho_j$ means slower relative loss reduction and gives that task
a larger target strength. Every ten model updates, separate Adam at learning
rate 0.01 adjusts $\log w_C$ and $\log w_D$ to reduce
$|s_C-a_C|+|s_D-a_D|$. Gradient norms and target strengths are detached:
the controller differentiates only through its two weights, without
second-order model derivatives. It skips adjustment if the reference is
unavailable or any shared gradient norm is at most $10^{-12}$.

Weights start at 0.05 and 0.01. After each controller step, the log
weights are projected to contrast bounds $[0.0125,a]$ and detection bounds
$[0.0025,a]$. The cap $a$ is 0.20 in the first width/weight study, 0.40
in subsequent capacity studies and 0.60 in the fresh 4,096-group comparison.
New weights apply to the next model update. Reconstruction remains at one.
Only training batches drive the controller. The empirical cap comparison
is in Appendix~\ref{app:higherbound}.

\subsection{Released baseline implementations}\label{app:baselines}

The pinned Goodfire classes \texttt{VanillaBSF}, \texttt{GrassmannianBSF},
and \texttt{GroupLassoBSF} are unchanged. Source identities accompany the paper.
Our sampling, fitting, and audits are separate, so this is not full replication.

\paragraph{Vanilla.}
Decoder directions start at unit length. The encoder starts as their transpose
with zero bias, then learns independently. Top-$k$ retains the $k$ largest
group norms with signed coefficients. Training minimizes reconstruction MSE,
with sparsity supplied by gating alone.

\paragraph{Grassmannian.}
QR factorization keeps each decoder group mutually perpendicular and unit
length whenever accessed. Its tied encoder uses those directions with one
unconstrained learned multiplier per group. Top-$k$ selects the requested $k$
active groups by scaled coefficients. The loss is reconstruction error.

\paragraph{Released Group Lasso.}
For group $g$, let $\theta_g\geq0$ be its learned activation threshold,
$a_g$ its encoder coefficient vector, and $z_g$ its coefficients after gating.
The implementation retains the entire signed vector when its length exceeds
the threshold: $z_g=a_g\ind[\|a_g\|>\theta_g]$, where the indicator is one
when its condition holds and zero otherwise. It does not shrink retained
coefficients. Thresholds start at a quantile of observed norms. During
backpropagation, a normal-density kernel approximates the threshold's
derivative, with bandwidth tracking the norm scale. Let $L_0$ count active
groups, $\mathbb E[L_0]$ be their mean count, and $\lambda$ be the
activity-penalty multiplier. Reconstruction loss plus
$\lambda\mathbb E[L_0]$ targets mean activity 16 by updating $\lambda$
separately. This hard gating differs from textbook group lasso, which
shrinks coefficients through soft thresholding.

\subsection{Numerical implementation and saved state}\label{app:implementation}
\paragraph{Numerics.}
We resize and crop images explicitly while retaining the processor's pixel
rescaling and normalization. Cached half-precision features are loaded as
float32. Positional means accumulate in float64 and are saved as float32.
The shared scale uses float64 energy accumulation over float32-centered features. Verified memory
maps stream normalized float32 arrays under Equation~\ref{eq:norm}, without
coordinate-wise standardization, PCA, or whitening.

\paragraph{Records and environment.}
Logs record loss values and weights, gradient norms and cosines, batch support,
preservation checks, selected and final states, group activity, inactive groups,
and shared readouts. Checkpoints retain the model, optimizer, heads,
and random-number states, allowing recovery from local or verified Drive saves.
Work since the last save may repeat. Batch hashes verify pairing rather than
bitwise-identical optimization.

The scaled run records T4, PyTorch 2.11.0+cu130, NumPy 2.1.3, SciPy 1.16.3, scikit-learn 1.6.1, Transformers 4.57.3, pycocotools 2.0.10 and einops 0.8.2. Final extraction uses Transformers 5.18.0 with a verified state-key rename and unchanged tensor values. The VOC SLiCS fits and execution profiles use A100-SXM4-40GB, CUDA 13.0,
Python 3.13.15 and the same PyTorch/NumPy/SciPy versions, with TF32 disabled.
Run-specific versions and identities are retained. Matched updates do not imply equal FLOPs or time.

\paragraph{Saved identifiers.}
GGF denotes mask-derived Initialization followed by Guided Learning with
fixed group membership. Adaptive refers to the task-loss weights.
Saved \texttt{agf\_g}, \texttt{GGF} and guided display labels map to
GGF in \path{evidence/terminology.csv}. Run IDs, model keys and source
hashes identify the saved records. Evidence paths in this Appendix are
relative to the companion archive's \path{code/} directory.

\subsection{Data stewardship and author responsibility}\label{app:disclosures}
COCO can depict identifiable people. Object-response scores neither
establish demographic fairness nor justify sensitive-attribute inference.
Data, pretrained weights, and third-party code remain subject to their
respective terms. Code, model identities, and evaluation records support
reproduction. Availability of any redistributed assets must be verified
before release. An AI assistant supported coding, analysis, figures, and
writing. The authors remain responsible for the results and manuscript.

\section{Data and evaluation protocol}\label{app:evaluation-overview}
\subsection{Mask construction and split allocation}\label{app:allocation}
\paragraph{Data roles.}
We split COCO train2017 by image ID, so no image belongs to two splits.
Training data update parameters and supervised assignments; development
selects checkpoints. Some audits select readouts on calibration data, whereas the matched
1,024-group baseline comparison fixes them from training data
(Appendix~\ref{app:matched-baselines}). Test data only
supply scores. Coverage checks use annotations,
before model outcomes, to retain 75 categories. Sampling enough examples
of scarce categories makes this a coverage-enriched benchmark, not a sample
with natural COCO frequencies.

The shared data contain 2,000 training images and three 400-image sets for
selection, confirmation, and readout calibration. In the 75-group study,
the confirmation cohort serves post-fit calibration; the separate
readout-calibration cohort is unused (Appendix~\ref{app:matched75}).
The initial baseline screen uses
800 test images. Scaled guided learning uses a separate 1,500-image test.
The matched baseline comparison uses another 1,500 images, excluding all
recorded split IDs and requiring at least 20 usable images per category.
It reuses the two development sets but freezes all runs before
extracting the final test features. The following subsection details coverage requirements for contrast and detection.

Training schedules differ across experiments and are recorded with their
results. The fixed-weight loss ablations select eligible development
checkpoints. The matched-comparison studies report scheduled final endpoints.
Experiments use seeds 21, 22 and 23 unless specified otherwise.

\paragraph{Geometry and storage.}
Let $W$ and $H$ be the original image width and height in pixels. The resize
factor is $t=256/\min(W,H)$, giving dimensions
$(\lfloor tW\rfloor,\lfloor tH\rfloor)$, where the floor symbol rounds down
to an integer. We center-crop to 224 pixels by halving the excess width and
height and rounding as in the saved implementation. COCO's \texttt{annToMask}
converts polygon outlines or compressed pixel runs into binary masks.
Non-crowd instances stay separate for object contrast, while detection uses
the union of all instances of a category. Different categories may overlap.
Crowd masks form a separate union used to exclude uncertain regions.

Each patch's category-union coverage is rounded to an 8-bit unsigned integer
and divided by 255 when loaded. Instance and crowd masks retain fractional
coverage. All thresholds apply to these stored values. An eligible contrast
requires known foreground coverage totaling at least two full patches and
at least four category-negative patches. Its foreground average uses all
known foreground coverage, including patches below the detection-positive
threshold.

\paragraph{Coverage rule for sampled-patch evaluation.}
For sampled-patch evaluation, a usable image must provide both a
positive and a negative patch in its fixed 32-patch audit sample, as well
as an eligible contrast computed from all patches. If $\mathrm{imageID}$
is the photograph's integer COCO identifier, the sampling seed is
$61{,}002+\mathrm{imageID}$. Coverage checks, calibration, and test use this
same sample. A category's presence somewhere in the original image is
insufficient if cropping removes usable evidence. The allocator first
supports scarce categories, then fills shared prevalence targets. Quotas
do not equalize prevalence because one image can support several categories.
These pool and usability rules leave 75 categories, excluding skis, sports
ball, baseball bat, toaster, and hair drier.
\begin{figure}[t]
\centering\includegraphics[width=205.2bp]{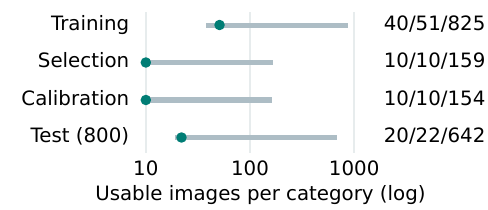}
\caption{Usable-image coverage in the COCO split allocation. Lines show category
minima to maxima. Dots show medians. Labels give minimum/median/maximum.
Counts require both audit detection evidence and an eligible object contrast.}
\label{tab:coverage}
\end{figure}
\paragraph{Scaled fresh-test allocation.}
Before fitting, a constrained solver found that 800 images could not meet
the declared quotas within the available pool. The revised allocation uses
1,500 images, the same 75 categories, and minimum quotas of 20 usable images
per category. Annotation-only scans add at most 500 candidate images for
each undercovered category. A binary covering solver selects a core set
meeting the quotas. Prevalence-based filling completes the split. We verify
the saved image IDs and coverage. Preparation stops if quotas cannot be met,
rather than relaxing them. The matched baseline comparison uses separate
quotas for contrast, positive detection and negative detection evidence.

\paragraph{Identity checks.}
We verify image IDs, feature/mask shards and crop geometry. The scaled
fresh test excludes the recorded confirmation and test cohorts; paired models
share images, normalization and category denominators. Readout and
checkpoint-selection roles are specified above.

\subsection{Additional audit conventions}\label{app:secondary}

\paragraph{Where the metrics are defined.}\label{app:metric-definitions}
Each metric has one canonical mathematical definition. Coherence votes,
presence-based concentration and prediction overlap are defined in
Appendix~\ref{app:matched75}; the supporting alignment and reconstruction
scores are given below. Synthetic factor recovery is defined in
Appendix~\ref{app:manifoldzoo}. Fixed-probe removal effects are in
Appendix~\ref{app:planned-consolidation}, worst-condition recognition in
Appendix~\ref{app:controlled-robustness}, and response fingerprints, focus,
reproducibility and direction agreement in
Appendix~\ref{app:category-fingerprints}. Resource ratios and per-image
operation counts are defined in Appendix~\ref{app:compute-reporting}.

\paragraph{Supporting alignment and reconstruction scores.}
In evaluation formulas, $\mathcal C$ denotes the evaluated category set.
Let $\mathcal O_c$ contain its eligible objects of category $c$, and
$g_c$ be that category's fixed contrast readout. Regional Capture is
\begin{equation}
C_c=1-\frac{\sum_{o\in\mathcal O_c}\|\delta x_o-\delta r_{o g_c}\|_2^2}
{\sum_{o\in\mathcal O_c}\|\delta x_o\|_2^2},\quad
\overline C=\frac1{|\mathcal C|}\sum_c C_c.
\label{eq:auditcap}\end{equation}
Errors and target energies are pooled within each category before
categories receive equal weight. Concept-SAE uses its native-token
calculation (Appendix~\ref{app:concept-sae-capture}). BSF patch scores are
gated norms, while SLiCS uses independent NNLS projections. The sampled
benchmark uses 32 patches per image; the 75-group audit uses all 196.
Patch AP measures ranking, rather than COCO segmentation AP. Reconstruction
VE is one minus normalized MSE relative to the evaluated split's global
mean \citep{gao2024scaling}.

Contrast uses all patches and eligible non-crowd instances. The sampled-patch benchmark evaluates 32 patches per image. Unlike the instance-normalized training loss, Equation~\ref{eq:auditcap} weights larger contrasts more heavily.
Without across-instance centering, it is not regression $R^2$.

Because Capture and detection can use different readouts, their scores need
not demonstrate both properties in one group. AP treats tied scores together,
excludes ambiguous and crowd-invalid patches, and sums precision-weighted
recall increments, measuring ranking rather than a fixed detection threshold
or bounding-box overlap.

\paragraph{Patch ranking and reconstruction.}
Let $\mathcal C_{\rm AP}$ contain categories with valid positive and
negative patches. Let $P_{c\ell}$ and $R_{c\ell}$ be precision and recall after including
all patches at distinct score level $\ell$ in descending order, with $R_{c0}=0$.
The non-interpolated average precision and category-macro score are
\begin{equation}
\mathrm{AP}_c=\sum_\ell(R_{c\ell}-R_{c,\ell-1})P_{c\ell},\quad
\overline{\mathrm{AP}}=\frac1{|\mathcal C_{\rm AP}|}\sum_c\mathrm{AP}_c.
\label{eq:patch-ap}\end{equation}
Tied scores enter together; ambiguous and crowd-invalid patches are
excluded. Categories must contain valid positives and negatives.
For $N$ evaluated $d$-dimensional vectors $x_i$ and their reconstructions
$\widehat x_i$, let $\mu_{\mathcal E}$ be the evaluated split's global mean. Define squared error
$E=\sum_i\|x_i-\widehat x_i\|_2^2$ and total variation
$V=\sum_i\|x_i-\mu_{\mathcal E}\|_2^2$. Then
\begin{equation}
\mathrm{MSE}=\frac{E}{Nd},\qquad \mathrm{VE}=1-\frac{E}{V}.
\label{eq:reconstruction-metrics}\end{equation}
MSE uses the training-normalized feature coordinates. Natural-image VE
uses the evaluated split's global mean $\mu_{\mathcal E}$; synthetic VE
instead uses the training mean, as specified in
Appendix~\ref{app:manifoldzoo}. Thus MSE and $1-\mathrm{VE}$ have different
denominators. No reconstruction score is imputed for independent SLiCS
projections.

\paragraph{Why measure object contrast Capture?}
Single-block recovery in Fel et al.~\citep{fel2026} tests whether a block
reconstructs a known synthetic factor's contribution. Natural images do
not provide that latent contribution. We instead construct an observable
regional target: the difference between mask-weighted object features and
category-negative features in the same image. PANet~\citep{wang2019panet}
provides precedent for separately pooling foreground and background
features. Their difference cancels a spatially constant feature offset.
It does not remove every contextual influence. Our choice tests whether
one group carries the measured object-versus-comparison contrast, rather
than claiming access to a ground-truth concept vector.

Normalized reconstruction error is also used to evaluate sparse
autoencoders \citep{gao2024scaling}. Our denominator differs: it is the
energy of the observed contrast, making zero contrast the reference
prediction. For a nonzero target $d$ and prediction $r$, the score is
$1-\|d-r\|^2/\|d\|^2=(2d^\top r-\|r\|^2)/\|d\|^2$.
Thus exact reconstruction scores one, no reconstructed contrast scores
zero, and worse predictions score below zero. Increasing response
magnitude alone cannot improve the score indefinitely: $r=ad$ scores
$2a-a^2$, maximized at $a=1$. A common rescaling of target and prediction
leaves Capture unchanged. Pooling errors and target energies within a
category avoids averaging unstable ratios for individual weak contrasts.
Macro-averaging then gives categories equal weight. This construction is
our adaptation, not the centered factor $R^2$ of the synthetic benchmark.

Capture complements AP: a group may rank object patches correctly while
reconstructing little of their contrast, or reconstruct a contrast while
responding elsewhere. Same-group AP, additional-group Capture and
matched-control removal diagnostics therefore qualify consolidation
claims. Neither Capture nor those diagnostics establish semantic
exclusivity or recovery of every factor of variation.

\paragraph{Additional reconstruction and detection diagnostics.}
Calibration maximizes F1, the harmonic mean of precision and recall. With
$TP$ correctly detected positives, $FP$ falsely detected negatives, and $FN$
missed positives, test reports precision, recall,
F1 and intersection over union:
\begin{equation}
\mathrm{F1}=\frac{2TP}{2TP+FP+FN},\qquad
\mathrm{IoU}=\frac{TP}{TP+FP+FN}.
\label{eq:f1-iou}\end{equation}
The sampled-patch diagnostic returns zero for a zero denominator;
the presence audit retains undefined precision separately. Recall and
false-positive rates use Equation~\eqref{eq:presence-rates}. Mask sensitivity changes foreground thresholds
with fixed readouts. It does not optimize the two/four support requirements
or validate every threshold. A calibration-greedy audit adds up to four
groups while contrast error decreases, without refitting coefficients or
directions. All-group Capture uses full reconstruction. These diagnostics measure reconstruction by selected groups.

A separate diagnostic fits a balanced logistic classifier to
responses from up to four groups ranked on calibration data. Its response
normalization also uses calibration data. The inverse regularization-strength
parameter is $C=1$, and the optimizer allows 500 iterations. Its extra flexibility differs from both single-group AP and the training heads. Reports retain category prevalence/support, mask eligibility, group IDs, thresholds, and negative outcomes. Appendix~\ref{app:heatmaps} discloses visual selection.

\paragraph{Uncertainty.}
For $n$ run-level scores $x_s$, indexed by $s=1,\ldots,n$, let
$\overline x$ be their mean and $s_x$ their sample SD:
\begin{equation}
\overline x=\frac1n\sum_s x_s,\qquad
s_x=\sqrt{\frac{\sum_s(x_s-\overline x)^2}{n-1}}.
\label{eq:mean-sample-sd}\end{equation}
For paired scores $x_s,y_s$, let $d_s=x_s-y_s$ have mean $\overline d$
and sample SD $s_d$. With $t_{0.975,n-1}$ the Student-$t$ quantile for
$n-1$ degrees of freedom, the reported paired 95\% interval is
\begin{equation}
\overline d\ \pm\ t_{0.975,n-1}\,s_d/\sqrt n.
\label{eq:paired-run-interval}\end{equation}
These intervals concern the declared run variation. The 75-group primary
figures report sample SD, not confidence bands. Percentile image-bootstrap
intervals use the 0.025 and 0.975 quantiles of the paired resampled
differences described below.

Paired bootstraps resample images with the same instance/patch multiplicities
for both models. Means retain all 75 categories. Unsupported full-cohort draws
are counted as invalid. The scaled comparison has 100 valid and zero invalid
draws. Matched baseline intervals use 200 valid draws. Unless stated otherwise, intervals are percentile intervals conditional
on fitted models and readouts. They omit training-seed, tuning, assignment
and researcher-selection uncertainty and receive no multiplicity correction.
Image-bootstrap VE differences remain point estimates. Training-run
VE intervals are reported where available. The continuation and capacity experiments reuse the inspected
1,500-image benchmark, so their comparisons are not independent confirmation.
The 4,096-group audit also reports Student-$t$ intervals across training starts.
The VOC SLiCS comparison uses paired training-run intervals rather than
image-bootstrap intervals (Appendix~\ref{app:voc-slics}). This limited replication does not support a well-powered population-level test.

\paragraph{Replication across comparisons.}
Most comparisons use three training runs per setting, including paired
capacity and continuation studies. The COCO GGF--SLiCS comparison uses
five training orders sharing one initializer. The baseline, loss-weight
and width/activity screens, Vanilla group-count check, cap pilot,
supervised Vanilla calibration and random-start diagnostic each use one
run per setting. The auxiliary 1,024-group pilot uses two runs.

\section{Model comparisons}\label{app:comparisons-overview}
We begin with the matched supervised Vanilla control, which distinguishes
fixed object-supervised training from GGF's initialization and controller.
The SLiCS and Concept-SAE comparisons then place these improvements
alongside alternative supervised representations. VOC2012 tests the
training recipe on another dataset. Capacity and baseline studies are reported separately because they use
different cohorts, budgets and readouts.
Component controls are collected in Appendix~\ref{app:components-current}.
Group-organization diagnostics are collected in Appendix~\ref{app:revision-protocol}.

\begin{figure*}[!tp]\centering
\includegraphics[width=\textwidth]{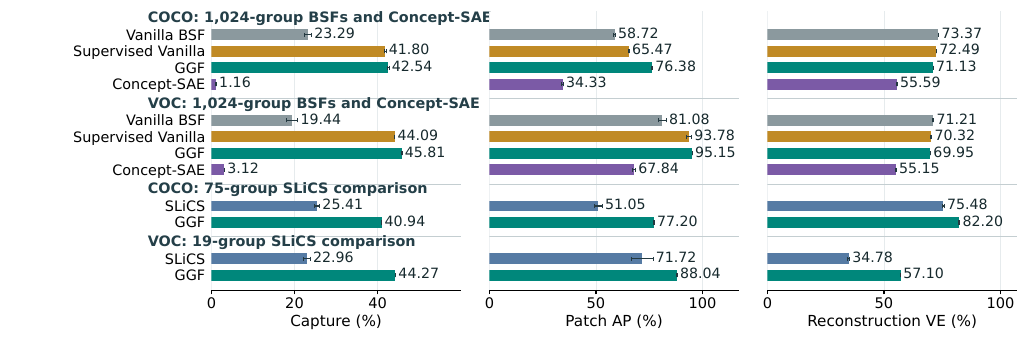}
\caption{Mean $\pm$ sample SD. Upper blocks compare 1,024-group BSFs
and Concept-SAE. Lower blocks compare GGF and SLiCS at 75 COCO and
19 VOC groups. Each block retains its own protocol and cohort
(Appendices~\ref{app:current-supervised-control}--\ref{app:voc-results}).}
\label{fig:released-main}\label{fig:main-results}\end{figure*}

\subsection{Matched object-supervised Vanilla comparison}\label{app:current-supervised-control}
Run \texttt{ef2f97f3dc53} compares reconstruction-only Vanilla,
fixed-weight supervised Vanilla and GGF at 1,024 groups, twelve directions
and Top-16. All endpoints receive 7,640 reconstruction warm-up and 3,000
continuation updates at Adam learning rate 0.0003. The random-initialized
arms share their warm-up. GGF uses mask-derived initialization and cap
0.40. Training-selected contrast/detection readouts remain fixed. The
same normalization, image cohorts and within-run sampling rules apply.
Reused checkpoints are verified by source and weight identity.

On COCO, fixed-weight calibration compares $(0.05,0.01)$, $(0.20,0.20)$ and
$(0.40,0.40)$ at the full continuation budget on the selection start.
Development Capture $+0.2\,\mathrm{AP}$ under the original warm-up MSE
allowance selects $(0.20,0.20)$, which satisfies that allowance. The chosen
pair applies to the remaining starts. Reconstruction weight is one, and
supervised Vanilla uses the same logistic classifier heads and both object
objectives as GGF. Each category has one head with a learned slope and
intercept on the selected group norm. It uses neither mask initialization
nor adaptive weighting. Relative to Vanilla, this control adds annotation
targets, contrast/detection objectives and trainable heads together. Their
individual effects are not identified by this comparison alone.

\begin{table}[htbp]\centering\small
\caption{Matched 1,024-group COCO comparison on the inspected audit cohort:
mean $\pm$ sample SD (\%). Capacity, activity and updates match.}
\label{tab:current-supervised-control}
\setlength{\tabcolsep}{3pt}
\begin{tabular}{@{}lrrr@{}}\toprule
Method & Capture & Patch AP & VE\\\midrule
Vanilla & $23.29\pm0.84$ & $58.72\pm0.32$ & $73.37\pm0.16$\\
Supervised & $41.80\pm0.24$ & $65.47\pm0.22$ & $72.49\pm0.10$\\
GGF & $42.54\pm0.21$ & $76.38\pm0.24$ & $71.13\pm0.03$\\\bottomrule
\end{tabular}\end{table}

GGF-minus-supervised-Vanilla gains are $0.73\pm0.32$ Capture,
$10.91\pm0.08$ AP and $-1.35\pm0.11$ VE points, with paired 95\%
training-start intervals $[-0.05,1.52]$, $[10.70,11.11]$ and
$[-1.62,-1.08]$. Compared with reconstruction-only Vanilla, gains are
$19.25\pm1.02$, $17.66\pm0.11$ and $-2.24\pm0.17$ points, respectively.
The supervised control isolates GGF's initialization/controller combination
under this calibration rule. The hyperparameter searches have different costs.
All per-model, category and paired-image records, frozen identities and
verified archive hashes are retained in \path{evidence/current_control/}.

\begin{center}\begin{minipage}{\linewidth}\centering\small
\setlength{\tabcolsep}{3pt}
\captionof{table}{Matched BSF control design.}
\label{tab:pending-control-design}
\begin{tabular}{@{}>{\raggedright\arraybackslash}p{.33\linewidth}>{\raggedright\arraybackslash}p{.24\linewidth}>{\raggedright\arraybackslash}p{.35\linewidth}@{}}\toprule
Arm & Initialization & Continuation\\\midrule
Vanilla BSF & Random & Reconstruction only\\
Supervised Vanilla & Random & All losses, fixed weights\\
GGF & Mask-derived & All losses, adaptive weights\\\bottomrule
\end{tabular}\end{minipage}\end{center}

\subsection{Supervised comparison with SLiCS}\label{app:supervised-five}
We compare GGF and SLiCS on both COCO and VOC, retaining one dictionary
per eligible category. The COCO study uses 75 groups and the VOC study
uses 19. Both use twelve directions per group and 10,640 training updates.
The dataset-specific protocols follow.

\paragraph{COCO data and frozen settings.}
SLiCS~\citep{li2025} applies alternating nonnegative-code/dictionary updates
to DINOv3 patches, adapting the published image-level inputs to patch-level evidence. Both methods share 2,000 training and 400 development
images, normalization, masks and 75 category slots. The 1,500-image
test excludes prior study splits. Settings, hashes and readouts freeze
before extraction. Test labels select no rate, checkpoint or group.
Run \texttt{a9c638e1ec0b} retains the complete audit.

\paragraph{Settings and their rationale.}
SLiCS assigns each concept a dictionary \citep[Sec.~3.1]{li2025}, giving
75 groups for our eligible categories. Their Sec.~4.2 relates width to
capacity and concept overlap. Twelve atoms match GGF's decoder, rather
than a published optimum. The 10,640 updates match our budget, replacing
their ten-iteration image-level schedule (Sec.~4.4). Spatial masks provide patch-level label support.

\paragraph{Matched quantities and remaining differences.}
Both methods share 900 decoder atoms and training-mask SVD initialization
\citep[Sec.~3.3]{li2025}, with SLiCS signs oriented by positive projection energy.
Training orders vary conditional on this initializer and fixed data.
GGF uses 7,640 reconstruction plus 3,000 guided updates, Top-32 and cap
0.60. Development-selected learning rate $10^{-3}$ comes from
$\{10^{-4},3\times10^{-4},10^{-3}\}$ and freezes for orders 24--25.
SLiCS receives 10,640 alternating updates on matching image schedules.
Mask support constrains training, not inference. The order budget freezes
before test access.

GGF has 1,383,452 learned parameters, including its untied encoder,
150 head parameters and two controller weights. SLiCS has 691,200
dictionary parameters. GGF activates 32 groups per patch, versus
approximately 75 for joint SLiCS reconstruction. Thus decoder capacity
and updates match, while total parameters, sparsity and computation differ.

\paragraph{Object readouts and full reconstruction.}
GGF uses its gated encoder codes. SLiCS object scores use an independent
nonnegative least-squares (NNLS) projection onto each category's twelve
directions, preserving its partial-concept readout \citep[Eq.~(3)]{li2025}. Both methods share contrast definitions, 32-patch detection samples, masks
and category denominators. Reconstruction VE instead
uses a joint NNLS solution over all 900 SLiCS directions: independent
concept projections are never summed as if they formed one reconstruction.
All 294,000 test patches are evaluated per model.

\paragraph{Solver verification and execution.}
A per-patch float64 projected-gradient KKT check enforces the $10^{-5}$
threshold against the full objective. Batched GPU FISTA warm starts
remove certified rows before cached float64 active-set refinement.
Excluded atoms can re-enter, with full SciPy NNLS up to 20,000 iterations
as fallback. An eight-patch probe checks objectives against independent
full NNLS before each dictionary's work, without retraining or selection.
Evaluation resumes verified CPU/GPU prefixes. Records retain precision
recovery and mixed execution. Nonunique codes may change activity at
equal reconstruction. Audit-call timers exclude deferred NNLS and are
not end-to-end inference times.

\paragraph{COCO training results.}
Table~\ref{tab:supervised-five-seeds} reports all final endpoints. GGF
improves Capture, AP and VE in every order. Training-order means are
40.94\%, 77.20\%, 82.20\% for GGF and 25.41\%, 51.05\%, 75.48\%
for SLiCS. Paired Student-$t$ intervals for GGF-minus-SLiCS are
$[14.78,16.28]$, $[23.63,28.66]$ and $[6.19,7.24]$ points, respectively.
These intervals condition on fixed data and initialization. Separately,
1,000 whole-image bootstrap draws per order retain all 75 categories.
Every Capture and AP interval excludes zero, conditional on the fitted
model rather than combining training-order and image uncertainty.

\begin{table*}[t]\centering\small
\caption{Values with $\pm$ show mean and sample SD. Completed COCO training on the separate supervised test cohort (\%).
All models have 75 groups of 12 directions and 10,640 updates.}
\label{tab:supervised-five-seeds}
\begin{tabular}{@{}llrrr@{}}\toprule
Seed & Method & Capture & Patch AP & VE\\\midrule
21 & GGF & 40.93 & 77.06 & 81.61 \\
21 & SLiCS & 26.25 & 52.26 & 75.33 \\
22 & GGF & 40.85 & 77.39 & 82.52 \\
22 & SLiCS & 24.51 & 49.69 & 75.49 \\
23 & GGF & 40.93 & 77.19 & 82.10 \\
23 & SLiCS & 25.26 & 53.56 & 75.40 \\
24 & GGF & 41.11 & 77.03 & 82.20 \\
24 & SLiCS & 25.82 & 51.02 & 74.96 \\
25 & GGF & 40.86 & 77.30 & 82.55 \\
25 & SLiCS & 25.22 & 48.72 & 76.22 \\
\midrule
Mean & GGF & $40.94\pm0.10$ & $77.20\pm0.15$ & $82.20\pm0.38$\\
Mean & SLiCS & $25.41\pm0.66$ & $51.05\pm1.94$ & $75.48\pm0.46$\\
\bottomrule\end{tabular}
\end{table*}

\paragraph{Development-selected contrast.}
The original co-primary contrast selects checkpoints by development
Capture $+0.2\,\mathrm{AP}$, including update zero. All GGF selectors
retain the final endpoint, while all SLiCS selectors retain the common SVD
initializer: 37.19\% Capture, 65.76\% AP and 67.87\% VE. Its zero
between-order SD reflects one identical checkpoint, not independent
initializations. GGF gains 3.75 Capture and 11.44 AP points, with paired
training-order intervals $[3.62,3.88]$ and $[11.25,11.63]$. Holm-adjusted
two-sided $p$ values are below $1.5\times10^{-7}$ for the two co-primary
object metrics. Every order's image interval also favors GGF. Those tests
retain their original primary designation. Final-versus-final comparisons
are separately reported secondary contrasts, not relabeled primary tests.

SLiCS training raises VE from 67.87\% to 75.48\% while reducing its
partial-projection object scores. Thus the larger final-endpoint object
gap partly reflects this trajectory. GGF's improvement over the selected
initializer is the more conservative object comparison. The result applies
to the stated patch-input implementation, not a claim about every supervised
method or a reproduction of published image-level retrieval results.

\paragraph{Capacity anchor and evidence.}
On this COCO cohort, saved 4,096-group GGF averages 42.69\% Capture,
78.45\% AP and 76.94\% VE. These images differ from the 1,500-image pilot cohort that gives
42.77\%/74.76\%/76.94\% in Appendix~\ref{app:pilotlatest}. This larger anchor is excluded
from the 75-group tests. Frozen identities, unrounded seed/category
records, intervals and provenance are retained in
\path{evidence/supervised_five_seed/}.

\paragraph{VOC transfer protocol.}\label{app:voc-slics}
VOC uses the splits in Appendix~\ref{app:voc-results}, fitting normalization
and mask-SVD initialization on training data. Its 19 category dictionaries
preserve SLiCS's design, giving 228 decoder directions per method. GGF
uses Top-8, cap 0.60 and COCO-selected learning rate $10^{-3}$ with the
same warm-up/continuation budget. Top-8 rounds the COCO active-group
fraction $19(32/75)$ before fitting. Both methods retain one fixed group
per category and a common training-mask SVD initializer, with SLiCS signs
oriented by positive projection energy. Unlike the denoising 1,024-group
study, this matched comparison uses clean reconstruction inputs and
excludes patches with crowd/void coverage above 0.1.
SLiCS receives 10,640 alternating updates. Seeds are 21--23 without a
new rate search. Final endpoints are the prespecified primary comparison.
The development-selected checkpoints remain separate diagnostic records. GGF has 350,476
learned parameters versus SLiCS's 175,104. SLiCS retains independent
category NNLS readouts and joint full-dictionary reconstruction, with
KKT tolerance $10^{-5}$, float64 recovery and a 20,000-iteration fallback.
Eight-patch checks for each SLiCS endpoint certify every row against
full CPU NNLS at the $10^{-5}$ KKT tolerance, with largest scaled
squared-error difference $3.48\times10^{-10}$. These category-matched
models are separate from the 1,024-group study.
VOC resource measurements use one device and remain separate from
COCO timing records (Appendix~\ref{app:compute-reporting}).

\paragraph{VOC results.}
The completed 19-group comparison gives GGF $44.27\pm0.09$\% Capture,
$88.04\pm0.11$\% Patch AP and $57.10\pm0.07$\% VE, versus
$22.96\pm0.85$\%, $71.72\pm5.07$\% and $34.78\pm0.44$\% for SLiCS
(Table~\ref{tab:voc-slics-final}). GGF improves every metric in every
paired run. Mean gains are 21.31, 16.33 and 22.32 points. Their paired
95\% Student-$t$ intervals are $[19.12,23.50]$, $[3.91,28.74]$ and
$[21.26,23.38]$ points (Table~\ref{tab:voc-slics-gains}). These intervals
describe variation across paired fits, without pooling models or images
from the COCO comparison. All 1,449 VOC validation images are evaluated,
including joint full-dictionary reconstruction for SLiCS. The frozen
run identity is \texttt{544561a40caa4025b92a}, with source lineage,
per-seed scores, solver checks and the completion receipt in
\path{evidence/voc_slics_final/}.
\begin{table}[htbp]\centering\small
\setlength{\tabcolsep}{3pt}
\caption{VOC category-matched comparison (\%). Both methods use 19 twelve-direction groups and 10,640 updates. Aggregates show mean $\pm$ sample SD.}
\label{tab:voc-slics-final}
\begin{tabular}{@{}llrrr@{}}\toprule
Seed & Method & Capture & Patch AP & VE\\\midrule
21 & GGF & 44.21 & 88.09 & 57.08\\
21 & SLiCS & 22.41 & 77.18 & 34.27\\
22 & GGF & 44.23 & 87.92 & 57.04\\
22 & SLiCS & 23.94 & 67.15 & 35.04\\
23 & GGF & 44.37 & 88.12 & 57.18\\
23 & SLiCS & 22.54 & 70.82 & 35.02\\
\midrule
Mean & GGF & $44.27\pm0.09$ & $88.04\pm0.11$ & $57.10\pm0.07$\\
Mean & SLiCS & $22.96\pm0.85$ & $71.72\pm5.07$ & $34.78\pm0.44$\\
\bottomrule\end{tabular}
\end{table}

\begin{table}[htbp]\centering\small
\setlength{\tabcolsep}{3pt}
\caption{Paired VOC GGF-minus-SLiCS gains in percentage points. SD and Student-$t$ intervals describe paired training-seed differences.}
\label{tab:voc-slics-gains}
\begin{tabular}{@{}lrr@{}}\toprule
Metric & Gain $\pm$ SD & 95\% interval\\\midrule
Capture & $21.31\pm0.88$ & [19.12, 23.50]\\
Patch AP & $16.33\pm5.00$ & [3.91, 28.74]\\
VE & $22.32\pm0.43$ & [21.26, 23.38]\\
\bottomrule\end{tabular}
\end{table}

\subsection{Supervised comparison with Concept-SAE}\label{app:concept-comparison}
\paragraph{Concept-SAE settings and their rationale.}\label{app:concept-sae-protocol}
Concept-SAE retains presence/mask supervision and residual free modules
\citep[Secs.~3.2--3.4]{ding2026conceptsae}, substituting DINOv3 features and
dataset masks. One token per eligible category and 36 free tokens follow
Sec.~4.2. The $14\times14$ grid matches DINOv3. BSF capacity differs.

Section~4.2 specifies Adam, batch size 64 and 30/50/30 epochs, with
tenfold rate decays after 20 tokenizer epochs and 30 aggregator epochs. Aggregation freezes the
tokenizer. Free-module training freezes concept modules. Aggregator/free
rates are $10^{-3}$. Tokenizer selection includes the published $10^{-3}$
on COCO development, then freezes for other starts/VOC. Empty-mask MSE
and constant-AP checks are additional diagnostics.

The pinned Concept-SAE release
(\texttt{956d19b}) supplies linear score/mask outputs, merge-$L_1$ weight
0.01, aggregator KL weight 0.01/temperature 0.02 and aggregation-weight
$L_1$ weight one. The paper instead uses a sigmoid for the presence
score and a merge penalty of 0.1. Its mask output is linear. Training follows its longer schedule, not the release's ten epochs.

\paragraph{Completed comparisons.}
Completed Concept-SAE achieves $1.16\pm0.07$\% Capture, $34.33\pm0.33$\%
AP and $55.59\pm0.12$\% VE, versus GGF's $42.54\pm0.21$\%,
$76.38\pm0.24$\% and $71.13\pm0.03$\% on the common COCO cohort
(mean $\pm$ sample SD). This tests native-token outputs under the stated
adaptation, not Concept-SAE's original benchmark. VOC results appear
in Table~\ref{tab:voc-primary}. All Concept-SAE fits complete the
30/50/30-epoch schedule and pass the empty-mask/constant-output checks.
On VOC, tokenizer and aggregator selection retains their last epochs.
Schedule completion and these checks establish a functioning fit, not
proof of convergence.

\paragraph{Parameter efficiency.}
GGF has 18,886,808 learned parameters (18,886,656 featurizer, 150 heads,
two controller), versus Concept-SAE's 175,125,423. Excluding the frozen DINOv3 backbone
and evaluation probes but counting all learned stages, GGF uses 10.8\%
as many parameters ($9.3\times$ fewer) while improving all COCO scores.
On VOC, 19 categories reduce GGF's heads to 38 parameters: the total is
18,886,696 including two controller parameters. Concept-SAE has
92,000,983 parameters. GGF uses 20.5\% as many ($4.9\times$ fewer).
The concept/free-token architecture is retained, so this is measured
parameter efficiency under the stated protocols, not a capacity-matched
architecture comparison.

\paragraph{Concept-SAE Capture calculation.}\label{app:concept-sae-capture}
Let $P_c$ be category $c$'s biased projector. For image $i$, predicted mask
$\mu_{ic}\in\mathbb R^{196}$ and presence $s_{ic}\in\mathbb R$ yield
$v_{ic}=P_c(s_{ic}\mu_{ic})\in\mathbb R^{768}$.
Let $r^{\rm CS}_{ipc}$ denote its contribution at patch $p$ and
$r^{\rm free}_{ip}$ the summed free-token contribution. The bias-free
aggregator $A\in\mathbb R^{196\times|\mathcal C|}$ gives
\begin{equation}
r^{\rm CS}_{ipc}=A_{pc}v_{ic},\qquad
\widehat x_{ip}=\sum_c r^{\rm CS}_{ipc}+r^{\rm free}_{ip}.
\label{eq:concept-sae-contribution}
\end{equation}
For object $o$ in image $i(o)$, the same weights $w_{op}$ from
Eq.~\eqref{eq:contrast} give
\begin{equation}
\delta r^{\rm CS}_{oc}
=\sum_p w_{op}r^{\rm CS}_{i(o)pc}
=\left(\sum_p w_{op}A_{pc}\right)v_{i(o)c}.
\label{eq:concept-sae-contrast}
\end{equation}
Capture pools errors over eligible objects before averaging categories:
\begin{equation}
\begin{split}
C^{\rm CS}_c&=1-
\frac{\sum_{o\in\mathcal O_c}\|\delta x_o-\delta r^{\rm CS}_{oc}\|_2^2}
{\sum_{o\in\mathcal O_c}\|\delta x_o\|_2^2},\\
\overline C^{\rm CS}&=\frac1{|\mathcal C|}\sum_c C^{\rm CS}_c.
\end{split}
\label{eq:concept-sae-capture}
\end{equation}
Category-token assignment is fixed, with no probe/decoder refit,
rescaling or clipping. Free tokens enter full VE, not category Capture.
Patch AP ranks native mask-times-presence scores $s_{ic}\mu_{icp}$.

\subsection{PASCAL VOC2012 comparisons}\label{app:voc-results}
The 1,024-group VOC comparison trains featurizers with COCO-selected
GGF defaults and VOC-only normalization, initialization, readouts and
evaluation probes. The separate 19-group SLiCS comparison appears in
Appendix~\ref{app:voc-slics}.
The official segmentation training split supplies 1,208 fit and 256
development images. All 1,449 official validation images supply held-out
evaluation. Annotation-only eligibility retains 19 object categories.
Flickr metadata are checked against COCO and exact decoded-image/Flickr
duplicates across VOC splits. This is not an exhaustive perceptual
near-duplicate audit. No GGF setting is selected on VOC validation.
As on COCO, supervised Vanilla's fixed loss pair is selected on VOC development under
the original warm-up MSE rule and frozen thereafter. This tests transfer
of the training recipe to a new dataset, not zero-shot transfer of
COCO-trained weights or evaluation on the withheld official test server.

\begin{table}[htbp]\centering\small
\setlength{\tabcolsep}{3pt}
\caption{VOC2012 validation scores (\%, mean $\pm$ sample SD). Readouts
are fixed on VOC training data. All models use the same images and categories.}
\label{tab:voc-primary}
\begin{tabular}{@{}lrrr@{}}\toprule
Method & Capture & Patch AP & VE\\\midrule
Vanilla BSF & $19.44\pm1.36$ & $81.08\pm1.91$ & $71.21\pm0.18$\\
Supervised Vanilla & $44.09\pm0.05$ & $93.78\pm1.09$ & $70.32\pm0.20$\\
GGF & $45.81\pm0.10$ & $95.15\pm0.05$ & $69.95\pm0.12$\\
Concept-SAE & $3.12\pm0.07$ & $67.84\pm0.73$ & $55.15\pm0.16$\\
\bottomrule\end{tabular}\end{table}

Against completed Concept-SAE, GGF gains 42.69 Capture, 27.31 AP and
14.80 VE points. Paired training-run 95\% intervals are $[42.57,42.81]$,
$[25.43,29.19]$ and $[14.58,15.02]$ points. All metrics favor GGF in
each paired run. The methods retain different native output interfaces.
This comparison does not equate Concept-SAE tokens with linear BSF groups.

Table~\ref{tab:voc-primary} compares reconstruction-only training with
the added object losses and classifier heads, then with GGF's initialization
and adaptive weighting. The supervised training package raises Vanilla
Capture from 19.44\% to 44.09\%. GGF reaches
45.81\%. Against supervised Vanilla, GGF gains 1.72 Capture and 1.36 AP
points at a 0.37-point VE cost. Paired training-run 95\% intervals are
$[1.36,2.08]$, $[-1.41,4.14]$ and $[-0.78,0.04]$ points. Both object
metrics improve in every paired run, but only Capture's interval excludes
zero. Conditional image-bootstrap AP intervals favor GGF for the first
two fits and include zero for the last. These do not replace training-run
uncertainty. Category means favor GGF on Capture for 18 of 19 categories
and AP for 11 of 19. Couch has lower Capture. One GGF endpoint exceeds the
common development-MSE allowance and remains reported without relaxing
or filtering by that rule.

\paragraph{Shared implementation and cross-dataset AP.}
Both datasets use the same Supervised Vanilla training code, BSF
architecture, contrast and detection objectives, logistic classifier heads,
Adam learning rate 0.0003 and 7,640/3,000-update warm-up/continuation budget.
Supervised Vanilla's contrast/detection loss weights are selected separately
on each dataset's development split. Both independent searches select
$(0.20,0.20)$.
VOC fits fresh weights, normalization and readouts from its own training
images. Its 19 categories require 38 head parameters, compared with 150
for COCO's 75 categories. Equal update counts therefore do not imply equal
passes through the differently sized training sets.

Supervised Vanilla's 93.78\% VOC Patch AP and 65.47\% COCO Patch AP average
over different categories, images and positive/negative patch populations.
The mask formats and annotation coverage also differ. Even unsupervised
Vanilla rises from 58.72\% on COCO to 81.08\% on VOC, consistent with a
broader dataset/evaluation difference rather than a change to the supervised
method. Category composition, object scale, context and patch prevalence
can affect AP, but these comparisons do not isolate their contributions.
Having fewer categories does not itself increase per-category binary AP.

Group-subset curves, removal controls and response spectra for these same models are analyzed in Appendix~\ref{app:voc-group-organization}.

\subsection{Fresh 4,096-group comparison}\label{app:pilotlatest}
\begin{figure*}[!tp]
\centering\includegraphics[width=.94\textwidth]{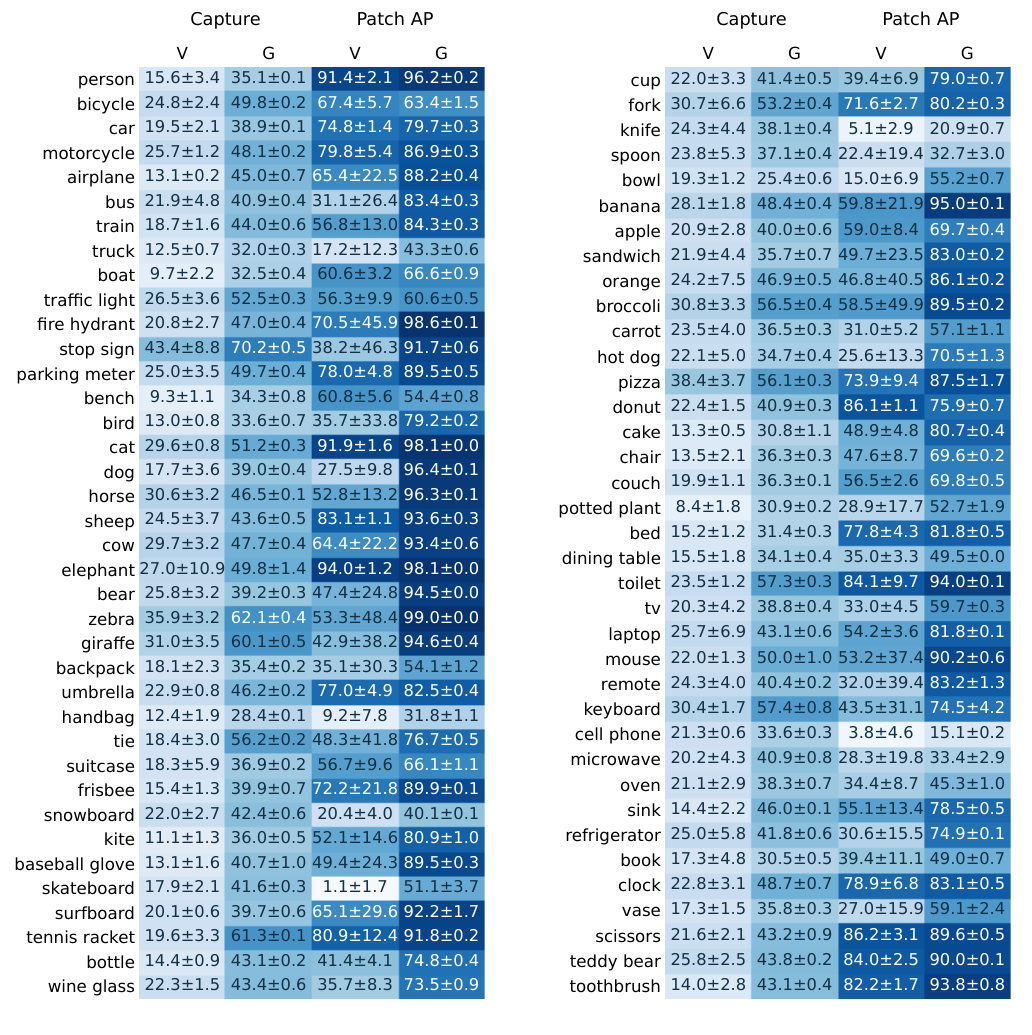}
\caption{From-scratch 4,096-group, 12D, Top-32 comparison on the inspected
1,500-image test cohort. Entries show per-category mean percentages and sample SD, comparing matched Vanilla (V) and GGF (G). Both receive 10,640
updates. GGF uses cap 0.60. Colors encode means on a shared 0--100 scale. Each entry includes sample SD. All 75 categories
are shown, including the three whose mean AP decreases. Capture and AP use
independently selected training readouts.}
\label{fig:current-categories}\label{tab:allcategories}
\label{fig:categories-2}\label{fig:categories-3}\label{fig:categories-4}
\label{fig:categories-5}\label{fig:categories-6}\label{fig:categories-7}\label{fig:categories-8}
\end{figure*}

This pilot uses newly initialized 4,096-group dictionaries with twelve
directions per group, Top-32 and adaptive cap 0.60. The first 75 GGF groups
use the twelve leading uncentered training-foreground directions. All
remaining groups start randomly. Vanilla is reconstruction-only and has no
semantic initialization. All six endpoints receive 7,640 reconstruction
warm-up and 3,000 additional updates. Constant-rate Adam at 0.0003 differs
from the OneCycle warm-up schedule. Paired branches share training streams.
The original feature cache, masks, normalization and 75 categories are reused.
The endpoints are fresh 12D fits, not expansions of trained 8D dictionaries.

\begin{table}[htbp]\centering\small
\caption{Completed pilot: development selection-set scores (\%). Seed
standard deviations are descriptive, not confidence intervals.}
\label{tab:pilotlatest}
\begin{tabular}{llrrr}\toprule
Seed & Model & Capture & AP & VE\\\midrule
21 & GGF & 42.48 & 80.46 & 77.29 \\
21 & Vanilla & 20.82 & 51.03 & 80.66 \\
22 & GGF & 42.28 & 80.54 & 76.69 \\
22 & Vanilla & 20.48 & 52.71 & 79.82 \\
23 & GGF & 42.24 & 80.33 & 76.49 \\
23 & Vanilla & 21.90 & 55.24 & 79.77 \\
\bottomrule\end{tabular}\end{table}

Mean GGF scores are 42.33\% Capture, 80.44\% AP and 76.82\% VE.
Vanilla scores 21.07\%, 52.99\% and 80.08\%. GGF seed standard
deviations are 0.13, 0.11 and 0.42 points. Vanilla's are 0.74, 2.12 and
0.50 points. Every seed favors guidance on object metrics and Vanilla
on reconstruction. The following evaluation scores these frozen endpoints
on the existing test cohort.

At the same 10,640-update budget, fresh 1,024/Top-16/cap-0.40 GGF is
available for seeds 21 and 22 only, averaging 42.19\% Capture, 82.41\% AP
and 70.90\% VE. On these common seeds the larger pilot averages 42.38\%,
80.50\% and 76.99\%. This changes group count, Top-$k$ and cap together,
so it does not isolate dictionary width. The missing third reference seed
prevents applying the original seed configuration-selection rule.
The larger dictionary has four times the core parameters: 75,546,624
versus 18,886,656 at 1,024 groups. Its 384 active coordinates also double
the Top-16 model's 192. Published 4,096-group experiments motivate this
candidate.

\paragraph{Test evaluation of the six saved models.}\label{app:pilot-test}
The evaluation-only run \texttt{445e2b6040fc} verifies all six model hashes
and TRAIN-selected readouts against the completed pilot before loading
test activations. It reuses the declared 1,500-image cohort, normalization, masks,
75 categories and deterministic 32-patch samples. This evaluation performs no gradient updates, readout
fitting or backbone extraction. Its scores do not use the separate
supervised-comparison holdout evaluated in Appendix~\ref{app:supervised-five}.

\begin{table*}[t]\centering\small
\caption{Values with $\pm$ show mean and sample SD. Fresh 4,096-group pilot on the existing test cohort (\%). All
models use 12D, Top-32 and 10,640 training updates. GGF cap is 0.60.}
\label{tab:pilot-test}
\begin{tabular}{llrrr}\toprule
Seed & Model & Capture & AP & VE\\\midrule
21 & GGF & 42.88 & 74.60 & 77.24\\
21 & Vanilla & 21.05 & 49.97 & 80.70\\
22 & GGF & 42.76 & 74.87 & 76.82\\
22 & Vanilla & 20.73 & 52.00 & 79.94\\
23 & GGF & 42.67 & 74.80 & 76.75\\
23 & Vanilla & 22.31 & 53.01 & 80.41\\\midrule
Mean & GGF & $42.77\pm0.10$ & $74.76\pm0.14$ & $76.94\pm0.27$\\
Mean & Vanilla & $21.37\pm0.84$ & $51.66\pm1.55$ & $80.35\pm0.39$\\
\bottomrule\end{tabular}\end{table*}

GGF's seed SDs are 0.10 Capture, 0.14 AP and 0.27 VE points.
Vanilla's are 0.84, 1.55 and 0.39. Each seed improves Capture and AP
while reducing VE. Category Capture improves in all 75 categories in
every run. Category AP improves in 72 of 75 categories after averaging
seeds, and in 64 categories in every run. These category counts describe
point estimates, not independent statistical replications.

\begin{table}[htbp]\centering\small
\caption{GGF-minus-Vanilla gains and paired 95\% whole-image intervals
on test, in percentage points. Each seed has 1,000 valid draws and zero
invalid draws. VE entries are point changes.}\label{tab:pilot-test-paired}
\begin{tabular}{lrrr}\toprule
Seed & Capture [interval] & AP [interval] & VE\\\midrule
21 & 21.82 [21.29,22.34] & 24.64 [23.09,26.23] & $-3.46$\\
22 & 22.03 [21.47,22.51] & 22.86 [21.40,24.52] & $-3.12$\\
23 & 20.35 [19.83,20.84] & 21.79 [20.31,23.34] & $-3.66$\\
\bottomrule\end{tabular}\end{table}

The mean paired gains are 21.40 Capture, 23.10 AP and $-3.41$ VE
points. Separate Student-$t$ 95\% intervals over the training starts
are [19.13,23.67], [19.52,26.67] and [$-4.10$,$-2.73$], respectively.
These imprecise seed intervals and the conditional image intervals
answer different uncertainty questions. Neither adjusts for multiple
comparisons or repeated inspection of this benchmark. All 75 categories
have finite scores in all six models. The original native-warm development
MSE allowance fails for all GGF endpoints. Reports retain this
failed allowance alongside all test scores.

Compared with the older expanded 1,024-group 12D GGF, the fresh pilot has
1.50 points lower Capture, 0.52 higher AP and 7.43 higher VE. Group count,
Top-$k$, cap, initialization history and training budget all differ, so
this is context rather than an isolated width comparison. The test scores
do not select a configuration. Full per-category scores, paired intervals,
model identities and source hashes are in \path{evidence/pilot4096_test/}.

\subsection{Matched native-baseline comparison}\label{app:matched-baselines}
The matched comparison reuses saved GGF models and trains the
remaining native controls. Every arm has 1,024 groups, three directions per group, and the same
2,000 training and 400 selection images. All models are scored on the same
1,500 test images, normalized with the original training statistics, across
the same 75 categories. Because this cohort was already inspected, the
comparison provides a shared evaluation rather than a new untouched
confirmation. No test labels enter the new training or readout selection.

\paragraph{Training budgets and reuse.}
Each warm-up performs 40 epochs, or 7,640 updates, with batches of 2,048
patches. The development-MSE minimum starts continuation. New Grassmannian
and Group Lasso warm-ups use the unchanged released classes. Saved Vanilla
and mask warm-ups are reused. Every arm has 2,500 plus 500 continuation updates, with Adam restarted
between phases. Saved Vanilla supplies the first 2,500 updates, and we train
its remaining 500-update phase alongside new continuation runs for
Grassmannian, Group Lasso, and mask reconstruction.
Mask GGF reuses the 3,000-update ordinary recovery endpoint. New work totals
74,340 gradient updates, with no new guided training.

Continuation uses eight sampled images per update, all 196 patches, learning
rate $0.0003$, scalar noise standard deviation $0.1\operatorname{std}(B)$,
gradient clipping at 5, and per-direction decoder normalization. The sampler
seeds are $71{,}000+s$ and $82{,}000+s$ for training seed $s$. Batch hashes
confirm identical image schedules across arms within each phase. Although the noise rule is shared, initialization consumes different random
draws across classes, so their noise arrays need not be identical.
Intermediate minimum-MSE checkpoints are retained as diagnostics. Primary comparisons use the final 3,000-update
endpoint in every arm, rather than each arm's best test score.

\paragraph{Readouts and activity.}
Each category's contrast and detector groups are chosen from training data
after warm-up and then fixed. Mask reconstruction and GGF share the
mask warm-up readouts. Native controls use their own warm-up readouts under
the same selection policy. Patch detection AP is scored from raw group norms rather than the learned
logistic classifier heads. Released Group Lasso retains its learned hard threshold and
mean-activity target of 16. Its measured mean is 17.99 on the test. This is
within the declared 20\% activity tolerance, but not identical Top-16 sparsity.
Table~\ref{tab:matched-resources} records the differences that remain.
\begin{agfresulttable}
\caption{Core parameter counts (millions), auxiliary head parameters, and active groups. Group Lasso activity is mean $\pm$ sample SD. Top-$k$ activity is fixed. All arms have 3,072 decoder directions. Update matching does not match computation.}\label{tab:matched-resources}
\begin{tabular}{@{}lrrr@{}}
\toprule
Model & Core (M) & Heads & Active \\
\midrule
Vanilla & 4.722 & 0 & 16.00 \\
Grassmannian & 2.360 & 0 & 16.00 \\
Group Lasso & 4.723 & 0 & $17.99\pm0.12$ \\
Mask + rec. & 4.722 & 0 & 16.00 \\
Mask GGF & 4.722 & 150 & 16.00 \\
\bottomrule
\end{tabular}

\end{agfresulttable}

\paragraph{Results and uncertainty.}
Table~\ref{tab:matched-seed-scores} reports all retained arms in each run. Capture
is recomputed from saved per-image contrast errors and target magnitudes.
All models share identical image IDs and contrast denominators. GGF
Capture, AP, and VE reproduce their previous reports exactly. Artifact hashes and checkpoint-recovery records pass validation.
\begin{agfresulttable}
\caption{Common-test results for every run and arm. All entries are percentages. Reconstruction and object-alignment metrics measure different properties.}\label{tab:matched-seed-scores}
\begin{tabular}{@{}llrrr@{}}
\toprule
Seed & Model & Capture & AP & VE \\
\midrule
21 & Vanilla & 21.178 & 58.722 & 65.196 \\
21 & Grassmannian & 16.437 & 60.261 & 64.973 \\
21 & Group Lasso & 20.764 & 56.757 & 64.564 \\
21 & Mask + rec. & 26.483 & 68.025 & 64.934 \\
21 & Mask GGF & 32.729 & 71.552 & 64.789 \\
22 & Vanilla & 20.393 & 60.640 & 65.293 \\
22 & Grassmannian & 16.892 & 60.942 & 64.962 \\
22 & Group Lasso & 21.510 & 55.919 & 64.489 \\
22 & Mask + rec. & 26.522 & 69.697 & 65.003 \\
22 & Mask GGF & 33.095 & 72.473 & 64.852 \\
23 & Vanilla & 21.241 & 58.811 & 65.160 \\
23 & Grassmannian & 16.223 & 61.787 & 65.018 \\
23 & Group Lasso & 21.388 & 53.265 & 64.294 \\
23 & Mask + rec. & 26.819 & 69.068 & 64.907 \\
23 & Mask GGF & 33.100 & 71.504 & 64.766 \\
\bottomrule
\end{tabular}

\end{agfresulttable}

Paired whole-image bootstraps keep each image's patches and objects together.
All 200 draws in each comparison retain the full category set. Their interpretation follows Appendix~\ref{app:secondary}. GGF has positive
Capture and AP intervals against every native control in every run.
\begin{agfresulttable}
\caption{Initialization and guided-loss increments, in percentage points. Init. compares mask reconstruction with Vanilla. Guided compares GGF with mask reconstruction. AP intervals are positive in every row and are included in the accompanying CSV.}\label{tab:matched-components}
\begin{tabular}{@{}llrrr@{}}
\toprule
Seed & Stage & Capture & 95\% interval & AP \\
\midrule
21 & Init. & 5.31 & [4.90, 5.74] & 9.30 \\
21 & Guided & 6.25 & [6.08, 6.40] & 3.53 \\
22 & Init. & 6.13 & [5.75, 6.52] & 9.06 \\
22 & Guided & 6.57 & [6.40, 6.74] & 2.78 \\
23 & Init. & 5.58 & [5.20, 5.91] & 10.26 \\
23 & Guided & 6.28 & [6.12, 6.43] & 2.44 \\
\bottomrule
\end{tabular}

\end{agfresulttable}

Mean gains attribute 5.671 Capture points and 9.539 AP points to mask
initialization plus reconstruction, followed by 6.367 Capture points and
2.913 AP points from guided losses. These controls separate initialization
from object-loss supervision, but do not establish equal total compute or
the optimal allocation of training between the stages.

\paragraph{Reconstruction eligibility.}
The native reference and its 2\% development-MSE allowance remain unchanged.
All Vanilla endpoints pass. Grassmannian passes in seeds 21 and 23. Group
Lasso, mask reconstruction and GGF fail in every run. This is determined
on development data, separately from test VE. Similar reconstruction scores
therefore do not imply compliance with the declared error limit.

Full per-category outcomes, paired comparisons, protocol, and
verification records accompany the paper under \path{evidence/matched_baselines/}.
Large activation and optimizer checkpoints stay in the recorded Drive
locations. The compact results archive retains source code and identities.

\subsection{The 256-group baseline screen}\label{app:screen256}
In the single-seed development screen, Vanilla had the highest native
Capture (21.31\%, versus 16.84\% for Grassmannian and 20.47\% for Group
Lasso). It also led on Capture and Patch AP after guided training.
Synthetic recovery and reconstruction supported this choice
(Table~\ref{tab:synthetic-means}). The 1,024-group comparisons show
metric tradeoffs: Grassmannian has higher Patch AP and Group Lasso
slightly higher Capture than Vanilla
(Table~\ref{tab:matched-seed-scores}). The supplementary code bundle records
the full screening settings and historical three-direction SLiCS pilots.
Finite-solver nonconvergence and differing checkpoint rules limit the
pilots, which are excluded from the twelve-direction comparison.

\section{Supervision and loss-weight studies}\label{app:supervision-overview}
These studies separate object supervision, mask initialization and task
weighting. The fixed-weight loss ablations select development
checkpoints, whereas the matched-component studies use scheduled endpoints.

\subsection{Twelve-direction component ablation}\label{app:components-current}
The completed component study (run \texttt{a2f215ed99c2}) uses 4,096 groups,
twelve directions, Top-32 and the declared training starts. Each endpoint receives 7,640
reconstruction warm-up and 3,000 continuation updates. We reuse the three
mask warm-ups and six completed Vanilla/GGF endpoints, adding only three
mask-initialized reconstruction-only fits and three fixed-guidance fits.
No new backbone extraction or unused holdout is required. All twelve
endpoints freeze before the new test audit on the existing 1,500 images.
Within each run the three mask-initialized arms share their starting
weights and fixed training-selected readouts. Capacity, activity, update
counts and image/noise streams match. FLOPs differ.

\paragraph{Training-matched fixed weights.}
Reconstruction keeps weight one. The fixed contrast and detection weights
are the arithmetic means of the saved adaptive weights at updates 500,
1,000, 1,500, 2,000, 2,500 and 3,000. They are respectively
$(0.36639,0.24090)$, $(0.35309,0.23918)$ and $(0.35819,0.23688)$ for
seeds 21, 22 and 23. Only training logs determine these constants, with
no development/test tuning. This matches the observed scale of guidance,
not the exact average over every update or an optimized fixed-weight
baseline. Adaptive GGF retains cap 0.60 and its saved controller schedule.

\begin{table*}[t]\centering\small
\caption{Values with $\pm$ show mean and sample SD. Component test results (\%). V: Vanilla, I: mask
initialization with reconstruction only, F: added fixed guidance, G: full
adaptive GGF. Every arm has the same capacity and update budget.}
\label{tab:components-seeds}
\begin{tabular}{@{}llrrr@{}}\toprule
Seed & Arm & Capture & AP & VE\\\midrule
21 & V &21.05&49.97&80.70\\
21 & I &23.61&68.73&79.80\\
21 & F &42.38&74.41&77.54\\
21 & G &42.88&74.60&77.24\\
22 & V &20.73&52.00&79.94\\
22 & I &23.53&68.38&79.33\\
22 & F &42.14&74.76&77.13\\
22 & G &42.76&74.87&76.82\\
23 & V &22.31&53.01&80.41\\
23 & I &23.53&68.53&79.54\\
23 & F &42.23&74.71&77.32\\
23 & G &42.67&74.80&76.75\\\midrule
Mean & V & $21.37\pm0.84$ & $51.66\pm1.55$ & $80.35\pm0.39$\\
Mean & I & $23.56\pm0.05$ & $68.55\pm0.17$ & $79.56\pm0.23$\\
Mean & F & $42.25\pm0.12$ & $74.63\pm0.19$ & $77.33\pm0.20$\\
Mean & G & $42.77\pm0.10$ & $74.76\pm0.14$ & $76.94\pm0.27$\\\bottomrule
\end{tabular}\end{table*}

\paragraph{Sequential gains and uncertainty.}
Table~\ref{tab:components-main} summarizes the four arms.
Table~\ref{tab:components-gains} reports paired seed intervals, computed
with Student's $t$ and two degrees of freedom. Each of the nine per-seed
component comparisons also has 1,000 paired whole-image bootstrap draws.
All are valid. Initialization and fixed guidance have positive Capture
and AP image intervals in every run. Adaptive Capture intervals are
positive in every run, while AP intervals include zero in seeds 22 and 23.
Seed intervals assess variation across fitted starts. Image intervals
condition on one fitted pair and the inspected cohort. Neither includes
researcher-selection uncertainty.

\begin{table*}[t]\centering\small
\caption{Sequential component gains: mean $\pm$ sample SD in percentage points, with paired
seed 95\% $t$ intervals. These describe this ablation sequence, not
independent causal shares.}
\label{tab:components-gains}
\begin{tabular}{@{}lrrr@{}}\toprule
Change & Capture & Patch AP & Reconstruction VE\\\midrule
Add initialization & $2.19\pm0.85$ & $16.89\pm1.68$ & $-0.79\pm0.17$\\
95\% interval & $[0.08,4.31]$ & $[12.72,21.06]$ & $[-1.20,-0.38]$\\
Add fixed guidance & $18.69\pm0.08$ & $6.08\pm0.36$ & $-2.23\pm0.03$\\
95\% interval & $[18.49,18.89]$ & $[5.18,6.98]$ & $[-2.30,-2.15]$\\
Use adaptive weights & $0.52\pm0.09$ & $0.13\pm0.06$ & $-0.40\pm0.16$\\
95\% interval & $[0.29,0.75]$ & $[-0.02,0.28]$ & $[-0.78,-0.01]$\\
\bottomrule\end{tabular}\end{table*}

Initialization improves mean category AP in 67 of 75 categories and Capture
in 46. Adding fixed guidance improves mean Capture in all 75 and AP in 65.
Adaptive weighting improves mean Capture in 71 and AP in 49. These counts
use category means across seeds and are not significance tests. Fixed
guidance obtains 97.6\% of GGF's Capture gain and 99.4\% of its AP gain
over Vanilla along this sequence, while retaining 0.40 more VE points.
The controller adds a smaller Capture benefit. A reliable AP
advantage over the tested fixed constants is not established.

Vanilla and initialization-only endpoints pass the original development
MSE allowance in all seeds. Both guided arms fail it in every run. The
study measures the alignment/reconstruction tradeoff, without showing
that reduced reconstruction is necessary or causes the alignment gains.
Contrast and detection are jointly enabled here. Their individual effects
are examined separately in the three-direction loss ablation below. Complete frozen
lineage, training-derived weights, category scores, paired intervals and
verification records are retained in \mbox{\path{evidence/components_current/}}.

The supplementary code bundle contains the three-direction warm-up,
continuation, random-start diagnostics and full loss-weight sweeps.

\subsection{Group width and adaptive loss weights}
\label{app:widthbalance}
This five-arm cap-0.20 study tests group width and loss weighting. Run \texttt{e4aeabc90db7} is complete for the declared training starts.
It reuses frozen DINOv3 features, the original 75 categories, 2,000 training
images, 400 selection images and the inspected 1,500-image test.
All 15 endpoints were frozen before test scoring. The comparison measures
improvement on this common cohort, not independent confirmation.

\paragraph{Initialization, training and controls.}
All arms use 1,024 groups and Top-16 gating. Width three supplies 3,072
directions and at most 48 active coefficients. Width four supplies 4,096
directions and at most 64. The corresponding encoder/decoder parameter
counts are 4,721,664 and 6,295,552. Guided models add 150 classifier
parameters. Vanilla has none. The four-direction control therefore matches
dictionary capacity, activity budget and updates, rather than supervision
or total computation.

For each category, we accumulate the training foreground outer-product
matrix defined in Appendix~\ref{app:initdetails}. We preserve its saved
first three float32 directions byte-for-byte, verify agreement with the
recomputed leading eigenspace, and obtain a fourth eigen-direction. We
project that direction away from the saved three, unit-normalize it, and
orient its sign by training projection energy. No development or test masks
enter this calculation. The remaining 949 groups are random. The category
initialization is shared across seeds. Reserve directions and training
order vary.

The original three-direction fixed-weight endpoint is reused, and the
three-direction adaptive arm reuses its reconstruction warm-up. The two
four-direction masked arms share one new warm-up per seed. Vanilla uses
a random four-direction warm-up. Each warm-up runs 40 epochs, or 7,640
updates, with the original Adam/OneCycle recipe. Continuation starts from
minimum selection-MSE warm-up weights and makes 2,500 plus 500 updates
at learning rate $3\times10^{-4}$. Both phases evaluate every 250 updates.
At the boundary, model Adam resets, but weights and trained heads carry
forward. Adaptive-controller state also carries forward. Final 3,000-update
endpoints are used regardless of eligibility, avoiding selection of only
successful checkpoints.

Continuation samples eight images with replacement and uses the original
noise scale and gradient clipping. Within each phase, sampled-image
schedules match across arms. Width-four noise streams also match. Different
tensor widths do not have identical noise realizations. Category readouts
are selected on training features after warm-up and stay fixed. The two
four-direction masked arms share readouts. Other warm-ups can yield different
assignments. The new work consists of six warm-ups and twelve continuations,
or 81,840 model-gradient updates. No backbone extraction or saved-control retraining is needed.

\begin{figure}[t]
\centering\includegraphics[width=.80\linewidth]{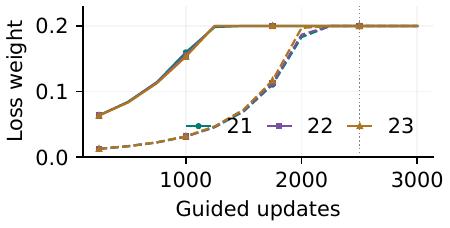}
\caption{Saved four-direction adaptive-weight trajectories. Lines connect
recorded values, not every controller step. Color identifies seed. Solid
and dashed lines identify contrast and detection. The dotted vertical line
marks the model-Adam reset at 2,500 updates. Controller state is retained.}
\label{fig:adaptive-weight-history}
\end{figure}

\paragraph{Observed weight trajectories.}
The cap is 0.20 in this study. Both object weights reach it in the four-direction runs (Figure~\ref{fig:adaptive-weight-history}), motivating
the larger-bound screen below.

\paragraph{Complete outcomes and uncertainty.}
Tables~\ref{tab:width-means} and~\ref{tab:width-seeds} report all arms.
Native eligibility uses each seed's original three-direction native
selection-MSE reference and its unchanged 2\% allowance, not test VE.
All four-direction endpoints in this cap-0.20 study pass. Both three-direction guided arms fail.
\begin{table*}[t]\centering\small
\caption{Values with $\pm$ show mean and sample SD. Five-arm test means (\%). Pass counts concern development MSE.}
\label{tab:width-means}
\begin{tabular}{@{}lrrrr@{}}\hline
Arm & Capture & AP & VE & Passes\\\hline
GGF 3 fixed & $32.97\pm0.21$ & $71.84\pm0.55$ & $64.80\pm0.04$ & 0/3\\
GGF 4 fixed & $33.89\pm0.10$ & $72.38\pm0.37$ & $66.36\pm0.03$ & 3/3\\
GGF 3 adaptive & $36.90\pm0.11$ & $74.37\pm0.80$ & $64.27\pm0.05$ & 0/3\\
GGF 4 adaptive & $37.80\pm0.07$ & $74.84\pm0.27$ & $65.76\pm0.02$ & 3/3\\
Vanilla BSF 4 & $21.31\pm0.49$ & $57.35\pm0.83$ & $66.74\pm0.05$ & 3/3\\\hline
\end{tabular}\end{table*}
\begin{agfresulttable}
\caption{All five endpoints in all seeds: test percentages.}
\label{tab:width-seeds}
\begin{tabular}{@{}rlrrr@{}}\hline
Seed & Arm & Capture & AP & VE\\\hline
21 & GGF 3 fixed &32.729&71.552&64.789\\
21 & GGF 4 fixed &33.786&72.263&66.317\\
21 & GGF 3 adaptive &36.804&74.278&64.251\\
21 & GGF 4 adaptive &37.727&74.879&65.740\\
21 & Vanilla BSF 4 &21.837&57.932&66.761\\\hline
22 & GGF 3 fixed &33.095&72.473&64.852\\
22 & GGF 4 fixed &33.989&72.086&66.370\\
22 & GGF 3 adaptive &36.874&75.213&64.325\\
22 & GGF 4 adaptive &37.868&74.543&65.773\\
22 & Vanilla BSF 4 &20.874&57.718&66.770\\\hline
23 & GGF 3 fixed &33.100&71.504&64.766\\
23 & GGF 4 fixed &33.890&72.796&66.379\\
23 & GGF 3 adaptive &37.025&73.628&64.238\\
23 & GGF 4 adaptive &37.795&75.086&65.781\\
23 & Vanilla BSF 4 &21.233&56.396&66.679\\\hline
\end{tabular}\end{agfresulttable}

Seven declared paired comparisons per seed give 21 comparisons, with 200
whole-image bootstrap draws and no multiplicity correction. Intervals follow Appendix~\ref{app:secondary}. VE differences are point estimates. Four-direction adaptive GGF improves
Capture, AP and VE over the original fixed-weight GGF in every run.
Its Capture/AP intervals are positive against original GGF, four-direction
fixed-weight GGF and four-direction Vanilla in all seeds. Its VE remains
below both four-direction controls. Against three-direction adaptive GGF,
AP decreases in seed 22, so width does not uniformly improve detection.
The research repository retains all 21 intervals in
\path{evidence/width_balance/paired_comparison.csv}, all endpoint scores,
and the category-level summary. The full experimental archive records
checkpoint hashes, batch identities, initialization verification and
controller histories. No new downstream-classifier evaluation is included.

Against matched four-direction Vanilla, adaptive GGF gains 16.48 Capture
and 17.49 Patch AP percentage points on average, with a 0.97-point VE cost.

\paragraph{Cap selection.}\label{app:higherbound}
On the 400-image development set, we compared caps 0.20, 0.40 and 0.60
with matched training updates for seeds 21--23. The 0.40 cap improved
Capture and Patch AP at a reconstruction cost. Raising it to 0.60
yielded little further gain, so 0.40 was selected before test scoring.
The supplementary code bundle contains the full development grid.

\paragraph{Test evaluation of the selected 0.40 cap.}\label{app:cap040test}
Run \texttt{eafd62bd8566} evaluates cap 0.40 for seeds 21, 22 and 23.
All saved final endpoints freeze before scoring the same cached
1,500 test images. We reuse the 0.20 GGF, four-direction Vanilla and
original GGF control reports and paired audit arrays, with no retraining,
test calibration or readout changes. The cap was chosen from development.
0.60, 0.80 and 1.00 are excluded. 

\begin{table*}[t]\centering\small
\caption{Values with $\pm$ show mean and sample SD. Means across runs (\%). All four-direction arms pass the
original native-reference development-MSE limit.}\label{tab:cap040test}
\begin{tabular}{@{}lrrr@{}}\hline
Model & Capture & Patch AP & VE\\\hline
GGF 4: cap 0.40 & $38.83\pm0.04$ & $75.19\pm0.25$ & $65.38\pm0.05$\\
GGF 4: cap 0.20 & $37.80\pm0.07$ & $74.84\pm0.27$ & $65.76\pm0.02$\\
Vanilla BSF 4 & $21.31\pm0.49$ & $57.35\pm0.83$ & $66.74\pm0.05$\\
Original GGF 3 & $32.97\pm0.21$ & $71.84\pm0.55$ & $64.80\pm0.04$\\\hline
\end{tabular}\end{table*}

Cap 0.40 improves Capture and AP over 0.20 in every run, with mean gains
of 1.03 and 0.35 percentage points. Mean VE falls by 0.38 points. Against
matched four-direction Vanilla, Capture and AP improve by 17.52 and
17.84 points, while VE is lower by 1.36 points. Against original GGF,
all three metrics improve, by 5.86 Capture, 3.35 AP and 0.58 VE points.
All nine paired Capture and AP intervals lie above zero. This confirms
an object-alignment gain on the shared test cohort, with a reconstruction
tradeoff. It does not show improvement in every category or independent
generalization. All original GGF controls fail the original native
development-MSE limit. Every four-direction endpoint in this cap-0.40 audit passes.

\begin{agfresulttable}
\caption{Cap-0.40 test scores by seed (\%).}\label{tab:cap040seed}
\begin{tabular}{@{}rrrr@{}}\hline
Seed & Capture & Patch AP & VE\\\hline
21 &38.823&75.217&65.331\\
22 &38.876&74.927&65.388\\
23 &38.796&75.423&65.423\\\hline
\end{tabular}\end{agfresulttable}

Each comparison uses 200 paired whole-image draws with shared image IDs
and the 75-category denominator (Appendix~\ref{app:secondary}).
The seed means, full category reports, endpoint identities and all
nine intervals are in \path{evidence/cap040_test/}.

\section{Capacity and continuation studies}\label{app:capacity-overview}
These studies vary group width or active-group count under matched
gradient-update budgets. Parameter counts and active-coordinate budgets
are also reported because equal updates do not imply equal computation.

\begin{figure*}[!tp]\centering
\includegraphics[width=\textwidth]{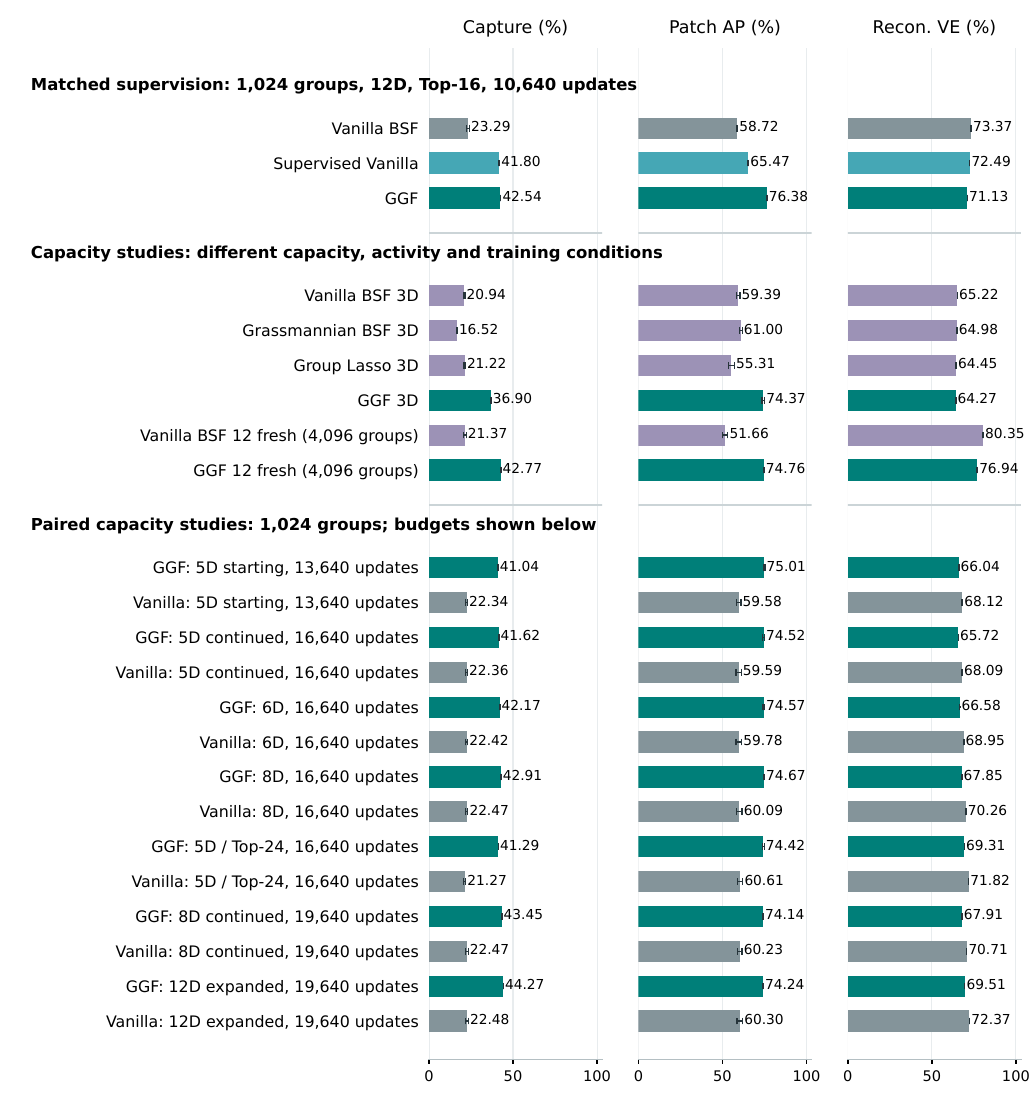}
\caption{COCO means with sample SD. Top: matched supervision
(Appendix~\ref{app:current-supervised-control}). Middle: comparisons at different capacities. Cross-row differences do not isolate width. Fresh 12D uses 4,096
groups, Top-32 and 10,640 updates. Bottom: every paired comparison in
Table~\ref{tab:capacitylatest}, with its update budget. Top-16 applies
except the stated Top-24 control. All bottom rows use 1,024 groups and
GGF cap 0.40. GGF caps are 0.20 (3D), 0.40 (matched), and
0.60 (fresh). Three-direction GGF and reconstruction-only rows come from
different studies. Group Lasso averages 17.99 active groups. All bars
start at zero. All GGF rows use adaptive weights.}
\label{fig:capacity-comparison-overview}\end{figure*}

The supplementary code bundle records the intermediate four- and
five-direction training and continuation studies.

\subsection{Matched capacity and continuation studies}\label{app:capacitylatest}
\paragraph{What changes and what remains fixed.}
The starting five-direction GGF and Vanilla checkpoints have 13,640
scheduled updates. The next comparison gives every arm another 3,000:
unchanged 5D/Top-16, expanded 6D/Top-16, expanded 8D/Top-16 and 5D/Top-24.
All use 1,024 groups and the declared training starts. GGF retains its cap-0.40 controller.
Vanilla trains reconstruction only. The final checkpoint supplies scores,
not the checkpoint with the highest test result.

Expansion preserves every learned encoder coefficient slot, bias entry and
decoder direction in its existing group. New seeded decoder directions
are projected away from the learned span and prior additions, then unit
normalized. Their encoder coefficients and biases start at zero, preserving
the starting gated code and reconstruction up to float32 rounding. Old Adam
moments, step counters, trained heads, controller state and reference losses
carry forward. New-coordinate moments start at zero and share the existing
Adam step counter. All directions may then fine-tune.

The 12D follow-up branches from completed 8D checkpoints at 16,640 updates.
One branch adds four coordinates and trains 3,000 updates. The other retains
8D and trains the same additional budget. Matched Vanilla branches follow
the same plan, giving twelve new endpoints at 19,640 updates. Training
uses eight images sampled with replacement, Adam at 0.0003, noise scale
0.1, gradient clipping at five and decoder unit normalization. Image and
noise stream hashes are checked within paired branches. Category Capture
and detector readouts stay fixed. Models freeze before scoring the same
inspected 1,500-image test cohort. This is a comparison on that
cohort, not an unused confirmation set.

\begin{table*}[t]\centering\small
\caption{Values with $\pm$ show mean and sample SD. Completed capacity studies: means (\%). All use
1,024 groups and cap 0.40 for GGF. Top-16 applies except the stated
Top-24 arm. Each Vanilla row matches its GGF capacity and update budget.
Starting checkpoints are included without retraining.}
\label{tab:capacitylatest}
\begin{tabular}{llrrrr}\toprule
Configuration & Model & Updates & Capture & Patch AP & VE\\\midrule
5D starting & GGF & 13,640 & $41.04\pm0.11$ & $75.01\pm0.26$ & $66.04\pm0.07$ \\
5D starting & Vanilla & 13,640 & $22.34\pm0.52$ & $59.58\pm1.40$ & $68.12\pm0.16$ \\
5D continued & GGF & 16,640 & $41.62\pm0.05$ & $74.52\pm0.47$ & $65.72\pm0.04$ \\
5D continued & Vanilla & 16,640 & $22.36\pm0.57$ & $59.59\pm1.58$ & $68.09\pm0.13$ \\
6D & GGF & 16,640 & $42.17\pm0.06$ & $74.57\pm0.45$ & $66.58\pm0.07$ \\
6D & Vanilla & 16,640 & $22.42\pm0.63$ & $59.78\pm1.64$ & $68.95\pm0.16$ \\
8D & GGF & 16,640 & $42.91\pm0.04$ & $74.67\pm0.39$ & $67.85\pm0.06$ \\
8D & Vanilla & 16,640 & $22.47\pm0.65$ & $60.09\pm1.61$ & $70.26\pm0.20$ \\
5D / Top-24 & GGF & 16,640 & $41.29\pm0.03$ & $74.42\pm0.41$ & $69.31\pm0.07$ \\
5D / Top-24 & Vanilla & 16,640 & $21.27\pm0.61$ & $60.61\pm1.53$ & $71.82\pm0.14$ \\
8D continued & GGF & 19,640 & $43.45\pm0.03$ & $74.14\pm0.23$ & $67.91\pm0.11$ \\
8D continued & Vanilla & 19,640 & $22.47\pm0.82$ & $60.23\pm1.47$ & $70.71\pm0.10$ \\
12D expanded & GGF & 19,640 & $44.27\pm0.03$ & $74.24\pm0.16$ & $69.51\pm0.12$ \\
12D expanded & Vanilla & 19,640 & $22.48\pm0.95$ & $60.30\pm1.71$ & $72.37\pm0.17$ \\
\bottomrule\end{tabular}\end{table*}

At 16,640 updates, 8D exceeds unchanged 5D by 1.29 Capture, 0.15 AP and
2.14 VE points. Top-24 at width five instead lowers Capture by 0.33 points
and AP by 0.10 points while raising VE by 3.59 points. This separates the
tested effects of within-group capacity and active-group budget. At 19,640
updates, 12D exceeds equally continued 8D by 0.82 Capture, 0.10 AP and 1.60
VE points. Every seed improves in all three point metrics. The 200-draw
paired whole-image bootstrap gives positive Capture intervals in every
seed. The AP interval is positive only in seed 22. These intervals condition
on fitted models and omit seed-level and researcher-selection uncertainty.

Twelve-direction models contain 18,886,656 core parameters versus
12,591,104 at width eight, and expose at most 192 versus 128 active
coefficients. GGF adds 150 training-only classifier parameters. Equal
updates therefore control training duration but not FLOPs or wall time.

\subsection{Width and activity sensitivity}\label{app:capacity}
\begin{figure*}[!tp]
\centering
\includegraphics[width=.94\textwidth]{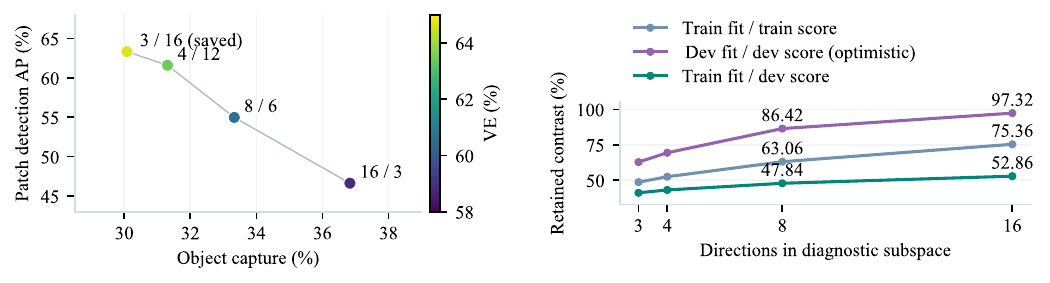}
\caption{(a) Matched-direction screen, selected development states. Same 3,072 total
directions and 48 active coordinates, with one start and reused development images. (b) Category-macro retained contrast energy (percent). The development-fitted
column is optimistic and must not be read as population recovery.}
\label{tab:wider}
\label{fig:width}
\label{tab:headroom}
\end{figure*}

The direction-width and activity screens use seed 21, 2,000 training images, the reused 400 development images, 75 categories and original loss weights. They do not open the final test. Their readouts are training-selected. The separate group-count check below uses mapping-selected readouts.

\paragraph{Total group count.}\label{app:groupcount}
The 1,024-group choice lies halfway on a logarithmic scale between the
released 256-group example and the paper's minimum DINOv3 sweep width,
4,096 \citep{goodfire2026,fel2026}. These anchors motivated the 1,024-group configuration.

A reconstruction-only Vanilla BSF check compares 1,024 and 4,096 groups at
width three and Top-16. Both use seed 21, the same 2,000 training images,
normalization, Adam/OneCycle recipe, and 40 epochs (7,640 updates). The saved
1,024 control was hash-verified after matching Torch, NumPy, and einops
versions. Minimum-selection-MSE checkpoints occur at 7,640 and 7,500 for
1,024 and 4,096 groups. A shared seed does not ensure identical batch/noise
streams across widths. Equal updates do not ensure equal computation.

A separate 400-image mapping split selects each model's contrast and detector
groups independently. Frozen readouts score all 75 categories on the same
400 confirmation/development images. These images were already
inspected. The final test remains unused. This follows our COCO schedule,
not the paper's full DINOv3 sweep protocol.

\begin{agfresulttable}
\caption{Follow-up Vanilla group-count check, seed 21, the confirmation/development cohort. Capture, Patch AP and VE are percentages. Parameters are millions.}\label{tab:groupcount}
\begin{tabular}{@{}lrr@{}}
\toprule
Metric & $G=1{,}024$ & $G=4{,}096$ \\
\midrule
Object contrast Capture & 21.32 & 20.43 \\
Patch detection AP & 61.99 & 62.18 \\
Reconstruction VE & 65.21 & 65.43 \\
Core parameters (M) & 4.72 & 18.89 \\
\bottomrule
\end{tabular}

\end{agfresulttable}

Increasing group count changes Capture by $-0.891$ percentage points
(paired 95\% interval $[-1.460,-0.360]$), Patch AP by $+0.193$ points
(interval $[-2.808,2.984]$), and VE by $+0.221$ points. All 100 image
resamples retain every category. Capture rises in 30 categories and falls
in 45. Mean AP improvement is inconclusive. These conditional intervals do
not establish a training-seed population effect. Lower Capture and fourfold
parameter cost support 1,024 under this budget, without identifying the best
GGF width or longer-training outcome. Records are in
\path{evidence/dictionary_size/} (results JSON and comparison CSV).

\paragraph{Fixed group count.}
The width/activity screen fixes 1,024 groups: width three gives 3,072 directions
for 768-dimensional activations, shared by matched controls. Here $b$ denotes
directions per group (main-text $r$), and $k$ denotes groups selected per
patch. At 1,024 groups, we test $b\in\{2,3,4\}$ and $k\in\{8,16,32\}$, reevaluate the saved
$b=3,k=16$ control, and give changed models 40 native epochs (7,640 updates)
plus 2,500 guided updates with shared image batches. Width changes capacity.
Activity changes selected coordinates. Equal updates do not match computation.
The group-count check is a separate experiment.
\begin{table*}[t]
\centering
\small
\setlength{\tabcolsep}{3pt}
\renewcommand{\arraystretch}{1.05}
\caption{The fixed-1,024-group screen reports selected development checkpoints. Width is the number of directions per group, and active groups is the Top-$k$ setting. The active coordinates column gives the number of retained scalar coefficients. Capture, Patch AP and VE are percentages. The first row reuses the saved control, and update zero denotes a retained native start.}
\label{tab:capacity}
\begin{tabular}{@{}rrrrrrrrr@{}}
\toprule
Width & \shortstack{Active\\groups} & Groups & Directions & \shortstack{Active\\coordinates} & \shortstack{Selected\\update} & Capture (\%) & Patch AP (\%) & VE (\%) \\
\midrule
3 & 16 & 1024 & 3072 & 48 & 2500 & 30.08 & 63.34 & 64.66 \\
2 & 8 & 1024 & 2048 & 16 & 2500 & 28.66 & 60.84 & 56.46 \\
2 & 16 & 1024 & 2048 & 32 & 2500 & 27.45 & 60.84 & 62.88 \\
2 & 32 & 1024 & 2048 & 64 & 500 & 21.14 & 59.29 & 69.65 \\
3 & 8 & 1024 & 3072 & 24 & 2500 & 30.52 & 59.15 & 57.92 \\
3 & 32 & 1024 & 3072 & 96 & 0 & 15.08 & 59.46 & 72.65 \\
4 & 8 & 1024 & 4096 & 32 & 2500 & 32.01 & 62.03 & 59.31 \\
4 & 16 & 1024 & 4096 & 64 & 750 & 28.20 & 64.16 & 66.35 \\
4 & 32 & 1024 & 4096 & 128 & 0 & 15.58 & 59.96 & 74.72 \\
\bottomrule
\end{tabular}

\end{table*}
Selection maximizes Capture $+0.2\,\mathrm{AP}$ within 2\% of each
native MSE. No setting meets all cross-setting replacement safeguards.
Table~\ref{tab:capacity-rejected} compares the $b=4,k=16$ endpoint, which
fails its native guard at seven audits, with the retained state. Models
$b=3,k=32$ and $b=4,k=32$ retain update zero despite higher endpoint Capture.
The CSV retains every selected and final state.

\paragraph{Fixed direction and activity budgets.}
This screen fixes 3,072 directions, 4,721,664 core parameters, and at most
48 selected coordinates. Models $b=4,k=12$, $b=8,k=6$, and $b=16,k=3$ each
receive new native initialization, 40 epochs, and 2,500 guided updates.
Wider groups reduce the group count, so these are new dictionaries rather
than continuations or rotations of $b=3,k=16$. Runtime and the number of
independent groups are not matched.

All three layouts pass their own native-MSE guards at every audit,
but lose AP and VE against original GGF. Table~\ref{tab:wider-states} gives
the $b=16,k=3$ outcomes. Native AP and VE are already low, so guided learning
alone did not cause the deficits. Capture improves in 69 of 75 categories
but AP in only 13. Late Capture gains persist, so this run does not establish a Capture plateau.

\begin{table*}[t]
\centering
\begin{minipage}[t]{.48\textwidth}
\centering\small\setlength{\tabcolsep}{3pt}
\caption{Width-four, Top-16 scores. Selection retains update 750.}\label{tab:capacity-rejected}
\begin{tabular}{@{}lrrr@{}}
\toprule
Model or state & Capture (\%) & AP (\%) & VE (\%) \\
\midrule
Selected: 750 & 28.20 & 64.16 & 66.35 \\
Final: 2,500 & 31.34 & 65.76 & 66.27 \\
\bottomrule
\end{tabular}
\end{minipage}
\hfill
\begin{minipage}[t]{.48\textwidth}
\centering\small\setlength{\tabcolsep}{3pt}
\caption{Width-16, Top-3 scores. NR means Capture was not reported.}\label{tab:wider-states}
\begin{tabular}{@{}lrrr@{}}
\toprule
Model or state & Capture (\%) & AP (\%) & VE (\%) \\
\midrule
Native & \textit{NR} & 45.54 & 59.45 \\
Selected: 2,250 & 36.83 & 46.63 & 58.86 \\
Final: 2,500 & 36.88 & 46.34 & 58.92 \\
\bottomrule
\end{tabular}
\end{minipage}
\end{table*}

\section{Response geometry and baseline checks}\label{app:geometry-overview}
\subsection{Coherence-linked group geometry}\label{app:coherence-pca}
Figure~\ref{fig:response-geometry} uses full-category, seed-21 models with
75 native groups each. The selected GGF glove group has three confirmed
specific pairs, versus one for each selected SLiCS and supervised Vanilla
group and zero for Vanilla. These examples were chosen to illustrate
coherence and failure cases before computing their PCA coordinates;
they are not representative group samples. All three original pair and
whole-group reviews remain frozen. The exhaustive comparison in
Table~\ref{tab:matched75-coherence} includes every group.

For each displayed group, we decode its native twelve-dimensional codes
into 768-dimensional contributions. BSFs retain their signed gated codes;
SLiCS retains its independent NNLS projection rather than a joint additive
reconstruction. The frozen checkpoints and normalization reproduce every
strongest marked patch index, with maximum relative mean-top-four response
error below $4.6\times10^{-7}$. No training or annotation changes are involved.

Each group's centered PCA uses its active patches from the same first
64 audit images in the original fixed order, chosen before projection.
For covariance eigenvalues $\lambda_j$ in descending order, the variance
retained by the first $k$ principal components is
$\sum_{j=1}^k\lambda_j/\sum_j\lambda_j$. Here $k$ counts PCA components,
separately from the encoder's group-activity setting. The first three
PCs retain 50.7\%, 50.2\%, 40.6\% and 35.3\% of reference variance for GGF, SLiCS, supervised Vanilla and Vanilla, respectively.
Clouds contain those reference patches; outlined points retain all four
original marks from each of the six reviewed images. Arrows locate the
strongest mark. Separate bases, signs and scales preclude interpreting
cross-panel coordinates as a shared space. No manifold surface is fitted;
these projections do not establish intrinsic dimension or topology.

PC1--3 set the RGB channels in both cloud and image overlays. Per-axis
limits start at reference percentiles 2 and 98, expand to include all
marked points, and map to $[0.04,0.96]$. Border, arrow and point-outline
colors instead annotate visible marked content from the frozen reviews;
they are not learned cluster assignments. Let $q_{99}$ be the 99th
percentile of that group's active reference norms. Opacity is
$0.61\min(\|z_g\|_2/q_{99},1)^{0.65}$. Activity-weighted colors use Gaussian smoothing
($\sigma=0.35$ patch) and clipped cubic interpolation. Insets are original
$96\times96$-pixel neighborhoods around the strongest $16\times16$ patch,
clamped to the original $224\times224$ crop. Floor projections are display
shadows. Saved arrays, image hashes, patch IDs and transforms reproduce
the figure without inferring new geometry or labels.

\subsection{Object-response visualizations}\label{app:heatmaps}
\begin{figure*}[t]\centering
\includegraphics[width=.95\textwidth]{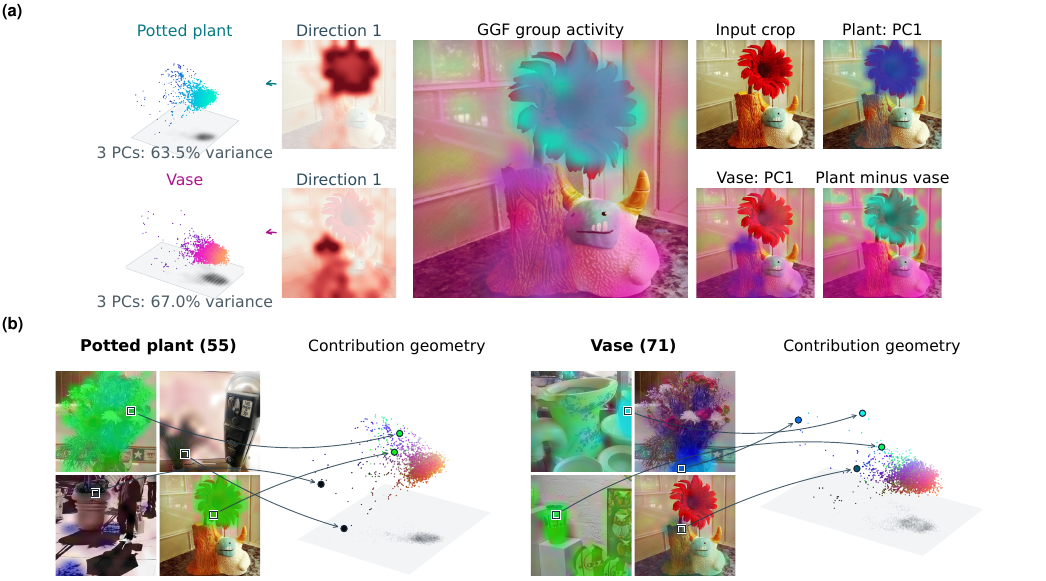}
\caption{\textbf{Object-group responses and geometry.}
(a) Plant (55) and vase (71) responses. Cloud/image colors encode PC1,
joint colors identify groups, and direction maps show coefficient magnitude.
(b) RGB encodes three contribution PCs with matching colors for patches
and clouds. Each arrow locates one marked patch. Both panels share the
GGF model and training-cloud samples. Overlays use activity-scaled opacity
and smooth interpolation.}
\label{fig:earlier-response-geometry}\end{figure*}

Figure~\ref{fig:earlier-response-geometry}(a) uses the selected plant/vase
photograph COCO 553000 with the fresh seed-23 GGF checkpoint: 4,096 groups,
12 directions per group, Top-32 and cap 0.60. The saved training detector
readouts assign potted plant to group 55 and vase to group 71. No
additional training is used for this visualization.

We collect gated coefficients on the first 256 images of the fixed training
cohort, retaining only patches where the displayed group is active. This
gives 13,911 plant-group responses and 5,455 vase-group responses. A fixed
uniform sample of 3,000 responses per group is plotted, without selecting
by masks, cluster shape or test performance. The same saved sample indices
are used across display revisions.

For each group, decoding its twelve coefficients produces a contribution
in the 768-dimensional activation space. We center these active training
contributions and fit PCA, whose orthogonal axes order their variation.
The first three axes retain 63.5\% of plant-group variance and 67.0\% of
vase-group variance. The remaining variance is absent from the 3D plot.
Neither its shape nor these percentages establish intrinsic concept dimension.
Point coordinates are the measured projected responses. Camera framing
and a density shadow change their presentation without changing coordinates.

In panel (a), the first principal coordinate determines color within each group:
blue--cyan--teal for plant and purple--magenta--orange for vase. The same
training-derived color function is applied to cloud points and image
patches. Its limits are the 2nd and 98th percentiles of that group's active
training coordinates. Values outside this range saturate for display.
The large joint-presence panel instead uses fixed cyan for plant and
magenta for vase. These categorical colors identify groups rather than
principal coordinates.

Overlay opacity depends on the full twelve-coefficient norm, scaled by
the 99th percentile of nonzero activity in the displayed image separately
for each group. This makes weaker responses visible but prevents comparing
absolute strength by brightness across groups. The difference panel
subtracts these normalized group activities. Red direction panels show
the absolute first native coefficient using separate within-group scales.
They do not depict the first PCA axis. Responses can include surrounding
objects and context, so these overlays are not segmentation predictions.

Smooth overlays apply a Gaussian filter with standard deviation
$\sigma=0.35$ patch to scalar responses, then cubic interpolation at patch
centers, clipped to the measured range. These display operations change no
scores, model outputs or original $14\times14$ patch values. The source
photograph is rendered at high resolution over the same model crop. No
image detail is generated. Checkpoint hashes, cohort IDs, projected arrays,
sampling indices and display scales accompany the figure.

Panel (b) uses the same saved PCA and training-cloud sample, but maps
PC1, PC2 and PC3 to red, green and blue. Each channel is scaled between
the 2nd and 98th percentiles of that group's active training coordinates,
clipped and mapped to $[0.04,0.96]$. Exactly the same function colors
image patches and cloud points. Four photographs per group illustrate
the responses. Three come from high-response images in the fixed training
cohort, and the fourth is the existing photograph COCO 553000. These are
illustrative examples, rather than a comparison of methods.

Each photograph contains many patch contributions, not a single geometry
point. An outlined patch with the largest gated coefficient norm in each
image is linked to its exact measured PCA coordinates. The eight highlighted
points supplement the fixed cloud sample without changing its coordinates.
Let $q_{99}$ be the 99th percentile of active training norms for that
group, shared across its four images. Opacity is
$0.76\min(\|z_g\|_2/q_{99},1)^{0.65}$. Colors are interpolated after weighting by activity.
The source photographs use the same resize and center crop as the model
inputs, rendered from their original pixels at higher display resolution.
The figure sources retain image IDs, file hashes, patch indices, PCA
coordinates and all display scales.

\subsection{Synthetic factor recovery with GGF}\label{app:manifoldzoo}
\begin{figure*}[!tp]
\centering\includegraphics[width=\textwidth]{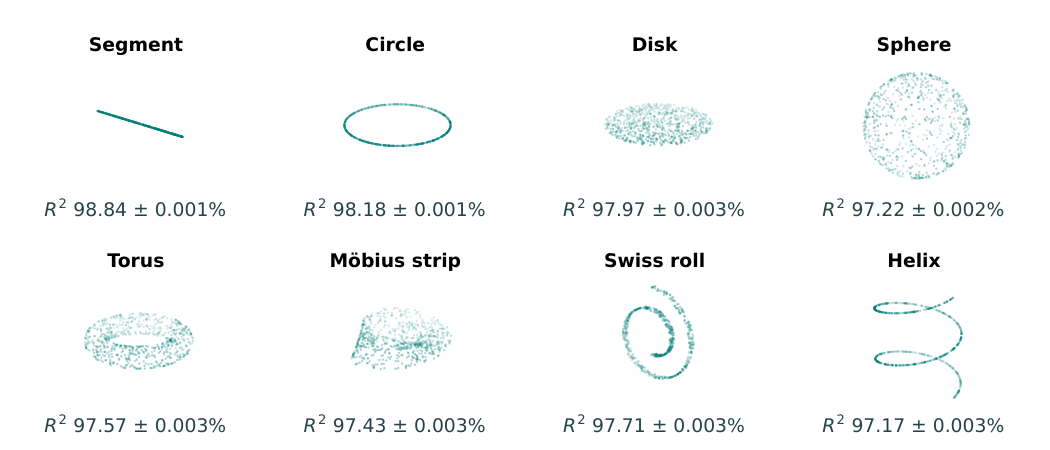}
\caption{Known factor families and GGF's recovery by geometry. Shapes show
samples from the benchmark generating functions of \citet{fel2026},
rendered by us and centered and scaled separately. Values show GGF's
assigned-group factor $R^2$, averaged within family, with mean $\pm$ sample SD.
Decoded predictions appear in Figure~\ref{fig:zoo-reconstructions}. There are 64
segments, 10 circles and 9 factors in each remaining family. The benchmark
embeds these factors into 128 dimensions and sums four per observation.}
\label{fig:zoo-geometries}
\end{figure*}

Manifold Zoo provides a controlled test where individual generating factors
are known. We compare a factor-supervised analogue of GGF with the released BSFs. Table~\ref{tab:zoo-main} reports the common recovery
and reconstruction metrics. Figure~\ref{fig:zoo-geometries} separates the
eight factor families. This reconstructs the benchmark of \citet{fel2026}
under the sampling protocol below, rather than reproducing its unpublished
arrays or all author settings.

\begin{figure*}[!tp]
\centering\includegraphics[width=.98\textwidth]{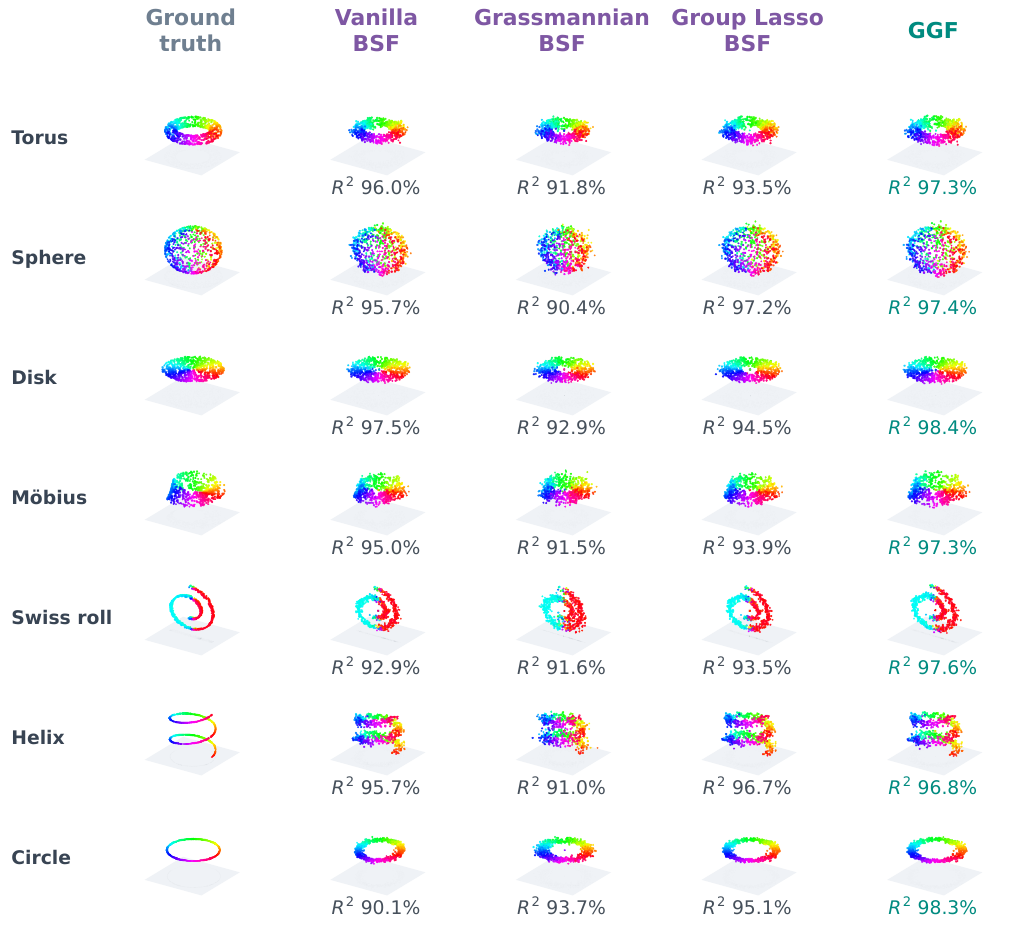}
\caption{Ground-truth factors and decoded single-group contributions on
Manifold Zoo \citep{fel2026}, using our saved seed-0 models. Rows show the
first indexed factor of each nonlinear family, selected without reference
to model scores, and the same 700 present-factor observations in every
column. Baseline groups are locked by calibration firing correlation.
GGF uses its assigned group. Each row shares the true factor's embedding
frame, center, scale and camera. Colors identify corresponding ground-truth
points. Gate-zero predictions are retained. These are model outputs on
regenerated benchmark observations. The
displayed $R^2$ uses all present-factor test observations and all 128
activation dimensions. Projections can hide off-factor error.}
\label{fig:zoo-reconstructions}
\end{figure*}

\begin{table*}[t]\centering\small
\caption{Mean $\pm$ sample SD. Seed synthetic means (\%). Test-matched recovery is a
secondary diagnostic. Primary recovery locks groups on calibration.
GGF final denotes the final endpoint of the selection-chosen trial.}
\label{tab:synthetic-means}\label{tab:zoo-guided-summary}\label{tab:zoo-main}
\begin{tabular}{@{}lrrr@{}}\toprule
Model & Locked $R^2$ & Test-matched $R^2$ & VE \\
\midrule
Vanilla & $96.781\pm0.533$ & $96.781\pm0.533$ & $97.372\pm0.043$ \\
Grassmannian & $86.782\pm0.776$ & $86.787\pm0.780$ & $92.049\pm0.183$ \\
Group Lasso & $93.295\pm1.548$ & $93.295\pm1.548$ & $96.694\pm0.131$ \\
GGF selected & $98.231\pm0.001$ & $98.231\pm0.001$ & $97.958\pm0.001$ \\
GGF final & $98.231\pm0.000$ & $98.231\pm0.000$ & $97.958\pm0.000$ \\
\bottomrule\end{tabular}
\end{table*}

\paragraph{Data and archived controls.}
Geometry seed 701 defines 128 factors in 128 dimensions: 64 segments and
64 curved factors cycling through circles, disks, spheres, tori, M\"obius
strips, Swiss rolls and helices. Each observation sums four independently
active, distinct factors. Training, selection, calibration and test contain
300,000, 20,000, 20,000 and 100,000 observations. Training-derived scaling
gives mean squared norm one without global centering. Summing the stored
true components reproduces the observations to $2.51\times10^{-8}$. This
checks the arithmetic ceiling, not learned recovery.

The saved baseline classes \citep{goodfire2026} match the source hashes
used by the new experiment after newline normalization. The original
dataset is reused with checksum-verified payload and exact array hashes
for all four splits. The nine selected baseline reports are imported
unchanged, with no new baseline fits. All models use 256 width-four
groups and requested activity four. Baseline seeds 0--2 and two learning
rates ($0.0004,0.003$) originally gave eighteen fits of 200 epochs
(29,200 updates), with batches of 2,048, Adam, OneCycleLR and no added
noise. Checkpoints minimize selection MSE. Group Lasso additionally
requires activity within 10\% of four. Its measured mean activity is
3.808 rather than exactly four.

\paragraph{Factor-supervised GGF.}
Six new fits use the same seeds, two learning rates and total update
budget. Initialization assigns one of the first 128 groups to each factor.
The leading four eigenvectors of the uncentered second moment of observed
training mixtures conditional on that factor's presence initialize its
decoder directions. The encoder begins at the decoder transpose. The
remaining 128 groups start randomly. Initialization uses presence labels,
not isolated component vectors. Factor-informed initial directions are
deterministic and shared across seeds. Reserve initialization and training
order vary. Reconstruction warm-up lasts 40 epochs. Guided learning
occupies the remaining 160. OneCycleLR spans all 200
epochs, without a reset at the transition.

For a training batch $\mathcal B$ of observations indexed by $n$,
let $S_n$ contain the four active factors, let
$\mathcal B_f=\{n\in\mathcal B:f\in S_n\}$, and let $\mathcal F_{\mathcal B}$
contain factors represented in the training batch. For the group assigned
to factor $f$, let $z_{nf}\in\mathbb R^r$ be its gated coefficient vector
and $D_f\in\mathbb R^{d\times r}$ its decoder, with $d=128$ and $r=4$.
These follow the column-vector convention of Equation~\eqref{eq:architecture}.
True component contributions $m_{nf}\in\mathbb R^d$ supply a synthetic
counterpart of the mask-contrast target. The supervised contribution loss is
\begin{equation}
L_{\rm factor}=\frac1{|\mathcal F_{\mathcal B}|}
\sum_{f\in\mathcal F_{\mathcal B}}
\frac{\sum_{n\in\mathcal B_f}\|D_fz_{nf}-m_{nf}\|_2^2}
{\sum_{n\in\mathcal B_f}\|m_{nf}\|_2^2}.
\label{eq:zoo-factor-loss}
\end{equation}
Each ratio normalizes the factor's mean squared contribution error by its
mean component energy, and factors receive equal weight. Balanced
factor-presence BCE uses the assigned group's gated coefficient norm,
training-only mean/scale measured after warm-up, and a learned positive
affine classifier. Its positive and negative pools receive equal weight
within each factor. There are 128 heads (256 additional parameters).
Reconstruction retains weight one. The same first-order gradient controller
adapts both supervised weights every ten guided updates, with exponent 1.5,
twenty reference batches and cap 0.60. This uses exact factor contributions
and presence, which are stronger targets than COCO masks. It is a synthetic
supervision experiment, not the unchanged COCO loss on another dataset.

Each seed's learning rate and checkpoint minimize selection reconstruction
MSE before model evaluation on synthetic test. All runs select learning
rate 0.0004, at updates 24,820, 24,090 and 27,010 for seeds 0, 1 and 2.
Every trial finishes its full budget. Table~\ref{tab:synthetic-means} also
reports the selected trial's final endpoint as a separate diagnostic.
The GGF core has 263,168 parameters, matching Vanilla's core. Supervision
adds heads and gradient computation. Archived BSFs use Torch's cu128
build and new GGF uses cu130, so the numerical environments are not identical.

\paragraph{Recovery and reconstruction.}

For factor $f$, let $m_{nf}$ be its true contribution and
$\widehat m_{nf}$ the chosen group's decoded contribution. With
$\overline m_f$ the true contribution averaged over observations where
the factor is present, recovery is
\begin{equation}
R_f^2=1-\frac{\sum_{n:f\in S_n}\|m_{nf}-\widehat m_{nf}\|_2^2}
{\sum_{n:f\in S_n}\|m_{nf}-\overline m_f\|_2^2}.
\label{eq:zoo-recovery}
\end{equation}
We average over all 128 factors. One denotes exact component recovery.
Zero matches the true factor's mean. No fitted readout or rescaling is
applied to predictions. For baselines, the primary comparison selects each
factor's group by correlation between its presence and group firing on
calibration observations, then freezes that assignment for test. GGF's
known assigned groups give identical scores to this calibration rule.
A paper-style diagnostic selects groups using test presence masks. Its
scores remain separate in Table~\ref{tab:synthetic-means}. Reconstruction
VE uses deviations from the training mean, unlike COCO's test-mean
denominator. Synthetic factor $R^2$ and COCO contrast Capture measure
different quantities.

\begin{agfresulttable}
\caption{Manifold Zoo by seed (\%). Recovery uses calibration-locked
groups. All GGF assigned-factor test scores exceed 90\%. The final row
counts those factors, out of 128.}\label{tab:synthetic-seeds}
\begin{tabular}{@{}lrrr@{}}\toprule
Model or metric & Seed 0 & Seed 1 & Seed 2 \\
\midrule
Vanilla recovery & 96.20 & 96.89 & 97.25 \\
Grassmannian recovery & 86.57 & 87.65 & 86.13 \\
Group Lasso recovery & 94.82 & 91.72 & 93.34 \\
GGF recovery & 98.231 & 98.232 & 98.232 \\
\midrule
Vanilla VE & 97.368 & 97.417 & 97.332 \\
Grassmannian VE & 92.001 & 92.252 & 91.895 \\
Group Lasso VE & 96.820 & 96.557 & 96.704 \\
GGF VE & 97.959 & 97.958 & 97.958 \\
GGF factors $R^2>0.90$ & 128 & 128 & 128 \\
\bottomrule\end{tabular}
\end{agfresulttable}

For Figure~\ref{fig:zoo-reconstructions}, we regenerate the fixed benchmark.
Seed-0 recovery and VE agree with archived values within
$3\times10^{-8}$ in absolute units. Full-dimensional scores accompany
common-frame projections.

GGF improves primary recovery by 1.450 percentage points and VE by 0.586
over Vanilla. Both improvements hold in every run. All 128 factors
exceed $R_f^2=0.90$ in their assigned group on test in each run. This is
100\% of factors passing a recovery threshold, not perfect reconstruction:
mean recovery is 98.231\%. GGF's seed SDs are 0.00057 recovery points and
0.00059 VE points. These describe training-start variation on one fixed
synthetic geometry, not uncertainty over generating dictionaries.
Mean factor-presence AP is 92.976\%. It is not COCO Patch AP, and archived
baseline AP is unavailable. The geometry breakdown ranges from 97.17\%
for helices to 98.84\% for segments. Neither this conditional synthetic
result nor the 93--97\% BSF recovery range establishes complete
natural-image concept recovery. Full records and source archives are
retained in \path{evidence/zoo_guided/}. The baseline JSON is in
\path{evidence/zoo_baselines/}.

\FloatBarrier
\section{Group organization and functional diagnostics}\label{app:revision-protocol}
This section brings the shared analysis protocol and its COCO and VOC2012 results together. The metrics distinguish useful individual contributions from stronger claims about semantic coherence.

\subsection{Response consistency and dependence on selected groups}\label{app:earlier-interpretability}
\label{sec:category-fingerprints}\label{sec:voc-results}\label{sec:controlled-robustness}
These diagnostics retain their original checkpoints, group budgets,
readouts and cohorts; they are not pooled with the 75-group study.
Figure~\ref{fig:category-fingerprints} combines category-response profiles
with tests of stability under appearance changes. Detailed protocols, category results, subgroup
Capture and selected/random removals follow below; whole-group focus
and coordinate-dependent direction agreement remain distinct.

Worst-condition recognition AP is $80.92\pm0.33$\% for GGF,
$55.16\pm3.37$\% for Concept-SAE and $80.52\pm1.22$\% for supervised
Vanilla. The GGF--Concept-SAE paired 95\% interval is positive; GGF's
means also exceed Vanilla and SLiCS. Subsequent paragraphs retain every
paired interval and the rotation/scale/brightness Capture scores.

\subsection{Readouts, subset curves and removal controls}\label{app:planned-consolidation}
\paragraph{Why these diagnostics.}
Capture tests reconstructed object contrast; AP tests spatial response
ranking. Their derivations and literature connections are given once in
Appendix~\ref{app:secondary}. The subset and removal tests below assess
complementary information and fixed-readout dependence.

\paragraph{Capture across group subsets.}
To test concentration without assuming a known manifold, a greedy rule
selects up to four groups on training object contrasts after fitting.
At each step it adds the native contribution giving the largest positive
training-error reduction. If none improves it, the unchanged prediction
is retained for larger budgets. This adapts compact-versus-distributed
capture analysis \citep{bhalla2026manifolds} to existing BSF groups.
For category $c$ and subset budget $q\in\{1,2,4\}$, let $S_{c,q}$ be
the selected group list and $C^{(q)}$ the resulting category-macro Capture.
These lists freeze before evaluation. Equation~\eqref{eq:auditcap}
is evaluated with $\delta r_{og_c}$ replaced by
$\sum_{g\in S_{c,q}}\delta r_{og}$, for $q\in\{1,2,4\}$.
No coefficient/decoder is refitted and encoder Top-$k$ is unchanged.
The increment $C^{(4)}-C^{(1)}$ measures additional native-contribution
recovery, not a feature-splitting rate: overlap, cancellation and context
also affect it. Small increments are informative only alongside substantial
one-group Capture. All-group Capture supplies reconstruction context.
The first selected group's norm is also scored for AP, ensuring the
same group is tested for reconstruction and localization.

\paragraph{Probe removal and its controls.}
Following probing \citep{alain2016probes} and feature ablation
\citep{gao2024scaling}, we fit an evaluation-only linear ridge probe to
training DINOv3 patches after freezing the featurizer. It predicts category mask fractions with penalty
$10^{-3}$ and an unpenalized intercept. One probe per dataset is shared
across BSF arms. It is separate from GGF's training logistic classifier heads and
changes neither featurizer training nor primary Capture/Patch AP.

For category $c$, let $\beta_c$ and $b_c$ be the fixed probe's weight
vector and intercept, so $f_c(x)=x^\top\beta_c+b_c$. Here $g_c$ is the
first training-greedy group. Keeping its code fixed, we subtract
$r_{g_c}(x)$ from $x$, preserving the reconstruction residual. The target effect is
\begin{equation}
\Delta\mathrm{AP}_c=\mathrm{AP}(f_c(x))-
\mathrm{AP}(f_c(x-r_{g_c}(x))).
\label{eq:probe-removal}
\end{equation}
For each category, we separately remove its first training-selected group.
This is not one handpicked group or an average over every dictionary group.
We macro-average category AP drops within each fitted model, then report
the mean and sample SD across fitted models. Four random groups are sampled from
the 32 nearest training activity/contribution-RMS neighbors of the
selected group. This is approximate matching, not exact energy matching.
The same label-blind 32-patch samples, crowd exclusions and positive/negative
mask thresholds as the primary AP evaluation are used. A larger drop
than random controls indicates that the selected contribution supports
the readout, not that information is absent from all remaining groups.
The probe is not refitted after removal. Because subtraction is measured
through a linear readout, its effects are readout dependence rather than
causal changes to the backbone's own computation.

Selectivity is checked separately through mean absolute changes in
off-target probe scores, not off-target AP. For each removed group,
we average $|r_g(x)^\top\beta_{c'}|$ over $c'\ne c$ and sampled patches,
then average categories. A large target AP drop does not imply a small
off-target change.

Active coefficient/contribution spectra use a common training cohort.
Direction, norm and PCA displays use training-derived bases and fixed image
rules. The BSF-family audit covers the primary arms on COCO and VOC.
Archived supporting-variant studies remain separate. Concept-SAE retains its native-output
evaluation rather than being relabeled as a linear block.

On the matched COCO models, GGF Capture rises from $42.54\pm0.21$\% with
one training-selected group to $52.14\pm0.30$\% with four. Vanilla rises
from $23.30\pm0.85$\% to $44.93\pm0.91$\%. Supervised Vanilla rises from
$41.80\pm0.24$\% to $50.81\pm0.37$\%. Values are means with sample SD.
Both supervised methods place more regional evidence in the first group,
while additional groups retain complementary evidence.
Removing the selected group reduces target probe AP by
$47.17\pm1.04$ points for GGF, $41.10\pm1.36$ for supervised Vanilla and
$6.32\pm1.32$ for Vanilla. Matched random removals average
$0.01\pm0.02$, $0.02\pm0.02$ and $0.02\pm0.01$ points, respectively.
The fixed probe is applied to original activations after subtracting
native group contributions. These are readout-based functional diagnostics,
not interventions on a backbone output.

\subsection{VOC2012 group organization}\label{app:voc-group-organization}
The model scores and dataset allocation are reported in Appendix~\ref{app:voc-results}. Here we examine how the same fitted models distribute object-associated information across groups.

\begin{figure*}[!tp]\centering
\includegraphics[width=\textwidth]{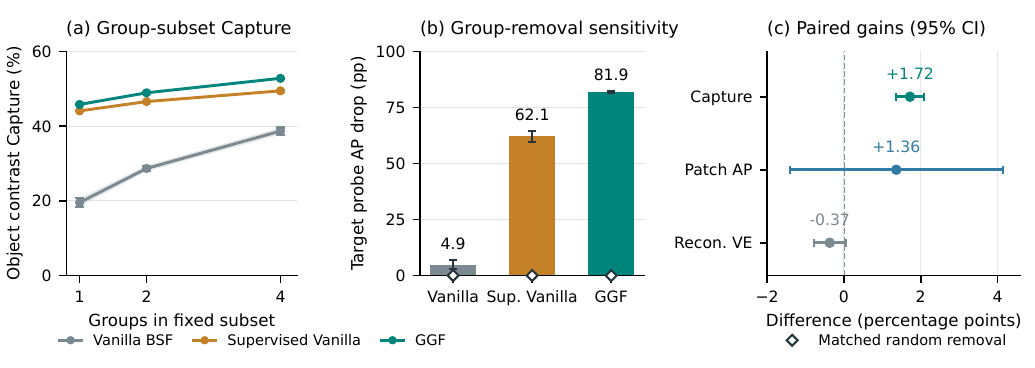}
\caption{VOC2012 group analyses. (a) Native contributions from
training-selected subsets. Lines and bands show mean $\pm$ sample SD.
The subset size changes, not encoder Top-$k$. (b) Target probe-AP drop
after subtracting one Capture-selected group. Bars show mean $\pm$
sample SD and white diamonds matched random removals. (c) GGF minus
supervised Vanilla, with paired training-run 95\% Student-$t$ intervals.
Higher Capture and removal sensitivity support useful individual groups.
Neither establishes semantic purity or removal of all complementary information.}
\label{fig:voc-consolidation}\end{figure*}

\begin{table*}[t]\centering\small
\setlength{\tabcolsep}{4pt}
\caption{VOC consolidation audit: mean $\pm$ sample SD. Capture (\%)
uses a post-fit training-greedy plan, separately from the primary readouts.
Removal effects are target-category probe-AP drops in percentage points.
The random column averages four activity/contribution-matched controls.}
\label{tab:voc-consolidation}
\begin{tabular}{@{}lrrrrrr@{}}\toprule
Method & One group & Two groups & Four groups & All groups & Selected removal & Random removal\\\midrule
Vanilla BSF & $19.57\pm1.15$ & $28.69\pm0.75$ & $38.70\pm0.95$ & $79.80\pm0.04$ & $4.94\pm1.93$ & $0.044\pm0.060$\\
Supervised Vanilla & $44.09\pm0.05$ & $46.58\pm0.22$ & $49.45\pm0.21$ & $79.63\pm0.07$ & $62.14\pm2.34$ & $0.031\pm0.050$\\
GGF & $45.81\pm0.10$ & $48.93\pm0.13$ & $52.81\pm0.10$ & $79.79\pm0.07$ & $81.93\pm0.56$ & $0.006\pm0.004$\\
\bottomrule\end{tabular}\end{table*}

\paragraph{What the subset curves establish.}
GGF preserves a Capture advantage at every measured subset size
(Figure~\ref{fig:voc-consolidation}a). Its one-to-four-group gain is 7.00
points, versus 5.36 for supervised Vanilla and 19.13 for Vanilla. The
supervised arms therefore capture much more contrast with one group and
need less additional contribution than unsupervised Vanilla. However,
GGF's larger marginal gain than supervised Vanilla does not establish
less fragmentation between those two methods. All-group Capture remains
near 80\% for every arm, so the first group is useful without containing
all object-related information. The Vanilla audit's one-group value
differs slightly from its primary score because the greedy plan selects
groups under its own post-fit training rule. No held-out image selects a
group, coefficient or subset.

\paragraph{What removal establishes.}
Removing one training-selected group separately for each of the 19
categories reduces target probe AP by $81.93\pm0.56$ points for GGF,
$62.14\pm2.34$ for supervised Vanilla and $4.94\pm1.93$ for Vanilla.
These are category-macro means with sample SD across fitted models.
Matched random removals average below 0.05 points
(Table~\ref{tab:voc-consolidation}). Thus GGF's selected contribution
supports the fixed object readout more strongly. This measures readout
dependence, not the absence of recoverable information in the remaining
groups: no replacement probe is fitted after removal. Larger drops also
do not establish greater selectivity. The off-target score-change audit
in Table~\ref{tab:voc-geometry} is larger for GGF. Because the frozen
probe is linear, removal directly subtracts the contribution's projection
onto its weights. It is not an intervention on DINOv3's downstream computation.

\paragraph{Within-group variation and semantic coherence.}
The training-active coefficient and decoded-contribution spectra measure
how many variance-bearing directions a selected group uses, rather than
whether those directions share a meaning. Table~\ref{tab:voc-geometry}
averages effective ranks over the Capture-selected group for each
category, then over training runs. GGF's contribution entropy rank is
higher than either control at the same width of twelve. This establishes
richer measured within-group variation, not higher semantic coherence or
recovery of a known natural-image manifold.

\begin{table}[htbp]\centering\small
\setlength{\tabcolsep}{3pt}
\caption{VOC selected-group variation and collateral readout effects:
mean $\pm$ sample SD of category means. Entropy rank uses training-active
patches. Off-target change is the mean absolute change in other probe
scores, multiplied by $10^3$, not an AP drop.}
\label{tab:voc-geometry}
\begin{tabular}{@{}lrrr@{}}\toprule
 & Coefficient & Contribution & Off-target\\
Method & entropy rank & entropy rank & change $\times10^3$\\\midrule
Vanilla BSF & $6.58\pm0.07$ & $4.99\pm0.15$ & $0.543\pm0.050$\\
Supervised Vanilla & $5.98\pm0.12$ & $4.21\pm0.11$ & $0.849\pm0.033$\\
GGF & $6.40\pm0.03$ & $6.28\pm0.03$ & $1.532\pm0.036$\\
\bottomrule\end{tabular}\end{table}

Signed-direction maps, norm overlays and contribution-PCA color maps
are generated for fixed annotation-positive person and potted-plant
images shared across methods. The image rule is independent of scores.
PCA bases and display limits use training data. This follows the
contribution-space visualization of Fel et al.~\citep[Sec.~3.5]{fel2026}:
rotating a block's coordinate basis can change individual coefficient
maps without changing its decoded contribution. PCA signs and display
scales also differ by model, so hue or brightness alone cannot rank
coherence. These displays make variation inspectable. The separate controlled-view
study below adds fixed transformations and qualitative inspection.

\subsection{Controlled appearance changes and qualitative coherence}
\label{app:controlled-robustness}
\begin{figure*}[t]\centering
\includegraphics[width=\textwidth]{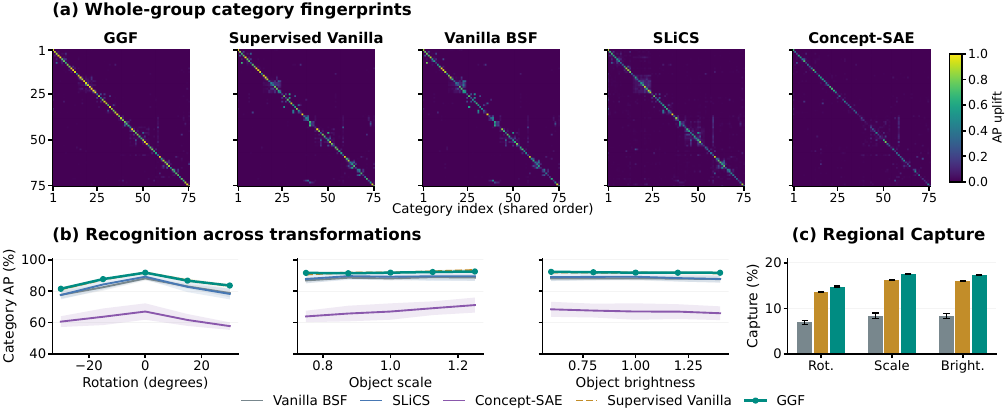}
\caption{\textbf{Object focus and robustness.}
Categories shown here were used for guidance.
(a) Rows index 75 fixed group or token readouts, and columns test the
same categories. Cells show mean above-prevalence AP uplift.
A stronger diagonal indicates greater object focus.
(b) Fixed readouts on six categories under controlled transformations.
(c) Single BSF groups preserve regional contrast. Bands and error bars
show mean $\pm$ sample SD. Protocols and SD maps appear in
Appendices~\ref{app:controlled-robustness}, \ref{app:category-fingerprints}.}
\label{fig:category-fingerprints}\label{fig:controlled-robustness}\end{figure*}

The preceding subset and removal tests assess information carried by a
group. Here we ask whether that group's object association persists
across changing appearances, then inspect the meaning of its strongest
responses. The models and readouts remain frozen throughout.

\paragraph{Cohort and transformations.}
The predeclared categories are person, car, dog, chair, potted plant and
vase. Annotation coverage and a stable image hash determine the cohort,
with at most 32 evaluation instances per category and one instance per
base image. The final pool contains 178 evaluation objects. Training
objects separately determine PCA bases and display scales. Existing
COCO audit images are reused, rather than introduced as a new test set.
Each annotated instance is extracted from the pinned crop, resized to a
144-pixel longest side and centered on a $224\times224$ neutral canvas.
We vary in-plane rotation over $\{-30,-15,0,15,30\}$ degrees, scale over
$\{0.75,0.875,1,1.125,1.25\}$ and object brightness over
$\{0.6,0.8,1,1.2,1.4\}$. Every object must pass annotation-only geometry
checks at every level. These are controlled two-dimensional image
changes, not three-dimensional pose or physical illumination.

\paragraph{Recognition with a fixed native readout.}
Each category-associated unit scores every object in the common
six-category pool. BSFs use their gated group norms. SLiCS and
Concept-SAE retain their native category response interfaces. Training
fixes BSF group selection, while the alternatives use their category
slots. AP ranks instances of the target category above the other five.
It is object-category recognition, distinct from Patch AP on natural
images. Let $\mathcal C$ be these six categories, $s$ index fitted models,
$a$ the transformation type and $\ell$ its level. Write $\mathrm{AP}_{cs}(a,\ell)$
for recognition AP, $A^{\min}_{cs}$ for its worst condition and
$\overline A^{\min}_s$ for the category mean:
\begin{equation}
A^{\min}_{cs}=\min_{a,\ell}\mathrm{AP}_{cs}(a,\ell),\qquad
\overline A^{\min}_s=|\mathcal C|^{-1}\sum_c A^{\min}_{cs},
\end{equation}
where $(a,\ell)$ ranges over the fifteen foreground conditions. Taking
the minimum within each category before averaging prevents strong
categories from hiding another category's weakest condition. The curve
in Figure~\ref{fig:controlled-robustness}(b) instead reports category-macro
AP at each level. Both summarize each fitted model before reporting mean
and sample SD across models. Table~\ref{tab:controlled-category-ap}
retains every category.

GGF's mean worst-condition AP exceeds Vanilla by 5.09 points, SLiCS by
4.29 and Concept-SAE by 25.76. Paired 95\% intervals are
$[-1.78,11.96]$, $[-2.90,11.47]$ and $[17.47,34.05]$ points,
respectively. Its 0.40-point difference from supervised Vanilla has
interval $[-2.32,3.11]$. The clearest recognition advantage is therefore
over Concept-SAE. The archived multiplicity-adjusted result for that
contrast is $p=0.0222$. The other contrasts do not establish a reliable
recognition advantage under this protocol.

\paragraph{Capture under the same changes.}
For the three BSF arms, Equation~\eqref{eq:auditcap} uses each category's
first frozen training-selected group. Squared errors and target energies
pool across objects and transformation levels within each category,
followed by equal category averaging. This tests whether an additive
group contribution retains regional object information, beyond merely
ranking objects correctly. The cutout foreground is compared with its
composited background, so these Capture values should not be equated
with the natural-image Capture in Figure~\ref{fig:released-main}.
Table~\ref{tab:controlled-category-capture} reports every category and
foreground transformation. Macro Capture for GGF is
$14.75\pm0.15$\%, $17.50\pm0.15$\% and $17.37\pm0.15$\% under rotation,
scale and brightness. Supervised Vanilla reaches $13.54\pm0.11$\%,
$16.13\pm0.09$\% and $16.04\pm0.10$\%, and Vanilla
$6.91\pm0.43$\%, $8.40\pm0.59$\% and $8.38\pm0.54$\%. SLiCS's independent partial projections and
Concept-SAE's native tokens are not added to this BSF-contribution audit.

\begin{table*}[t]\centering\small
\setlength{\tabcolsep}{5pt}
\caption{Worst-condition object-category AP (\%, mean $\pm$ sample SD).
For each category, the minimum AP across the fifteen foreground
conditions is computed within a fitted model before aggregation.}
\label{tab:controlled-category-ap}
\begin{tabular}{@{}lrrrrr@{}}\toprule
Category & Vanilla & Supervised Vanilla & SLiCS & Concept-SAE & GGF\\\midrule
Person & $91.72\pm0.73$ & $96.37\pm1.30$ & $92.91\pm2.85$ & $91.43\pm4.96$ & $98.00\pm0.13$\\
Car & $79.33\pm8.60$ & $81.82\pm3.52$ & $84.37\pm5.89$ & $55.59\pm4.31$ & $75.02\pm3.04$\\
Dog & $92.76\pm1.45$ & $93.20\pm1.52$ & $95.12\pm1.42$ & $91.55\pm1.84$ & $94.92\pm0.05$\\
Chair & $52.95\pm8.44$ & $58.48\pm4.66$ & $55.47\pm11.13$ & $16.63\pm4.30$ & $56.20\pm3.37$\\
Potted plant & $64.37\pm6.55$ & $81.94\pm0.80$ & $83.50\pm11.03$ & $43.66\pm9.83$ & $91.67\pm1.49$\\
Vase & $73.83\pm5.82$ & $71.30\pm8.46$ & $48.40\pm11.79$ & $32.07\pm3.51$ & $69.68\pm0.90$\\
\bottomrule\end{tabular}\end{table*}

\begin{table*}[t]\centering\small
\setlength{\tabcolsep}{6pt}
\caption{Controlled single-group Capture by category and transformation
(\%, mean $\pm$ sample SD). Squared errors and target energies pool over
objects and levels within each category before computing Capture.}
\label{tab:controlled-category-capture}
\begin{tabular}{@{}llrrr@{}}\toprule
Category & Transformation & Vanilla & Supervised Vanilla & GGF\\\midrule
Person & Rotation & $7.41\pm2.03$ & $16.17\pm0.45$ & $17.86\pm0.27$\\
Person & Scale & $7.61\pm2.02$ & $16.36\pm0.44$ & $18.35\pm0.31$\\
Person & Brightness & $7.65\pm2.00$ & $16.27\pm0.48$ & $18.25\pm0.32$\\
Car & Rotation & $11.15\pm1.10$ & $16.60\pm0.37$ & $15.08\pm0.29$\\
Car & Scale & $14.82\pm1.42$ & $21.28\pm0.51$ & $20.04\pm0.39$\\
Car & Brightness & $14.63\pm1.53$ & $21.01\pm0.50$ & $19.78\pm0.39$\\
Dog & Rotation & $11.18\pm0.53$ & $25.37\pm0.49$ & $27.50\pm0.59$\\
Dog & Scale & $11.63\pm0.55$ & $26.72\pm0.43$ & $28.87\pm0.56$\\
Dog & Brightness & $11.68\pm0.49$ & $26.76\pm0.41$ & $28.96\pm0.55$\\
Chair & Rotation & $2.59\pm0.80$ & $6.55\pm0.39$ & $7.10\pm0.24$\\
Chair & Scale & $4.11\pm1.14$ & $9.94\pm0.41$ & $10.64\pm0.28$\\
Chair & Brightness & $4.18\pm1.11$ & $10.07\pm0.39$ & $10.79\pm0.25$\\
Potted plant & Rotation & $3.32\pm0.73$ & $7.96\pm0.33$ & $11.25\pm0.13$\\
Potted plant & Scale & $3.99\pm0.90$ & $10.00\pm0.25$ & $13.46\pm0.16$\\
Potted plant & Brightness & $3.96\pm0.90$ & $9.64\pm0.31$ & $12.85\pm0.08$\\
Vase & Rotation & $5.78\pm0.67$ & $8.62\pm0.16$ & $9.73\pm0.03$\\
Vase & Scale & $8.25\pm1.19$ & $12.51\pm0.11$ & $13.65\pm0.10$\\
Vase & Brightness & $8.19\pm1.16$ & $12.50\pm0.08$ & $13.57\pm0.09$\\
\bottomrule\end{tabular}\end{table*}

\paragraph{Qualitative semantic assessment.}
For every category-associated unit, we inspect the eight strongest
distinct original views from the same evaluation pool. The 90 cards
cover every method, category and fitted model in this study. Mechanical
ranking fixes the examples before inspection. Method identifiers were
hidden during the initial assessment, and the authors subsequently
verified the images and common-theme explanations. This is a
descriptive qualitative assessment, not an independent human agreement
study or a statistical coherence test.

The criterion is a recognizable common object, part, shape, material or
other meaningful visual property across the strongest responses. A
property may span categories. Matching the assigned category and having
an interpretable within-concept variation axis are separate questions.
GGF and supervised Vanilla consistently show recognizable object themes.
The reviewed GGF responses are more consistently coherent than those of
unsupervised Vanilla, SLiCS and Concept-SAE, although many of their
units also show clear themes. Concept-SAE's chair-associated units
include the clearest mixed-response cases. The selected groups and
restricted object pool bound these observations.

\paragraph{Reading PCA maps.}
Following contribution-space PCA \citep{fel2026}, each BSF group's first
three training-derived principal coordinates set the RGB channels in
Figure~\ref{fig:earlier-response-geometry}. A group norm indicates where it
responds, while PCA color shows the relative location of its contribution
in that group's coordinate system. Bases, signs and display limits differ
across groups. Compare colors within a group's coordinate system rather
than treating them as common semantic labels or an absolute response scale.
Figure~\ref{fig:response-geometry} applies contribution PCA to the matched
75-group coherence examples using a fixed audit reference cohort instead
(Appendix~\ref{app:coherence-pca}). Its border and outline colors annotate
marked content separately from the RGB coordinates.

\subsection{All-category fingerprints and direction agreement}
\label{app:category-fingerprints}
\begin{figure*}[!tp]\centering
\includegraphics[width=\textwidth]{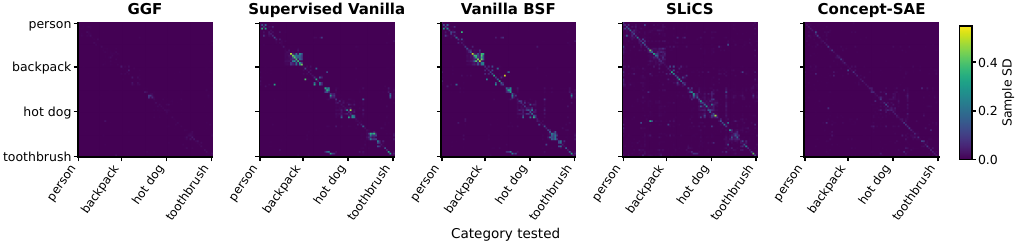}
\caption{Sample SD of the category-response matrices in
Figure~\ref{fig:category-fingerprints}(a). Each fitted model's matrix
averages its two image halves. Rows, columns and category order match
Figure~\ref{fig:category-fingerprints}(a), with a common SD scale across methods.}
\label{fig:category-fingerprint-SD}\end{figure*}

We test whole-group object focus/reproducibility across all five native
interfaces and, separately, within-group direction agreement for the
four twelve-direction methods against energy-matched regroupings.
\paragraph{Models, data and fixed readouts.}
We reuse GGF, supervised Vanilla, unsupervised Vanilla, SLiCS and
Concept-SAE checkpoints. No parameters, probes,
loss weights or group memberships are updated. The data are all 400
images in the cached COCO development pool, with all 75 categories
retained by the original coverage rule. Shared features, normalization,
masks and image IDs are identical across methods. We verify checkpoint
and feature checksums before inference and checkpoint checksums afterward.
All 1,024 groups in each BSF, 75 SLiCS dictionaries, and 111 Concept-SAE
tokens are profiled. Primary summaries average the 75 existing
category-associated readouts: the training-selected contrast group for
each BSF category and the corresponding native category dictionary or
token for the alternatives. These assignments remain fixed. We neither
search for a better group on these images nor discard poorly responding
readouts. A group shared by two category assignments contributes to both
category summaries. All-unit results remain archived separately from
this equal-category comparison.

\paragraph{Two image halves and support requirements.}
A fixed random seed generates 200/200 image partitions. We retain the
first partition with at least three annotation-positive images per
category in each half, using masks alone and no model outputs. Each
category has at least ten positive images in the complete pool.
The halves estimate fingerprints independently, rather than selecting
directions in one set and reporting their best responses in the other.
Every direction and category enters the defined summaries.
These development images were already inspected. This is an exploratory
cross-image check within categories used for guidance, rather than an
independent confirmation of transfer to categories excluded from
training or discovery of the model's own concepts.

Detection uses all 196 patches per image. A patch is positive at category
coverage $\geq0.5$, negative at coverage $\leq0.1$, and otherwise ignored.
Crowd coverage above 0.1 is ignored. This full-grid diagnostic differs
from the primary benchmark's fixed 32-patch samples. For instance
contrasts, known foreground coverage must sum to at least one patch and
there must be eight category-negative patches. Foreground pooling uses
coverage weights and comparison pooling uses the category-negative mean,
as in Equation~\eqref{eq:auditcap}. There are 1,936 usable object
instances. Handbag has only two usable contrast-instance images in one
half. Its raw results are retained, while the supported Capture and
direction-contribution summaries use the 74 categories with at least
three usable contrast images in both halves. Detection and direction-response
summaries retain all 75 categories.

\paragraph{What counts as an object-associated fingerprint.}
Semantic alignment with annotated visual concepts motivates the
category-response analysis \citep{bau2017}. Here $\mathcal C$ contains
all 75 retained categories. For image half $h\in\{A,B\}$,
let $A^{h}_{gc}$ be exact Patch AP for native unit $g$ and category $c$,
including ties, and $\pi^{h}_c$ the valid patches' positive prevalence.
The unit's response fingerprint is
\begin{equation}
F^{h}_{gc}=\max\left(0,
\frac{A^{h}_{gc}-\pi^{h}_c}{1-\pi^{h}_c}\right),
\label{eq:fingerprint-ap}
\end{equation}
Prevalence-aware evaluation motivates adjusting for category frequency
\citep[Sec.~4.1.1]{everingham2015retrospective}. Our transformation
subtracts positive-patch prevalence and rescales by the remaining distance
to perfect AP. It differs from their normalization of precision before
averaging. BSFs use gated norms, SLiCS uses its independent per-category
NNLS coefficients, and Concept-SAE uses its original mask-times-presence
response. Training logistic classifier heads are not used.

We separately record signed single-group Capture $C^{h}_{gc}$ for every
unit and category, extending the single-block recovery perspective
\citep{fel2026} to the regional contrasts in
Equation~\eqref{eq:auditcap}. Thus response focus and reconstructed
object information have separate fingerprints. For Concept-SAE, Capture
uses each native token's actual additive output, including the original
nonlinear projector, bias and token aggregator
(Appendix~\ref{app:concept-sae-capture}). Its 36 free tokens contribute
to the all-unit archive. They have no category mask, so their auxiliary
spatial response is their decoded contribution norm. Primary comparisons
use the original 75 concept-token responses. SLiCS dictionaries remain
independent projections and are not summed into a joint reconstruction.

Write $F_g^h=(F_{gc}^h)_{c\in\mathcal C}$ for a unit's category profile.
For a nonnegative profile $v$, define $n(v)=v/\|v\|_2$, with $n(0)=0$.
Whole-group reproducibility is
$n(F^A_g)^\top n(F^B_g)$, and is also computed separately for positive
Capture profiles. A fixed category-permutation control measures the
similarity remaining when category identities in one half are scrambled.
Let $\overline F=(F^A+F^B)/2$ be the mean fingerprint matrix.
The intended-category response share for readout $g_c$ is
\begin{equation}
Q_c=\frac{\overline F_{g_c c}}
{\sum_{c'}\overline F_{g_c c'}}.
\label{eq:fingerprint-focus}
\end{equation}
A zero denominator gives zero. $Q_c$ is a share of response-profile
weight, not an object-information percentage or a probability. We retain
AP and Capture beside it because a narrow profile can be weak. Figure~\ref{fig:category-fingerprints}(a)
shows the unnormalized $\overline F$ matrix with one common color scale.

\paragraph{Inspecting individual directions.}
For direction $j\in\{1,\ldots,r\}$, let $D_{gj}$ be column $j$ of $D_g$
and $z_{ipgj}$ component $j$ of $z_g(x_{ip})$. For object $o$, define
$\delta z_{ogj}=\sum_p w_{op}z_{ipgj}$, using its image $i$.
Its directional contrast contribution is $a_{ogj}=\delta z_{ogj}D_{gj}$;
the whole group contributes $m_{og}=\sum_j a_{ogj}$. Let
$\mathcal O_c^h$ contain eligible objects of category $c$ in half $h$,
and $C^h_{g\setminus j,c}$ be Capture after removing direction $j$.
The signed directional fingerprint is
\begin{equation}
U^{h}_{gjc}=C^{h}_{gc}-C^{h}_{g\setminus j,c},
\label{eq:direction-utility}
\end{equation}
where $C^{h}_{g\setminus j,c}$ uses $m_{og}-a_{ogj}$. Codes, decoder and
gate stay fixed during this subtraction. It asks whether that direction
helps that group's object reconstruction. For nonorthogonal
directions the exact difference is
\begin{equation}
U^{h}_{gjc}=
\frac{\sum_{o\in\mathcal O^h_c}
\left[2a_{ogj}^{\top}(\delta x_o-m_{og})+\|a_{ogj}\|_2^2\right]}
{\sum_{o\in\mathcal O^h_c}\|\delta x_o\|_2^2}.
\label{eq:direction-utility-exact}
\end{equation}
This retains interactions among directions instead of treating them as
orthogonal or ranking decoder vectors alone. Negative values identify
directions whose removal improves that regional reconstruction and are
preserved in the raw archive.

A second fingerprint $R^h_{gjc}$ measures response without using the
Capture objective. In every eligible image containing category $c$, we
subtract mean absolute coefficient strength on category-negative patches
from its category-coverage-weighted foreground mean, then average images
equally. Let $\mathcal I_c^h$ contain these eligible images in half $h$.
Writing $f_{icp}$ for normalized foreground-coverage weights and
$\overline b_{icp}=b_{icp}/\sum_{p'}b_{icp'}$ for uniform valid
category-negative weights gives
\begin{equation}
R^h_{gjc}=\frac1{|\mathcal I_c^h|}\sum_{i\in\mathcal I_c^h}
\sum_p(f_{icp}-\overline b_{icp})|z_{ipgj}|.
\label{eq:direction-response}\end{equation}
Each eligible image has equal weight. This uses all relevant patches
rather than choosing visually favorable examples or only the largest responses. We compare positive
parts of $U$ and $R$ separately, since a shared helpful category profile
and a shared response profile are distinct properties.

\paragraph{Agreement and matched regrouping control.}
\begin{figure*}[!tp]\centering
\includegraphics[width=.96\textwidth]{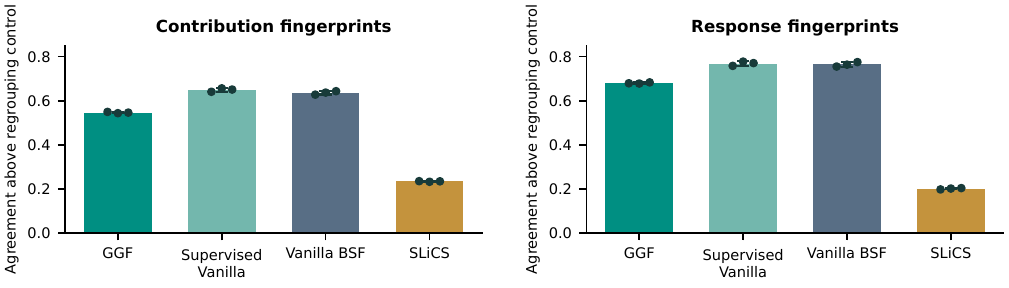}
\caption{Category profiles show agreement among group directions in their
contributions (left) and responses (right). Bars report lift over
energy-matched random regrouping, with mean $\pm$ sample SD across fitted
models and dots for individual runs. All twelve directions and eligible
categories are included.}
\label{fig:direction-fingerprints}\end{figure*}

For either direction fingerprint $V\in\{U,R\}$, let
$p^h_{gj}=n(\max(V^h_{gj},0))$. For direction indices
$j,\ell\in\{1,\ldots,r\}$, agreement for a width-$r$ group is
\begin{equation}
W_g=\frac{1}{r(r-1)}\sum_{j\ne\ell}
(p^A_{gj})^{\top}p^B_{g\ell},\qquad r=12.
\label{eq:direction-agreement}
\end{equation}
All 132 ordered pairs enter, comparing different directions across
separate image halves. Zero positive profiles earn zero instead of being
removed. Normalizing each category profile separates the category pattern
from a direction's overall amplitude. We also archive the fraction of
directions with a positive profile and same-direction reproducibility.

To ask whether agreement exceeds that expected from directions with
similar activity, let $z_{gj}$ denote its scalar coefficient and
$\mathbb E_A$ average over half-A patches. Its decoded energy is
$E_{gj}=\mathbb E_A[z_{gj}^2]\|D_{gj}\|_2^2$. We sort directions into ten
equal-count energy bins and generate 100 fixed random permutations
within those bins. Each permutation reassigns direction identities to
the existing width-twelve slots and is applied to both image halves.
Thus widths and slot energy bins are preserved, as is each direction's
cross-half fingerprint. The reported grouping lift is $W_g$ minus the
mean agreement of these random regroupings. This is a control on the
fingerprint statistic, not a newly trained or physically exchanged model.
The control matches approximate decoded energy rather than every
activation statistic. The permutations provide a descriptive
reference rather than a calibrated hypothesis-test distribution.

\paragraph{Native interfaces and coordinate dependence.}
GGF, both Vanilla controls and SLiCS expose twelve additive directions
per group. Concept-SAE remains in every whole-token comparison without
artificial scalar coordinates. Direction agreement characterizes the
learned basis. For BSFs, a coordinated orthogonal rotation of the encoder
and decoder preserves group norms and decoded contributions
\citep[Sec.~3.5]{fel2026}, while individual-direction fingerprints can
change. Whole-group focus and reproducibility are therefore more direct
measures of object-associated coherence. This symmetry does not extend
to arbitrary rotations of SLiCS's nonnegative codes.

\paragraph{Aggregation, uncertainty and verification.}
For each model, we average scores over the fixed category readouts, giving
categories equal weight. We then report the mean and sample SD across
training runs. Random regrouping repetitions and directions are not
treated as independent model replicates. Paired seed differences and all
category results remain available in the evidence archive. We verified
the removal formula against explicit subtraction and exact AP against
the standard tied-score implementation. Across all saved models,
180 checks against the archived native interfaces agree to
within $8.0\times10^{-6}$ AP. This confirms interface fidelity rather
than creating a new performance comparison.

\paragraph{Whole-group results.}
GGF's intended-category response share and response-profile
reproducibility exceed each comparator in every seed
(Table~\ref{tab:fingerprint-groups}). Figure~\ref{fig:category-fingerprint-SD}
shows variation in the response matrices across seeds. Capture-profile
reproducibility is similar for GGF and SLiCS, and supervised Vanilla is
slightly higher. GGF therefore has stronger whole-group object focus,
with a particularly clear response-profile advantage over the supervised
alternatives, rather than an advantage on every fingerprint property.

\begin{table*}[t]\centering\small\setlength{\tabcolsep}{4pt}
\caption{Whole-group fingerprints: mean $\pm$ sample SD. Response and Capture reproducibility are cosine similarities.
Category share is the intended category's fraction of positive response
profile weight. Capture covers 74 supported categories, and the other
summaries cover the 75 fixed category readouts. These full-grid development
diagnostics are separate from the primary benchmark scores.}
\label{tab:fingerprint-groups}
\begin{tabular}{@{}lrrrrr@{}}\toprule
Method & Capture (\%) & Patch AP (\%) & \shortstack{Response\\reproducibility} & \shortstack{Capture\\reproducibility} & \shortstack{Category\\share (\%)}\\\midrule
GGF & $41.6\pm0.2$ & $80.6\pm0.3$ & $0.9594\pm0.0013$ & $0.9685\pm0.0001$ & $77.1\pm0.4$\\
Supervised Vanilla & $40.5\pm0.2$ & $62.7\pm1.4$ & $0.9265\pm0.0110$ & $0.9728\pm0.0030$ & $63.9\pm1.6$\\
Vanilla BSF & $22.1\pm0.9$ & $52.4\pm2.3$ & $0.9091\pm0.0220$ & $0.9638\pm0.0036$ & $59.8\pm2.4$\\
SLiCS & $25.0\pm0.7$ & $52.9\pm1.7$ & $0.8893\pm0.0166$ & $0.9680\pm0.0004$ & $46.5\pm1.3$\\
Concept-SAE & $1.1\pm0.1$ & $32.4\pm0.4$ & $0.7818\pm0.0017$ & $0.5577\pm0.0080$ & $55.4\pm0.6$\\
\bottomrule\end{tabular}\end{table*}

\paragraph{Direction-level results and their interpretation.}

Both the contribution and response fingerprints give GGF higher
grouping lift than SLiCS in every seed
(Figure~\ref{fig:direction-fingerprints},
Table~\ref{tab:fingerprint-directions}). The Vanilla controls have higher
agreement than GGF under both definitions. These results distinguish
GGF's stronger whole-group object focus from the closer alignment of
individual native direction profiles in the controls. Alongside the
qualitative object themes, the comparison supports greater
object-associated coherence over SLiCS and Concept-SAE at the whole-unit
level, with additional internal-direction evidence against SLiCS.
Object labels do not describe every shared shape, part or texture, so
this operational definition concerns object-associated organization.

\begin{table}[t]\centering\small\setlength{\tabcolsep}{3pt}
\caption{Direction grouping lift: mean $\pm$ sample SD. Contribution fingerprints compare signed Capture-removal
effects through their positive parts. Response fingerprints compare
foreground/background coefficient differences.}
\label{tab:fingerprint-directions}
\begin{tabular}{@{}lrr@{}}\toprule
Method & Contribution & Response\\\midrule
GGF & $0.5461\pm0.0029$ & $0.6798\pm0.0027$\\
Supervised Vanilla & $0.6488\pm0.0079$ & $0.7683\pm0.0098$\\
Vanilla BSF & $0.6356\pm0.0078$ & $0.7641\pm0.0101$\\
SLiCS & $0.2342\pm0.0014$ & $0.2012\pm0.0031$\\
\bottomrule\end{tabular}\end{table}

\subsection{Matched 75-group coherence and concentration}\label{app:matched75}
\paragraph{Purpose of the group budget.}
SLiCS uses one 12-direction dictionary per category, while the BSF
methods allow a choice of group count. A common 75-group setting for
the 75-category evaluation therefore matches SLiCS's native count and
makes exhaustive inspection feasible. Vanilla, supervised Vanilla and
GGF retain 75 twelve-direction groups. This does not assume that each
learned group represents one named category; parameter counts,
activity and native inference still differ. Concept-SAE 111 remains
exploratory, and native category-split SLiCS 37/38 is an unmatched
seed 21 reference. The 1,024- and 4,096-group experiments retain their
original protocols and are never pooled with this study.

\begin{table*}[!tbp]\centering\small
\caption{GPT coherence scores by seed (\%). Column headings 21--27 are
training seeds. Sup. Vanilla denotes supervised Vanilla BSF.
Category-split scores include all native groups.}
\label{tab:matched75-per-run}
\begin{minipage}[t]{.49\textwidth}\centering
\setlength{\tabcolsep}{2.1pt}
\begin{tabular*}{\linewidth}{@{\extracolsep{\fill}}lrrrrrrr@{}}\toprule
\multicolumn{8}{@{}l}{\textbf{Specific-pair meaning}}\\\midrule
Method & 21 & 22 & 23 & 24 & 25 & 26 & 27\\
\addlinespace[2pt]\multicolumn{8}{@{}l}{\textit{Category-split guidance}}\\
GGF & 81.78 & 78.00 & 79.33 & 81.33 & 77.56 & 79.56 & 81.56\\
Sup. Vanilla & 68.89 & 77.33 & 72.67 & 70.67 & 70.00 & 75.11 & 74.67\\
Vanilla BSF & 54.67 & 63.11 & 61.33 & 58.67 & 58.22 & 68.44 & 58.67\\
\addlinespace[2pt]\multicolumn{8}{@{}l}{\textit{All category labels}}\\
GGF & 93.33 & 93.78 & 95.11 & 92.89 & 90.22 & 93.78 & 94.67\\
SLiCS & 80.89 & 83.56 & 84.44 & 87.11 & 80.89 & 81.33 & 85.78\\
Sup. Vanilla & 63.11 & 65.33 & 67.56 & 70.22 & 72.00 & 66.67 & 70.22\\
Vanilla BSF & 54.67 & 57.33 & 54.22 & 61.78 & 63.56 & 66.22 & 61.33\\
\bottomrule\end{tabular*}
\end{minipage}
\hfill
\begin{minipage}[t]{.49\textwidth}\centering
\setlength{\tabcolsep}{2.1pt}
\begin{tabular*}{\linewidth}{@{\extracolsep{\fill}}lrrrrrrr@{}}\toprule
\multicolumn{8}{@{}l}{\textbf{Pairwise visual structure}}\\\midrule
Method & 21 & 22 & 23 & 24 & 25 & 26 & 27\\
\addlinespace[2pt]\multicolumn{8}{@{}l}{\textit{Category-split guidance}}\\
GGF & 81.33 & 75.78 & 77.56 & 70.67 & 71.11 & 71.11 & 70.67\\
Sup. Vanilla & 69.33 & 75.56 & 72.89 & 61.78 & 63.11 & 66.00 & 67.33\\
Vanilla BSF & 57.78 & 60.89 & 61.33 & 55.11 & 52.00 & 59.11 & 54.22\\
\addlinespace[2pt]\multicolumn{8}{@{}l}{\textit{All category labels}}\\
GGF & 90.22 & 93.78 & 89.78 & 82.67 & 80.89 & 83.11 & 81.33\\
SLiCS & 86.22 & 82.22 & 84.44 & 78.67 & 76.44 & 74.22 & 80.89\\
Sup. Vanilla & 62.67 & 64.00 & 67.56 & 62.67 & 62.22 & 60.00 & 60.89\\
Vanilla BSF & 53.33 & 55.11 & 56.44 & 51.56 & 54.22 & 56.44 & 56.00\\
\bottomrule\end{tabular*}
\end{minipage}
\par\medskip
\begin{minipage}[t]{.49\textwidth}\centering
\setlength{\tabcolsep}{2.1pt}
\begin{tabular*}{\linewidth}{@{\extracolsep{\fill}}lrrrrrrr@{}}\toprule
\multicolumn{8}{@{}l}{\textbf{Clear whole-group theme}}\\\midrule
Method & 21 & 22 & 23 & 24 & 25 & 26 & 27\\
\addlinespace[2pt]\multicolumn{8}{@{}l}{\textit{Category-split guidance}}\\
GGF & 86.67 & 82.00 & 83.33 & 84.67 & 80.00 & 82.67 & 80.00\\
Sup. Vanilla & 75.33 & 78.67 & 82.67 & 72.67 & 71.33 & 75.33 & 72.00\\
Vanilla BSF & 65.33 & 69.33 & 74.67 & 57.33 & 57.33 & 66.67 & 60.00\\
\addlinespace[2pt]\multicolumn{8}{@{}l}{\textit{All category labels}}\\
GGF & 100.00 & 100.00 & 100.00 & 96.00 & 100.00 & 100.00 & 100.00\\
SLiCS & 97.33 & 97.33 & 100.00 & 96.00 & 97.33 & 96.00 & 96.00\\
Sup. Vanilla & 77.33 & 72.00 & 69.33 & 72.00 & 72.00 & 60.00 & 72.00\\
Vanilla BSF & 61.33 & 72.00 & 68.00 & 70.67 & 66.67 & 61.33 & 62.67\\
\bottomrule\end{tabular*}
\end{minipage}
\hfill
\begin{minipage}[t]{.49\textwidth}\centering
\setlength{\tabcolsep}{2.1pt}
\begin{tabular*}{\linewidth}{@{\extracolsep{\fill}}lrrrrrrr@{}}\toprule
\multicolumn{8}{@{}l}{\textbf{Clear theme with discriminating specificity}}\\\midrule
Method & 21 & 22 & 23 & 24 & 25 & 26 & 27\\
\addlinespace[2pt]\multicolumn{8}{@{}l}{\textit{Category-split guidance}}\\
GGF & 86.00 & 81.33 & 82.67 & 84.67 & 78.67 & 82.00 & 79.33\\
Sup. Vanilla & 74.67 & 77.33 & 81.33 & 72.67 & 71.33 & 75.33 & 72.00\\
Vanilla BSF & 65.33 & 69.33 & 74.67 & 57.33 & 57.33 & 66.67 & 57.33\\
\addlinespace[2pt]\multicolumn{8}{@{}l}{\textit{All category labels}}\\
GGF & 100.00 & 100.00 & 100.00 & 96.00 & 100.00 & 100.00 & 98.67\\
SLiCS & 97.33 & 97.33 & 100.00 & 96.00 & 96.00 & 94.67 & 96.00\\
Sup. Vanilla & 77.33 & 70.67 & 66.67 & 72.00 & 70.67 & 60.00 & 70.67\\
Vanilla BSF & 60.00 & 69.33 & 64.00 & 69.33 & 62.67 & 58.67 & 61.33\\
\bottomrule\end{tabular*}
\end{minipage}
\end{table*}

\begin{table*}[!tbp]\centering\small
\setlength{\tabcolsep}{3pt}
\caption{GPT whole-group outcomes and specificity, pooled counts across
seven training orders. Disagreements count groups with non-identical
reviewer votes on the panel's measure. Unknown votes are individual
insufficient-evidence votes.}
\label{tab:matched75-distributions}
\begin{tabular*}{.90\textwidth}{@{\extracolsep{\fill}}llrrrrrrr@{}}\toprule
\multicolumn{9}{@{}l}{\textbf{Theme outcome}}\\\midrule
Setting & Method & Clear & Partial & No theme & Insufficient & Unresolved & Disagreements & Unknown votes\\\midrule
Split & GGF & 869 & 177 & 4 & 0 & 0 & 37 & 0\\
Split & Supervised Vanilla & 792 & 245 & 11 & 0 & 2 & 75 & 1\\
Split & Vanilla BSF & 338 & 176 & 11 & 0 & 0 & 57 & 0\\
\midrule
All & GGF & 522 & 3 & 0 & 0 & 0 & 8 & 0\\
All & SLiCS & 510 & 15 & 0 & 0 & 0 & 25 & 0\\
All & Supervised Vanilla & 371 & 143 & 11 & 0 & 0 & 47 & 0\\
All & Vanilla BSF & 347 & 166 & 11 & 0 & 1 & 57 & 0\\
\bottomrule\end{tabular*}
\par\medskip
\begin{tabular*}{.90\textwidth}{@{\extracolsep{\fill}}llrrrrrr@{}}\toprule
\multicolumn{8}{@{}l}{\textbf{Theme specificity}}\\\midrule
Setting & Method & Specific & Broad, discriminating & Generic & Unclear & Unresolved & Disagreements\\\midrule
Split & GGF & 697 & 343 & 8 & 0 & 2 & 230\\
Split & Supervised Vanilla & 718 & 321 & 7 & 0 & 4 & 260\\
Split & Vanilla BSF & 310 & 203 & 5 & 0 & 7 & 156\\
\midrule
All & GGF & 432 & 92 & 1 & 0 & 0 & 95\\
All & SLiCS & 378 & 145 & 2 & 0 & 0 & 104\\
All & Supervised Vanilla & 302 & 213 & 7 & 0 & 3 & 141\\
All & Vanilla BSF & 276 & 230 & 15 & 0 & 4 & 145\\
\bottomrule\end{tabular*}
\end{table*}

\paragraph{Literature basis and choice of review tasks.}
ACE treats meaningfulness and coherence of visual examples as separate
requirements and tests coherence by asking people to identify the
conceptually different image among six \citep[Secs.~2,4]{ghorbani2019ace}.
CRAFT uses crop-level intruder detection, with the intruder and other
examples drawn from the same object class \citep[Sec.~4.2]{fel2023craft}.
These studies motivate inspecting shared content beyond a broad category
label. We apply that principle to fixed strongest-response patches.
Three disjoint pairs use all six views without reusing an image and ask
whether the marked regions share a specific meaning. A separate whole-set
task checks that one theme spans the group: three individually coherent
pairs could otherwise depict three different concepts. Requiring five
of six matches permits one exception while demanding a common theme
across most views. This is our explicit operational cutoff; the six-image
ACE task uses an inserted intruder and does not validate this cutoff.
Specificity is recorded separately so that a generic description does not
automatically establish a discriminating meaning.

Automated assessment of strongest-response examples also has a close
precedent. ELUDe compares grids of nine top-activating images using three
vision-language judges and majority consensus
\citep[App.~B.1]{bagci2026elude}. Its judgments are preferences between
two units' grids. Our absolute judgments concern marked pairs or one
six-view set, and use three fresh contexts of the same reviewer model.
The cited evaluation designs motivate our tasks without making their
scores interchangeable.

\paragraph{Fits, category allocation and leakage checks.}
Training orders 21--27 share initialization and source image pools.
BSFs use width 12 and Top-32; new fits use 10,640 updates. Two complementary
category folds guide on 37 or 38 categories. The 75-group capacity remains
fixed as guidance is reduced. Guide-only annotations determine
initialization, targets, losses, controllers and model selection;
excluded categories do not supply these signals. Only guided development
outcomes in order 21 select supervised settings, frozen for orders 22--27.
Excluded objects can still appear in reconstruction images. The task is
withheld supervision categories, not unseen visual exposure. Frozen
native units are calibrated after fitting; calibration may use the
excluded-category labels, so the test is not zero-shot recognition.
Shared Vanilla is one native bank per run, not two independent models.
Original weight, native-response, crop and card hashes were checked.

\begin{table*}[!tbp]\centering\small
\setlength{\tabcolsep}{4pt}
\caption{All category labels, negative quantile 0.95: mean $\pm$ sample SD. }
\label{tab:matched75-full_category-all_categories-95}
\begin{tabular}{@{}lrrrrrr@{}}\toprule
Method & Groups & Coverage & False pos. & Precision & Pos. overlap & Neg. overlap\\\midrule
GGF & 1 & $90.62\pm0.10$ & $5.05\pm0.07$ & $33.79\pm0.27$ & \textemdash & \textemdash\\
GGF & 2 & $93.15\pm0.23$ & $8.87\pm0.05$ & $23.90\pm0.10$ & $51.13\pm0.65$ & $17.72\pm0.45$\\
GGF & 4 & $94.70\pm0.26$ & $15.34\pm0.14$ & $16.51\pm0.11$ & $33.62\pm0.43$ & $15.04\pm0.35$\\
GGF & 8 & $96.50\pm0.09$ & $26.78\pm0.20$ & $11.11\pm0.07$ & $18.09\pm0.33$ & $9.75\pm0.23$\\
SLiCS & 1 & $79.60\pm0.79$ & $4.66\pm0.11$ & $32.76\pm0.61$ & \textemdash & \textemdash\\
SLiCS & 2 & $84.48\pm0.94$ & $7.91\pm0.29$ & $24.37\pm0.50$ & $54.76\pm3.07$ & $23.76\pm1.52$\\
SLiCS & 4 & $87.80\pm0.58$ & $13.51\pm0.53$ & $17.34\pm0.40$ & $37.06\pm1.83$ & $17.79\pm1.49$\\
SLiCS & 8 & $91.32\pm0.76$ & $23.70\pm0.87$ & $11.82\pm0.27$ & $20.57\pm1.17$ & $11.70\pm0.76$\\
Supervised Vanilla & 1 & $62.59\pm1.83$ & $4.97\pm0.17$ & $26.22\pm1.16$ & \textemdash & \textemdash\\
Supervised Vanilla & 2 & $70.89\pm2.03$ & $9.36\pm0.26$ & $18.66\pm0.72$ & $41.87\pm4.58$ & $8.09\pm1.43$\\
Supervised Vanilla & 4 & $78.21\pm1.67$ & $17.32\pm0.46$ & $12.83\pm0.39$ & $25.78\pm1.85$ & $6.75\pm0.76$\\
Supervised Vanilla & 8 & $84.82\pm1.20$ & $30.87\pm0.73$ & $8.76\pm0.15$ & $14.22\pm0.51$ & $5.30\pm0.29$\\
Vanilla BSF & 1 & $55.11\pm1.77$ & $4.97\pm0.12$ & $23.82\pm1.17$ & \textemdash & \textemdash\\
Vanilla BSF & 2 & $63.94\pm1.87$ & $9.46\pm0.21$ & $17.09\pm0.76$ & $37.41\pm2.42$ & $7.82\pm1.21$\\
Vanilla BSF & 4 & $72.05\pm2.06$ & $17.41\pm0.41$ & $11.95\pm0.40$ & $25.87\pm1.93$ & $6.61\pm0.55$\\
Vanilla BSF & 8 & $79.98\pm1.65$ & $31.12\pm0.51$ & $8.23\pm0.15$ & $14.45\pm0.87$ & $5.11\pm0.26$\\
\bottomrule\end{tabular}\end{table*}

\begin{table*}[!tbp]\centering\small
\setlength{\tabcolsep}{4pt}
\caption{Categories withheld from guidance, negative quantile 0.95: mean $\pm$ sample SD. }
\label{tab:matched75-category_split-excluded-95}
\begin{tabular}{@{}lrrrrrr@{}}\toprule
Method & Groups & Coverage & False pos. & Precision & Pos. overlap & Neg. overlap\\\midrule
GGF & 1 & $66.13\pm1.87$ & $5.09\pm0.14$ & $26.93\pm0.91$ & \textemdash & \textemdash\\
GGF & 2 & $74.42\pm1.14$ & $9.35\pm0.15$ & $19.33\pm0.35$ & $45.63\pm1.46$ & $11.55\pm0.79$\\
GGF & 4 & $82.63\pm1.00$ & $17.04\pm0.23$ & $13.67\pm0.19$ & $27.10\pm1.17$ & $9.01\pm0.54$\\
GGF & 8 & $89.02\pm0.47$ & $29.97\pm0.34$ & $9.36\pm0.06$ & $14.58\pm0.60$ & $6.64\pm0.21$\\
Supervised Vanilla & 1 & $59.20\pm1.76$ & $5.27\pm0.23$ & $24.21\pm0.84$ & \textemdash & \textemdash\\
Supervised Vanilla & 2 & $68.28\pm1.77$ & $9.67\pm0.25$ & $17.89\pm0.56$ & $41.63\pm2.50$ & $9.46\pm1.53$\\
Supervised Vanilla & 4 & $76.58\pm1.45$ & $17.54\pm0.45$ & $12.65\pm0.22$ & $27.93\pm1.49$ & $7.55\pm0.90$\\
Supervised Vanilla & 8 & $83.42\pm0.98$ & $31.07\pm0.61$ & $8.65\pm0.14$ & $14.53\pm0.87$ & $5.60\pm0.40$\\
Vanilla BSF & 1 & $57.15\pm1.99$ & $5.22\pm0.10$ & $23.62\pm0.93$ & \textemdash & \textemdash\\
Vanilla BSF & 2 & $65.33\pm2.24$ & $9.75\pm0.22$ & $16.99\pm0.71$ & $40.19\pm1.17$ & $8.00\pm1.54$\\
Vanilla BSF & 4 & $73.46\pm1.88$ & $17.82\pm0.45$ & $11.91\pm0.47$ & $27.27\pm1.59$ & $6.68\pm0.83$\\
Vanilla BSF & 8 & $80.93\pm1.94$ & $31.51\pm0.63$ & $8.19\pm0.28$ & $15.06\pm0.84$ & $5.25\pm0.35$\\
\bottomrule\end{tabular}\end{table*}

\begin{table*}[!tbp]\centering\small
\setlength{\tabcolsep}{4pt}
\caption{Guided categories in category-split fits, negative quantile 0.95: mean $\pm$ sample SD. }
\label{tab:matched75-category_split-guided-95}
\begin{tabular}{@{}lrrrrrr@{}}\toprule
Method & Groups & Coverage & False pos. & Precision & Pos. overlap & Neg. overlap\\\midrule
GGF & 1 & $90.07\pm0.20$ & $4.71\pm0.03$ & $34.86\pm0.12$ & \textemdash & \textemdash\\
GGF & 2 & $92.22\pm0.17$ & $8.93\pm0.08$ & $23.50\pm0.16$ & $41.03\pm1.26$ & $14.51\pm0.54$\\
GGF & 4 & $94.38\pm0.16$ & $16.29\pm0.24$ & $15.71\pm0.17$ & $24.45\pm1.06$ & $10.69\pm0.68$\\
GGF & 8 & $96.17\pm0.16$ & $29.38\pm0.32$ & $10.08\pm0.07$ & $13.15\pm0.59$ & $7.03\pm0.33$\\
Supervised Vanilla & 1 & $81.28\pm1.82$ & $4.77\pm0.15$ & $32.66\pm0.49$ & \textemdash & \textemdash\\
Supervised Vanilla & 2 & $85.74\pm1.29$ & $8.86\pm0.23$ & $22.56\pm0.40$ & $54.27\pm2.43$ & $9.72\pm1.12$\\
Supervised Vanilla & 4 & $90.00\pm0.96$ & $16.97\pm0.16$ & $14.67\pm0.23$ & $29.89\pm0.99$ & $7.57\pm0.52$\\
Supervised Vanilla & 8 & $93.27\pm0.64$ & $30.50\pm0.29$ & $9.58\pm0.14$ & $14.97\pm0.74$ & $5.68\pm0.36$\\
Vanilla BSF & 1 & $57.15\pm1.99$ & $5.22\pm0.10$ & $23.62\pm0.93$ & \textemdash & \textemdash\\
Vanilla BSF & 2 & $65.33\pm2.24$ & $9.75\pm0.22$ & $16.99\pm0.71$ & $40.19\pm1.17$ & $8.00\pm1.54$\\
Vanilla BSF & 4 & $73.46\pm1.88$ & $17.82\pm0.45$ & $11.91\pm0.47$ & $27.27\pm1.59$ & $6.68\pm0.83$\\
Vanilla BSF & 8 & $80.93\pm1.94$ & $31.51\pm0.63$ & $8.19\pm0.28$ & $15.06\pm0.84$ & $5.25\pm0.35$\\
\bottomrule\end{tabular}\end{table*}

\begin{table*}[!tbp]\centering\small
\setlength{\tabcolsep}{4pt}
\caption{All category labels, negative quantile 0.90: mean $\pm$ sample SD. }
\label{tab:matched75-full_category-all_categories-90}
\begin{tabular}{@{}lrrrrrr@{}}\toprule
Method & Groups & Coverage & False pos. & Precision & Pos. overlap & Neg. overlap\\\midrule
GGF & 1 & $93.48\pm0.46$ & $9.96\pm0.15$ & $21.93\pm0.21$ & \textemdash & \textemdash\\
GGF & 2 & $96.55\pm0.11$ & $16.68\pm0.10$ & $15.46\pm0.09$ & $65.71\pm1.24$ & $22.87\pm0.46$\\
GGF & 4 & $97.60\pm0.11$ & $27.02\pm0.22$ & $10.88\pm0.06$ & $47.81\pm0.66$ & $19.82\pm0.19$\\
GGF & 8 & $98.88\pm0.15$ & $42.69\pm0.33$ & $7.71\pm0.06$ & $31.64\pm0.32$ & $15.94\pm0.25$\\
SLiCS & 1 & $85.98\pm1.23$ & $9.25\pm0.31$ & $21.87\pm0.62$ & \textemdash & \textemdash\\
SLiCS & 2 & $91.26\pm0.71$ & $14.88\pm0.51$ & $16.35\pm0.45$ & $68.71\pm2.60$ & $28.46\pm1.68$\\
SLiCS & 4 & $94.22\pm0.45$ & $23.62\pm0.86$ & $11.94\pm0.26$ & $55.92\pm1.43$ & $24.72\pm1.30$\\
SLiCS & 8 & $96.38\pm0.45$ & $37.75\pm1.49$ & $8.43\pm0.21$ & $37.89\pm1.37$ & $19.31\pm1.05$\\
Supervised Vanilla & 1 & $75.30\pm2.62$ & $9.67\pm0.32$ & $18.96\pm1.06$ & \textemdash & \textemdash\\
Supervised Vanilla & 2 & $83.54\pm1.97$ & $17.55\pm0.32$ & $13.29\pm0.50$ & $57.41\pm4.39$ & $12.05\pm1.78$\\
Supervised Vanilla & 4 & $89.87\pm1.30$ & $30.39\pm0.56$ & $9.32\pm0.21$ & $41.38\pm1.81$ & $11.60\pm0.91$\\
Supervised Vanilla & 8 & $94.63\pm0.84$ & $49.87\pm0.66$ & $6.47\pm0.12$ & $26.61\pm1.15$ & $9.75\pm0.54$\\
Vanilla BSF & 1 & $68.65\pm2.15$ & $9.69\pm0.24$ & $17.57\pm0.53$ & \textemdash & \textemdash\\
Vanilla BSF & 2 & $77.03\pm1.53$ & $17.35\pm0.34$ & $12.64\pm0.22$ & $51.98\pm1.29$ & $13.14\pm0.96$\\
Vanilla BSF & 4 & $85.27\pm0.97$ & $30.18\pm0.36$ & $8.95\pm0.07$ & $40.52\pm1.34$ & $11.54\pm0.45$\\
Vanilla BSF & 8 & $91.76\pm1.16$ & $49.82\pm0.55$ & $6.28\pm0.09$ & $26.83\pm1.15$ & $9.46\pm0.52$\\
\bottomrule\end{tabular}\end{table*}

\begin{table*}[!tbp]\centering\small
\setlength{\tabcolsep}{4pt}
\caption{Categories withheld from guidance, negative quantile 0.90: mean $\pm$ sample SD. }
\label{tab:matched75-category_split-excluded-90}
\begin{tabular}{@{}lrrrrrr@{}}\toprule
Method & Groups & Coverage & False pos. & Precision & Pos. overlap & Neg. overlap\\\midrule
GGF & 1 & $79.17\pm1.27$ & $9.73\pm0.12$ & $19.45\pm0.34$ & \textemdash & \textemdash\\
GGF & 2 & $87.74\pm0.55$ & $17.52\pm0.28$ & $13.86\pm0.15$ & $62.92\pm1.76$ & $15.58\pm0.98$\\
GGF & 4 & $93.50\pm0.38$ & $29.35\pm0.22$ & $9.96\pm0.08$ & $46.71\pm0.89$ & $14.92\pm0.28$\\
GGF & 8 & $97.05\pm0.43$ & $47.35\pm0.49$ & $7.04\pm0.07$ & $29.33\pm0.52$ & $12.28\pm0.34$\\
Supervised Vanilla & 1 & $71.73\pm1.16$ & $9.95\pm0.14$ & $17.88\pm0.37$ & \textemdash & \textemdash\\
Supervised Vanilla & 2 & $81.45\pm0.98$ & $18.05\pm0.23$ & $12.81\pm0.25$ & $55.38\pm2.88$ & $13.02\pm1.16$\\
Supervised Vanilla & 4 & $88.62\pm0.86$ & $30.83\pm0.43$ & $9.13\pm0.14$ & $42.45\pm1.63$ & $11.86\pm0.82$\\
Supervised Vanilla & 8 & $93.88\pm0.58$ & $50.22\pm0.37$ & $6.44\pm0.08$ & $27.41\pm1.27$ & $9.84\pm0.44$\\
Vanilla BSF & 1 & $69.45\pm1.53$ & $9.67\pm0.15$ & $17.78\pm0.37$ & \textemdash & \textemdash\\
Vanilla BSF & 2 & $78.75\pm1.02$ & $17.44\pm0.35$ & $12.79\pm0.29$ & $55.23\pm1.85$ & $13.19\pm1.56$\\
Vanilla BSF & 4 & $85.90\pm0.66$ & $30.10\pm0.74$ & $9.03\pm0.24$ & $42.53\pm1.17$ & $11.50\pm1.25$\\
Vanilla BSF & 8 & $92.20\pm0.60$ & $50.03\pm1.00$ & $6.27\pm0.16$ & $27.51\pm1.16$ & $9.33\pm0.63$\\
\bottomrule\end{tabular}\end{table*}

\begin{table*}[!tbp]\centering\small
\setlength{\tabcolsep}{4pt}
\caption{Guided categories in category-split fits, negative quantile 0.90: mean $\pm$ sample SD. }
\label{tab:matched75-category_split-guided-90}
\begin{tabular}{@{}lrrrrrr@{}}\toprule
Method & Groups & Coverage & False pos. & Precision & Pos. overlap & Neg. overlap\\\midrule
GGF & 1 & $93.68\pm0.17$ & $9.64\pm0.08$ & $22.36\pm0.14$ & \textemdash & \textemdash\\
GGF & 2 & $96.67\pm0.21$ & $17.05\pm0.18$ & $15.24\pm0.11$ & $56.75\pm0.88$ & $20.49\pm0.54$\\
GGF & 4 & $97.57\pm0.26$ & $28.12\pm0.20$ & $10.63\pm0.05$ & $41.44\pm1.53$ & $17.87\pm0.55$\\
GGF & 8 & $98.77\pm0.16$ & $46.10\pm0.43$ & $7.21\pm0.05$ & $25.26\pm0.95$ & $13.21\pm0.35$\\
Supervised Vanilla & 1 & $88.06\pm1.60$ & $9.57\pm0.25$ & $21.62\pm0.44$ & \textemdash & \textemdash\\
Supervised Vanilla & 2 & $92.63\pm0.62$ & $17.06\pm0.24$ & $14.82\pm0.22$ & $68.58\pm2.28$ & $13.69\pm1.29$\\
Supervised Vanilla & 4 & $95.68\pm0.38$ & $29.67\pm0.24$ & $10.00\pm0.13$ & $44.51\pm1.18$ & $12.26\pm0.56$\\
Supervised Vanilla & 8 & $97.87\pm0.27$ & $49.15\pm0.60$ & $6.75\pm0.09$ & $26.71\pm1.18$ & $10.03\pm0.40$\\
Vanilla BSF & 1 & $69.45\pm1.53$ & $9.67\pm0.15$ & $17.78\pm0.37$ & \textemdash & \textemdash\\
Vanilla BSF & 2 & $78.75\pm1.02$ & $17.44\pm0.35$ & $12.79\pm0.29$ & $55.23\pm1.85$ & $13.19\pm1.56$\\
Vanilla BSF & 4 & $85.90\pm0.66$ & $30.10\pm0.74$ & $9.03\pm0.24$ & $42.53\pm1.17$ & $11.50\pm1.25$\\
Vanilla BSF & 8 & $92.20\pm0.60$ & $50.03\pm1.00$ & $6.27\pm0.16$ & $27.51\pm1.16$ & $9.33\pm0.63$\\
\bottomrule\end{tabular}\end{table*}

\paragraph{Coherence scores.}\label{app:coherence-formulas}
For group $g$, fixed pair $p\in\{1,2,3\}$ and reviewer context
$r\in\{1,2,3\}$, let $y^e_{gpr}$ indicate a yes vote on endpoint
$e\in\{\mathrm{semantic},\mathrm{visual}\}$. With $\mathcal G$ the complete
planned group bank, the majority indicator and pair score are
\begin{equation}
m^e_{gp}=\mathbf1\!\left[\sum_{r=1}^3y^e_{gpr}\geq2\right],\quad
S_e=\frac{1}{3|\mathcal G|}\sum_{g\in\mathcal G}\sum_{p=1}^3m^e_{gp}.
\label{eq:pair-coherence}\end{equation}
Let $a_{gr}$ indicate a clear whole-group judgment and $b_{gr}$ a
specific or broad-but-discriminating specificity judgment. Define
$A_g=\mathbf1[\sum_r a_{gr}\geq2]$ and
$B_g=\mathbf1[\sum_r b_{gr}\geq2]$. The two whole-group rates are
\begin{equation}
S_{\rm whole}=\frac{\sum_g A_g}{|\mathcal G|},\qquad
S_{\rm discr}=\frac{\sum_g A_gB_g}{|\mathcal G|}.
\label{eq:whole-coherence}\end{equation}
A clear judgment requires one theme matching at least five of six HIGH
views. A vote that is unclear, insufficient or unresolved is not counted as yes
or clear, and never reduces these planned denominators.
Outcome and specificity distributions count groups; disagreement counts
groups with non-identical votes, and unknown counts raw
insufficient-evidence votes.

\paragraph{All-group evidence and blinded review.}
Pair and whole-group reviews use GPT-6.1 Sol
(\texttt{gpt-6.1-sol}) with high reasoning effort
\citep{openai2026gpt61sol}. Archived execution metadata verifies the
inherited model settings for all actual reviewer identities, including
the reused seed 21 reviews. Fresh contexts are review replicates, not
independent models or human raters.

Each group's six strongest audit images are selected by the signed mean
of its four largest native patch responses over 196 positions, without
clipping. Deterministic image ties and hash presentation fix the six
views and three disjoint pairs. Up to four positive marked patches and
their original-resolution zoom strips are inspected. Category/method
names, native unit identities, numerical scores and other reviews are
hidden. The semantic endpoint requires a specific shared object, part,
action, relation, material or region at both marks. Generic edges or a
broad umbrella category are insufficient. Specific visual structure has
its own yes/no/unclear vote; cross-category geometry and textures count.
Scores divide majority-yes pairs by all three planned pairs per group,
with unclear outcomes retained. A group is not required to match its
intended training category to be coherent.

A supplemental, separately blinded whole-group review asks for one
theme across all six HIGH responses: clear at 5/6 or 6/6, partial at 3/6 or 4/6,
and no clear theme at 0--2/6. Specificity is separately specific, broad
but discriminating, generic/nondiscriminating or unclear. Majority-clear
plus at least two discriminating-specificity votes is secondary. LOW
views are absent, so HIGH--LOW contrast is not assessed. Unknown evidence
and unresolved votes remain in the denominator. The whole-group request
followed the pair results; it is not an independent confirmatory
allocation, and these are AI reviewers working in separate fresh contexts, rather than human raters.

There are 675 matched groups per run and 4,725 across all seven orders.
Matched pair review covers 14,175 fixed pairs per context. Including 75
native reference groups from order 21 gives 14,400 pairs per context and
43,200 two-endpoint assessments across three contexts. Whole review
covers 4,725 matched groups and yields 14,175 ratings. The order 21
judgments and provenance are reused after evidence-identity audits.
Orders 22--23 add 339 packets for each review task and orders 24--27
add 675 for each. Review annotations are finalized before aggregation
and retained with their source hashes. Independent recounts reproduce
majorities, denominators, unique group IDs and source hashes, and verify
that original seed 21 scores and completed numeric/concentration results
remain unchanged.

\paragraph{Presence-based concentration and false positives.}
All 1,900 crops receive blinded category-presence labels: 400 calibration
and 1,500 audit crops. Calibration uses the original confirmation cohort,
which is separate from training, development and audit images.
A fixed 40-crop second AI review covers 3,000 category
cells; canonical labels remain unchanged after QA. Negative response
quantiles 0.95 (primary) and 0.90 (sensitivity) define per-group thresholds,
with strict exceedance. Calibration balanced accuracy ranks all 75
candidate groups for each category. Subsets of 1/2/4/8 ranked groups
freeze before audit scoring and use an OR detector. Coverage is the
fraction of known target-present crops detected; false positives are
the fraction of known target-absent crops detected. Unknown presence
labels are not imputed negative. At least five positive and ten negative
crops are required in both splits. Truck and hot dog lack calibration
support; all 75 targets stay in support tables, with rates over 73 supported
targets, never zero imputation. No group is selected because it won a
coherence review. Positive/negative overlap, rank-wise unique detection
and incremental coverage are retained in the accompanying statistics.
The one-to-eight-group curve concerns detection concentration around
these predefined targets, not an interchangeable reconstruction-Capture
score or a proof of fragmentation of every discovered concept.

The two rates are conventional true-positive and false-positive rates
\citep{fawcett2006roc}. Restricting the number of selected native groups
makes them a test of response concentration: how much target coverage is
accessible through a small subset, at its measured false-positive cost.
The 95th-percentile rule sets a common calibration operating point;
it does not force the audit false-positive rate to equal 5\%.
Budgets 1, 2, 4, 8 examine progressively larger, nested subsets. Unlike a
standard threshold-swept ROC curve, these curves vary subset size while
retaining calibration-fixed thresholds.

\begin{table*}[!tbp]\centering\small
\setlength{\tabcolsep}{3pt}
\caption{Full-grid category-mask Patch AP and regional Capture (\%) by
training seed, with mean $\pm$ sample SD over seeds 21--27.
Each metric selects its group on calibration data and scores it on audit
data. The split settings pool categories across complementary folds.
The same Vanilla checkpoint contributes to both target strata.}
\label{tab:matched75-AP}
\begin{tabular*}{.95\textwidth}{@{\extracolsep{\fill}}lrrrrrrrr@{}}\toprule
\multicolumn{9}{@{}l}{\textbf{Full-grid Patch AP}}\\\midrule
Method & Mean $\pm$ SD & 21 & 22 & 23 & 24 & 25 & 26 & 27\\\midrule
\multicolumn{9}{@{}l}{\textit{All category labels}}\\
GGF & $74.38\pm0.30$ & 73.75 & 74.45 & 74.60 & 74.43 & 74.31 & 74.43 & 74.70\\
SLiCS & $49.93\pm1.02$ & 49.43 & 49.14 & 51.67 & 50.47 & 48.61 & 50.40 & 49.81\\
Supervised Vanilla & $31.12\pm2.40$ & 33.25 & 28.70 & 31.97 & 30.83 & 34.86 & 29.93 & 28.30\\
Vanilla BSF & $20.96\pm1.13$ & 19.61 & 20.37 & 21.47 & 23.16 & 21.08 & 20.64 & 20.38\\
\multicolumn{9}{@{}l}{\textit{Category split, guided targets}}\\
GGF & $69.72\pm0.29$ & 69.92 & 69.87 & 69.91 & 69.46 & 69.77 & 69.92 & 69.17\\
Supervised Vanilla & $55.65\pm2.14$ & 51.21 & 56.53 & 57.12 & 56.68 & 55.24 & 55.28 & 57.52\\
Vanilla BSF & $22.35\pm1.43$ & 20.83 & 20.79 & 22.64 & 24.79 & 23.29 & 22.52 & 21.59\\
\multicolumn{9}{@{}l}{\textit{Category split, withheld targets}}\\
GGF & $22.92\pm1.17$ & 23.03 & 20.47 & 23.58 & 23.06 & 23.03 & 23.05 & 24.22\\
Supervised Vanilla & $22.19\pm0.69$ & 22.15 & 22.52 & 21.94 & 21.31 & 22.32 & 23.43 & 21.64\\
Vanilla BSF & $22.35\pm1.43$ & 20.83 & 20.79 & 22.64 & 24.79 & 23.29 & 22.52 & 21.59\\
\bottomrule\end{tabular*}
\par\medskip
\begin{tabular*}{.95\textwidth}{@{\extracolsep{\fill}}lrrrrrrrr@{}}\toprule
\multicolumn{9}{@{}l}{\textbf{Regional Capture}}\\\midrule
Method & Mean $\pm$ SD & 21 & 22 & 23 & 24 & 25 & 26 & 27\\\midrule
\multicolumn{9}{@{}l}{\textit{All category labels}}\\
GGF & $40.96\pm0.11$ & 40.99 & 40.86 & 40.99 & 41.16 & 40.85 & 40.90 & 41.01\\
SLiCS & $26.05\pm0.60$ & 26.61 & 24.98 & 25.97 & 26.48 & 25.69 & 25.97 & 26.64\\
Supervised Vanilla & $28.73\pm0.62$ & 29.33 & 27.72 & 29.35 & 28.88 & 28.80 & 28.99 & 28.04\\
Vanilla BSF & $9.08\pm0.56$ & 9.64 & 8.57 & 8.90 & 10.03 & 8.53 & 8.99 & 8.91\\
\multicolumn{9}{@{}l}{\textit{Category split, guided targets}}\\
GGF & $43.55\pm0.15$ & 43.37 & 43.65 & 43.35 & 43.75 & 43.56 & 43.53 & 43.61\\
Supervised Vanilla & $38.60\pm0.42$ & 39.12 & 38.38 & 38.24 & 38.07 & 39.17 & 38.57 & 38.66\\
Vanilla BSF & $8.52\pm0.46$ & 8.88 & 8.24 & 8.09 & 9.37 & 8.14 & 8.50 & 8.43\\
\multicolumn{9}{@{}l}{\textit{Category split, withheld targets}}\\
GGF & $20.34\pm0.10$ & 20.26 & 20.17 & 20.32 & 20.40 & 20.37 & 20.50 & 20.35\\
Supervised Vanilla & $17.71\pm0.58$ & 17.48 & 16.92 & 17.62 & 18.11 & 17.25 & 18.69 & 17.88\\
Vanilla BSF & $8.52\pm0.46$ & 8.88 & 8.24 & 8.09 & 9.37 & 8.14 & 8.50 & 8.43\\
\bottomrule\end{tabular*}
\end{table*}

\begin{table*}[!tbp]\centering\small
\setlength{\tabcolsep}{4pt}
\caption{Native full-bank reconstruction on the audit set. MSE and VE
are mean $\pm$ sample SD across seeds 21--27. VE is reported for the full
bank in each setting. The SLiCS concept projections do not supply a
comparable additive full-bank reconstruction.}
\label{tab:matched75-reconstruction}
\begin{tabular}{@{}llrr@{}}\toprule
Setting & Method & MSE & VE\\\midrule
All labels & GGF & $0.1762\pm0.0032$ & $0.8237\pm0.0032$\\
All labels & SLiCS & N/A & N/A\\
All labels & Supervised Vanilla & $0.1428\pm0.0020$ & $0.8571\pm0.0020$\\
All labels & Vanilla BSF & $0.1417\pm0.0021$ & $0.8582\pm0.0022$\\
Category split & GGF & $0.1379\pm0.0003$ & $0.8620\pm0.0003$\\
Category split & Supervised Vanilla & $0.1382\pm0.0009$ & $0.8617\pm0.0009$\\
Category split & Vanilla BSF & $0.1293\pm0.0004$ & $0.8706\pm0.0004$\\
\bottomrule\end{tabular}\end{table*}

\paragraph{Concentration and overlap formulas.}\label{app:presence-formulas}
Let $s_{ig}$ be group $g$'s crop-level response for image $i$: the mean
of its four largest signed native patch responses across all 196 positions,
without clipping. For category $c$, let $\mathcal P_c$ and $\mathcal N_c$
be crops with known positive and negative presence. Calibration sets group threshold
$\tau_{gc}$ to negative-response quantile $q$. Let $\mathrm{TPR}_{gc}$ and
$\mathrm{FPR}_{gc}$ be that threshold's calibration true- and false-positive
rates. Balanced accuracy $\mathrm{BA}_{gc}=(\mathrm{TPR}_{gc}+1-\mathrm{FPR}_{gc})/2$
ranks groups. For subset budget $k\in\{1,2,4,8\}$, separate from encoder
Top-$k$, let $K_{c,k}$ be its frozen top-ranked subset. The OR prediction is
\begin{equation}
\widehat y_{ic}^{(k)}=
\mathbf1\!\left[\max_{g\in K_{c,k}}(s_{ig}-\tau_{gc})>0\right].
\label{eq:presence-or}\end{equation}
Let $TP_c$ and $FP_c$ count these detections in $\mathcal P_c$ and
$\mathcal N_c$, respectively. Coverage $\mathrm{Cov}_c$, false-positive
rate $\mathrm{FPR}_c$ and precision $\mathrm{Prec}_c$ are
\begin{equation}
\mathrm{Cov}_c=\frac{TP_c}{|\mathcal P_c|},\quad
\mathrm{FPR}_c=\frac{FP_c}{|\mathcal N_c|},\quad
\mathrm{Prec}_c=\frac{TP_c}{TP_c+FP_c}.
\label{eq:presence-rates}\end{equation}
Precision is undefined when there are no positive predictions. Unknown
presence labels enter neither set. Each rate is averaged equally over
categories with the required support; undefined values are retained as
missing, not imputed zero.

For sign $\sigma\in\{+,-\}$, let $D^\sigma_{gc}$ be the crops in
$\mathcal P_c$ or $\mathcal N_c$ detected by group $g$ alone. With
$\mathcal E^\sigma_{c,k}$ the selected-group pairs having nonempty union,
mean prediction overlap is
\begin{equation}
J^\sigma_{c,k}=\frac1{|\mathcal E^\sigma_{c,k}|}
\sum_{(g,h)\in\mathcal E^\sigma_{c,k}}
\frac{|D^\sigma_{gc}\cap D^\sigma_{hc}|}{|D^\sigma_{gc}\cup D^\sigma_{hc}|}.
\label{eq:prediction-jaccard}\end{equation}
Zero-union pairs are excluded and counted separately. Overlap is
undefined for a one-group subset. Incremental coverage at rank $j$ is
the fraction detected by that group and no preceding selected group;
unique coverage is the fraction detected by that group and none of the
other selected groups. Both use the same positive or negative denominators
as Equation~\eqref{eq:presence-rates}.

\paragraph{Aggregation and uncertainty.}
All concentration tables use 0--100 scales for coverage, false positives,
precision and Jaccard overlap, with 73 supported of 75 planned targets per
run. One-group overlap is undefined and never zero-imputed.

The matched 75-group summaries use seven run-level scores
(seeds 21--27). Sample SD
uses the $n-1$ denominator at the training-order level. Coherence combines equal-size 75-group folds
within each run; shared Vanilla counts once. Presence, Patch AP and
Capture use equal-category macro averages after fold aggregation.
Reconstruction is averaged across fitted banks for each method and run.
Only then are means and SD computed across runs. Full-grid category-mask
AP in this study differs from the 32-patch-per-image benchmark;
they are reported separately. All 75 AP categories remain supported.
The archive retains every per-run value, per-bank/fold distribution,
specificity outcome, unknown, disagreement and source fingerprint.
Shared initialization and image pools preclude interpreting these SDs
as independent-model population uncertainty. No new superiority test
or confidence claim is inferred. Independent human validation, visual
target alignment and a genuinely new-image test remain outstanding.

\paragraph{Cross-model coherence check.}\label{app:cross-model}
We repeated the blinded evaluation with
Gemini 3.5 Flash-Lite \citep{google2026gemini35flashlite}
(API ID \texttt{google/gemini-3.5-flash-lite}) via OpenRouter and
Google AI Studio Flex, with minimal reasoning effort and temperature zero. Gemini judged the same 4,725 matched groups
from orders 21--27. Each group received one fresh call for the three
fixed pairs and a separate call for the whole-group task. Both calls
received the frozen PNG and lossless crops of its numbered views and
zoom strips. The same rubric applied, with prompts that explicitly
excluded the review overlays as evidence and requested concise pair
explanations. Method identities and GPT judgments were hidden. Pilot
calls informed the prompt choices and are excluded from the results.

\begin{table*}[!tbp]\centering\small
\setlength{\tabcolsep}{4pt}
\caption{Cross-model check with Gemini 3.5 Flash-Lite. Scores are mean
$\pm$ sample SD across seven training orders (\%). Whole + discr. requires
a clear theme rated specific or broad but discriminating. Rejections count
missing pair-task and whole-group judgments, respectively. Shared Vanilla
is counted once in the category-split setting.}
\label{tab:matched75-gemini}
\begin{tabular}{@{}lrrrrr@{}}\toprule
Method & Specific pairs & Visual pairs & Whole clear & Whole + discr. & Rejections\\\midrule
\multicolumn{6}{@{}l}{\textit{All category labels}}\\
GGF & $79.37\pm2.86$ & $69.59\pm3.41$ & $92.38\pm2.14$ & $92.38\pm2.14$ & 0/1\\
SLiCS & $69.84\pm1.65$ & $62.79\pm3.63$ & $86.48\pm5.68$ & $86.10\pm6.10$ & 2/8\\
Supervised Vanilla & $51.62\pm3.36$ & $47.17\pm2.77$ & $58.29\pm2.74$ & $57.90\pm2.97$ & 0/18\\
Vanilla BSF & $43.43\pm4.83$ & $41.33\pm4.22$ & $49.33\pm3.77$ & $48.57\pm4.34$ & 3/27\\
\midrule
\multicolumn{6}{@{}l}{\textit{Category-split guidance}}\\
GGF & $64.13\pm1.24$ & $58.63\pm1.75$ & $72.00\pm2.52$ & $71.81\pm2.66$ & 3/25\\
Supervised Vanilla & $59.40\pm2.98$ & $54.89\pm2.53$ & $65.62\pm5.50$ & $65.43\pm5.60$ & 3/27\\
Vanilla BSF & $47.17\pm5.06$ & $45.02\pm5.06$ & $51.62\pm3.33$ & $51.05\pm3.75$ & 5/25\\\bottomrule
\end{tabular}
\end{table*}

Of the 9,450 replies, 16 pair-task and 131 whole-group replies failed
rubric validation. These remain missing, with planned denominators of
14,175 pairs and 4,725 whole groups. GGF leads all aggregate endpoints
even if every missing baseline judgment counts as positive. It leads
specific-pair coherence in every order under both guidance settings.
Gemini's absolute coherence scores are lower than the GPT-majority
scores. Among comparable judgments, exact agreement with the GPT
majority is 77.4\% for specific-pair coherence (10,932/14,127),
75.1\% for visual structure (10,613/14,127), 78.0\% for whole-group
coherence (3,582/4,594), and 54.2\% for specificity (2,490/4,594).
The method rankings therefore agree in aggregate across both models,
while absolute rates and specificity depend on the reviewer. Both
models use the same image evidence, and this comparison uses one
Gemini judgment per task and the majority of three GPT contexts.

\section{Computational resources and measurement scope}\label{app:compute-reporting}
\subsection{Fitting costs across datasets}\label{app:fit-costs}
\begin{figure*}[!tp]\centering
\includegraphics[width=\textwidth]{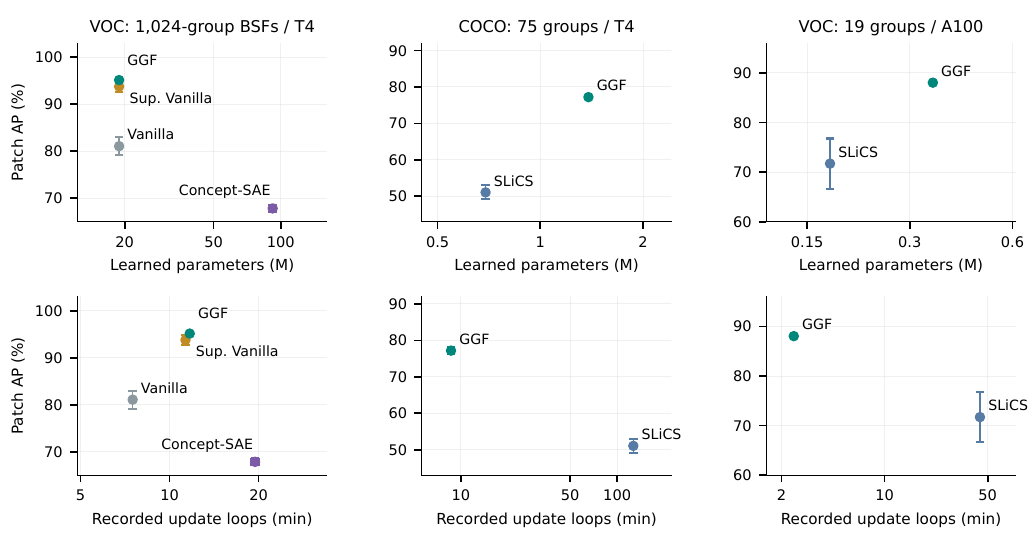}
\caption{Performance and fitting resources in three separate comparisons.
Columns retain the VOC 1,024-group/T4, COCO 75-group/T4 and VOC
19-group/A100 protocols. Top: learned parameters versus Patch AP.
Bottom: recorded update-loop minutes versus Patch AP. Error bars show
sample SD where measured. Higher points have better AP, while points
farther left use fewer parameters or less loop time. Timers exclude
validation, checkpoint transport and search. Devices and native training
schedules are not pooled.}
\label{fig:final-resources}
\end{figure*}

\begin{figure*}[!tp]\centering
\includegraphics[width=\textwidth]{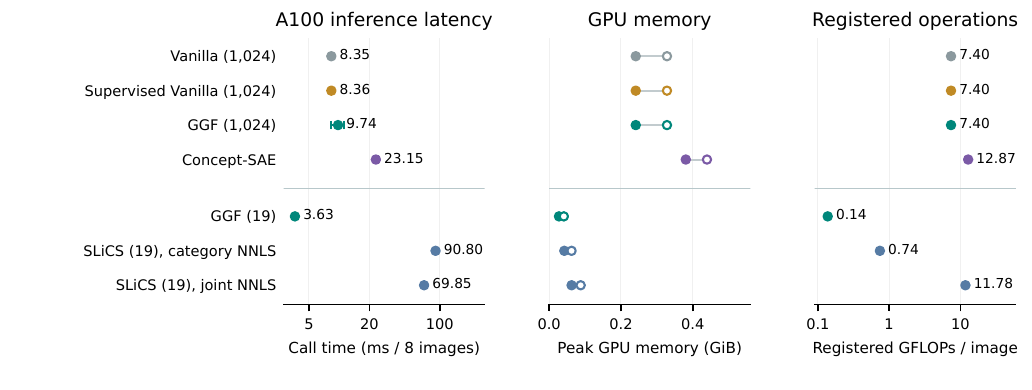}
\caption{A100 execution profiles on the same eight VOC training images.
Left: inference-call time, with sample SD over three independent calls.
Center: maximum allocated (filled) and reserved (open) GPU memory.
Right: registered GFLOPs per image from one instrumented call.
Labels in parentheses give BSF/SLiCS group counts. SLiCS category and
joint NNLS serve object scoring and full reconstruction, respectively.
The separator distinguishes the 1,024-group/native-token study from the
19-group study. Latency and operation counts use logarithmic axes.
These calls exclude the frozen DINOv3 backbone and retain each method's native outputs.}
\label{fig:execution-profiles}\end{figure*}

\begin{table}[!htbp]\centering\small
{\setlength{\tabcolsep}{3pt}
\caption{VOC fitting resources under two protocols. Update loops: minutes, mean $\pm$ sample SD. GPU peaks: maximum allocated/reserved GiB across fits. Devices and group counts differ between blocks (Appendix~\ref{app:compute-reporting}).}
\label{tab:compute-main}
\begin{tabular}{@{}lrrr@{}}\toprule
Method & Params (M) & Update min & GPU peaks\\\midrule
\multicolumn{4}{l}{T4: 1,024 groups for BSF arms}\\
Vanilla BSF & 18.89 & $7.50\pm0.02$ & 0.74/0.85\\
Supervised Vanilla & 18.89 & $11.32\pm0.03$ & 0.87/2.10\\
GGF & 18.89 & $11.70\pm0.04$ & 1.04/2.21\\
Concept-SAE & 92.00 & $19.47\pm0.66$ & 1.32/1.48\\
\midrule\multicolumn{4}{l}{A100: 19-group SLiCS comparison}\\
GGF & 0.350 & $2.42\pm0.03$ & 0.074/0.102\\
SLiCS & 0.175 & $44.31\pm0.08$ & 0.055/0.074\\
\bottomrule\end{tabular}
}\end{table}

\begin{table}[!htbp]\centering\small
{\setlength{\tabcolsep}{3pt}
\caption{COCO 75-group T4 fitting resources. Loop time is mean $\pm$ sample SD in minutes. GPU peaks were not recorded.}
\label{tab:coco-slics-cost}
\begin{tabular}{@{}lrr@{}}\toprule
Method & Parameters & Update min\\\midrule
GGF & 1,383,452 & $8.66\pm0.33$\\
SLiCS & 691,200 & $126.96\pm1.63$\\
\bottomrule\end{tabular}
}\end{table}

\begin{table}[!htbp]\centering\small
{\setlength{\tabcolsep}{3pt}
\caption{VOC A100 fit-invocation time in minutes (mean $\pm$ sample SD). This includes initialization and development checks but excludes final evaluation.}
\label{tab:voc-fit-wall}
\begin{tabular}{@{}lrr@{}}\toprule
Method & Update loops & Fit invocation\\\midrule
GGF & $2.42\pm0.03$ & $6.40\pm0.36$\\
SLiCS & $44.31\pm0.08$ & $59.26\pm0.90$\\
\bottomrule\end{tabular}
}\end{table}

Fitting costs and saved-model execution answer separate questions.
Tables~\ref{tab:compute-main}--\ref{tab:voc-fit-wall} and
Figure~\ref{fig:final-resources} retain each hardware, dataset and native
group-count protocol rather than pooling them.

\paragraph{Resource summaries.}
Let $\Theta_{\rm learned}$ contain learned tensors $\theta$, excluding
frozen DINOv3 and evaluation probes. With $\operatorname{numel}$ counting
tensor entries, parameters total
$\sum_{\theta\in\Theta_{\rm learned}}\operatorname{numel}(\theta)$. Allocated and reserved
memory are separate recorded maxima, including resumed fitting
invocations. For method $m$, let $\overline T_m$ be its unrounded mean time
within one protocol block, $R_m$ its time ratio to GGF, and $H_{\rm SV}$
GGF's percentage time overhead over supervised Vanilla (SV):
\begin{equation}
R_m=\frac{\overline T_m}{\overline T_{\rm GGF}},\quad
H_{\rm SV}=100\left(\frac{\overline T_{\rm GGF}}{\overline T_{\rm SV}}-1\right).
\label{eq:resource-ratios}\end{equation}
For $F_{\rm call}$ registered operations in a call with $N_{\rm images}$
images, operations per image are $F_{\rm call}/N_{\rm images}$.
Means and sample SD follow Equation~\eqref{eq:mean-sample-sd}, with the
declared training runs or independent calls as units.

\begin{table}[!tb]\centering\small
\setlength{\tabcolsep}{2.4pt}
\caption{VOC A100 execution profiles excluding the DINOv3 backbone. Times are mean $\pm$ sample SD over three calls. GPU peaks list allocated and reserved memory. GFLOPs use one instrumented call and count each fused multiply-add as two operations.}
\label{tab:execution-profiles}
\begin{tabular}{@{}lrrrr@{}}\toprule
Phase & Images & \shortstack{Call\\(ms)} & \shortstack{Peak GPU\\(GiB)} & \shortstack{GFLOPs\\per image}\\\midrule
\multicolumn{5}{@{}l}{\textit{GGF (19)}}\\
Inference & 8 & $3.63\pm0.11$ & 0.027/0.041 & 0.137\\
Ordinary update & 8 & $43.39\pm1.22$ & 0.057/0.068 & 0.415\\
Controller update & 8 & $57.89\pm0.08$ & 0.058/0.070 & 0.663\\
\multicolumn{5}{@{}l}{\textit{SLiCS (19)}}\\
Category NNLS & 8 & $90.80\pm0.50$ & 0.042/0.062 & 0.742\\
Joint NNLS & 8 & $69.85\pm0.68$ & 0.063/0.088 & 11.779\\
Training update & 8 & $244.30\pm11.07$ & 0.054/0.070 & 1.614\\
\multicolumn{5}{@{}l}{\textit{GGF (1,024)}}\\
Inference & 8 & $9.74\pm1.45$ & 0.241/0.328 & 7.399\\
Ordinary update & 8 & $60.41\pm0.84$ & 0.590/0.748 & 25.907\\
Controller update & 8 & $85.77\pm0.05$ & 0.624/0.854 & 40.710\\
\multicolumn{5}{@{}l}{\textit{Vanilla (1,024)}}\\
Inference & 8 & $8.35\pm0.03$ & 0.241/0.328 & 7.399\\
Training update & 8 & $17.80\pm0.09$ & 0.453/0.594 & 18.497\\
\multicolumn{5}{@{}l}{\textit{Supervised Vanilla (1,024)}}\\
Inference & 8 & $8.36\pm0.10$ & 0.241/0.328 & 7.399\\
Training update & 8 & $59.07\pm0.16$ & 0.590/0.748 & 25.907\\
\multicolumn{5}{@{}l}{\textit{Concept-SAE}}\\
Inference & 8 & $23.15\pm0.97$ & 0.381/0.439 & 12.869\\
Stage 1 update & 64 & $198.00\pm36.34$ & 0.577/0.643 & 8.820\\
Stage 2 update & 64 & $152.37\pm20.27$ & 0.576/0.643 & 4.500\\
Stage 3 update & 64 & $383.13\pm9.52$ & 1.319/1.477 & 21.405\\
\bottomrule\end{tabular}
\end{table}
\paragraph{Training schedules and timings.}
T4 VOC BSFs use 1,024 groups, 7,640 warm-up and 3,000 continuation
updates, and eight images per update. Each Vanilla arm is charged the
shared random warm-up; supervised Vanilla includes its selected loss
pair. Concept-SAE sums its three native stages, with 2,090 updates and
effective batch 64 in eight-image microbatches. The measured loops give
GGF 3.4\% overhead over supervised Vanilla and a $1.66\times$ speedup
over Concept-SAE. Concept-SAE's COCO loops average 76.23 minutes;
corresponding BSF loop coverage is incomplete.

The category-matched SLiCS studies each use 10,640 updates. Their GGF
loop speedups are $14.67\times$ on COCO/T4 with 75 groups and
$18.33\times$ on VOC/A100 with 19 groups. The VOC fit-invocation ratio
is instead $9.26\times$: it includes initialization and development
checks, while excluding final evaluation, resource profiling and result
packaging. All underlying times and their sample SD remain in the tables.

\paragraph{Timer boundaries and memory.}
BSF loops include losses, backpropagation, model/controller updates and
synchronization after batch/noise preparation. SLiCS includes its NNLS
solve and dictionary update; Concept-SAE includes microbatch handling.
Loop timers exclude initialization, feature preparation, development
checks, checkpoint transport and search.

Faster loops need not use less memory. In the VOC 19-group study,
GGF's 350,476 learned parameters include an untied encoder, decoder,
classifier heads and controller; SLiCS has 175,104 dictionary parameters.
GGF has larger allocated/reserved fitting peaks there. In the T4
1,024-group comparison, allocated memory favors GGF over Concept-SAE,
while reserved memory does not. Tables retain every peak. COCO 75-group
GPU peaks were not recorded. Loading a completed checkpoint measures
reuse, not its original fitting cost.

\subsection{Execution profiles and operation coverage}\label{app:execution-profiles}

We profile three independent cold calls on disposable copies of the
seed-21 VOC models on A100-SXM4-40GB, with TF32 off. Copies reset between
calls and source checkpoints stay intact. Timers span input preparation
through synchronized return and exclude construction, loading,
validation and checkpoint I/O.

\paragraph{Registered FLOPs and coverage.}\label{app:flop-coverage}
PyTorch's registered formulas count a fused multiply-add as two
operations and cover executed matrix/convolution operations, including
backward for updates. Elementwise operations, reductions, indexing,
eigendecompositions and external SciPy/BLAS CPU work remain uncounted.
Thus Table~\ref{tab:execution-profiles} reports partial GFLOPs per image,
with coverage limits in \path{FLOP_COVERAGE.json}. That file, the
per-phase operator inventories and hardware records are in
\path{evidence/voc_slics_final/}.

\paragraph{Execution measurements.}
Inference uses the same eight training images. BSFs execute encoder,
gate and decoder; Concept-SAE retains its native output. SLiCS uses
independent category NNLS for object scores and joint NNLS for full
reconstruction. GGF's measured inference speedups are $2.38\times$
over Concept-SAE at 1,024 groups, and $25.02\times$/$19.25\times$ over
SLiCS category/joint NNLS at 19 groups. The table gives all means, SDs
and peak allocated/reserved memory; the peaks are maxima, not means.

Training uses batches of eight for BSFs and SLiCS, and effective batch
64 in eight-image microbatches for Concept-SAE. BSFs restore available
optimizer state and Concept-SAE starts fresh Adam moments for each stage
update. We time controller updates separately because they run every
ten steps. These cold calls complement the whole-fit timings.

\clearpage
\nobalance
\onecolumn
\section{Qualitative comparisons}
\label{app:qualitative-current}
\label{app:matched-blinded-examples}
\begingroup\centering
\noindent{\small\textbf{(a) GGF; group 14; G0197}\hfill
\footnotesize Specific pairs: 3/3; visual pairs: 3/3.}\par
{\footnotesize Overlapping feathers along a bird wing or body. Theme matches: 6/6, 6/6, 6/6 (clear).}\par
\noindent\makebox[\textwidth][c]{\includegraphics[width=.82\textwidth]{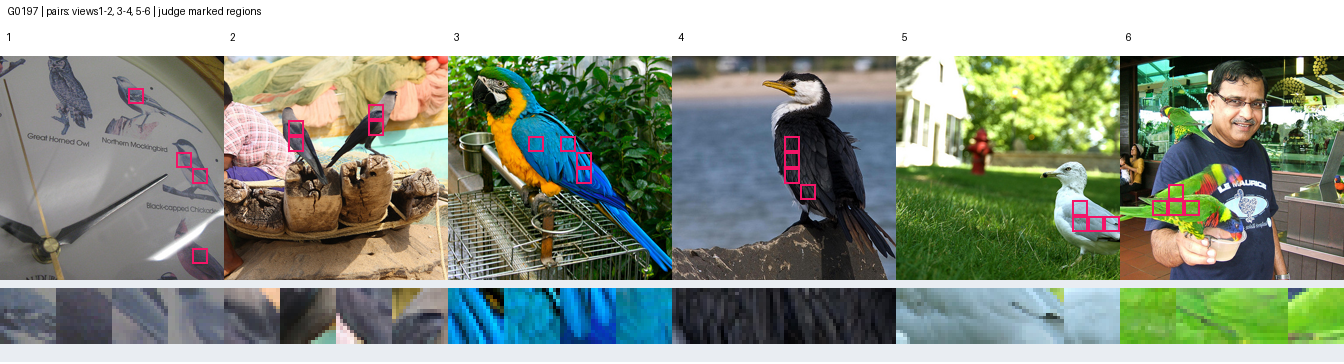}}\par\smallskip
\noindent{\small\textbf{(b) SLiCS; group 14; G0606}\hfill
\footnotesize Specific pairs: 1/3; visual pairs: 1/3.}\par
{\footnotesize Animal hair or feather coverings, with fine strands or layered edges. Theme matches: 6/6, 4/6, 6/6 (clear).}\par
\noindent\makebox[\textwidth][c]{\includegraphics[width=.82\textwidth]{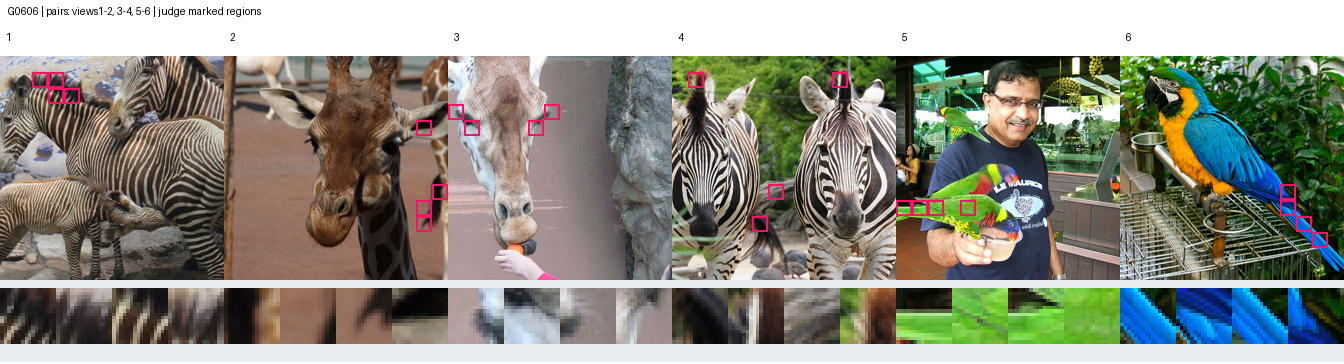}}\par\smallskip
\noindent{\small\textbf{(c) Supervised Vanilla BSF; group 35; G0057}\hfill
\footnotesize Specific pairs: 1/3; visual pairs: 1/3.}\par
{\footnotesize Human fingers wrapped around or gripping a handheld object. Theme matches: 3/6, 3/6, 6/6 (partial).}\par
\noindent\makebox[\textwidth][c]{\includegraphics[width=.82\textwidth]{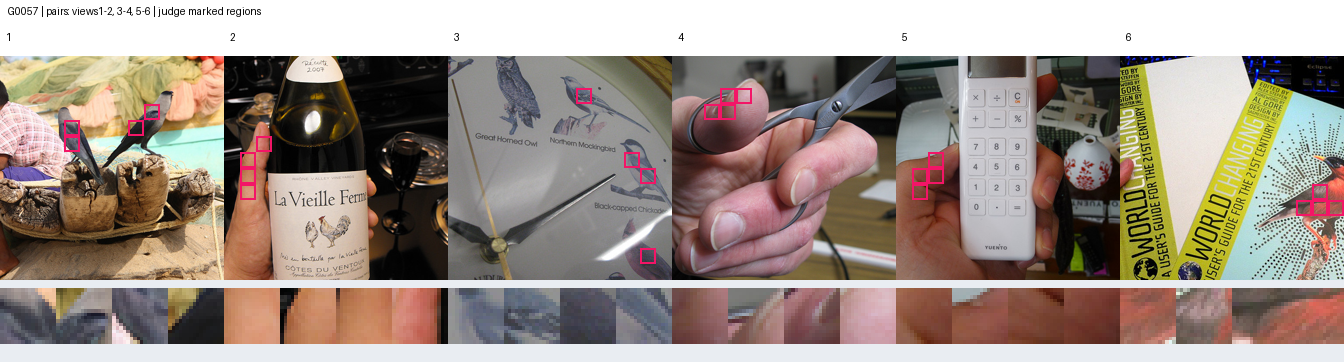}}\par\smallskip
\noindent{\small\textbf{(d) Vanilla BSF; group 35; G0136}\hfill
\footnotesize Specific pairs: 1/3; visual pairs: 1/3.}\par
{\footnotesize Birds or bird depictions, especially their heads and plumage. Theme matches: 4/6, 3/6, 4/6 (partial).}\par
\noindent\makebox[\textwidth][c]{\includegraphics[width=.82\textwidth]{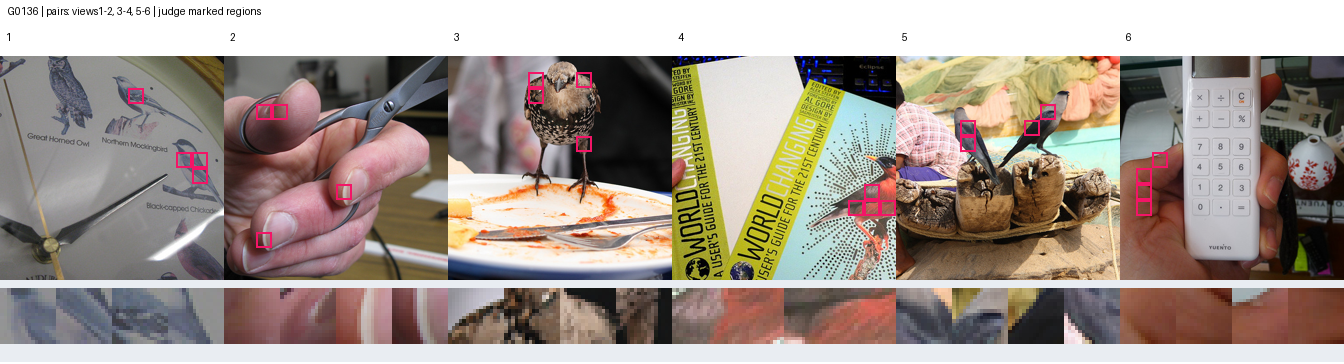}}\par\smallskip
\captionof{figure}{\textbf{Matched bird readouts.} GGF consistently marks feather surfaces.
SLiCS combines hair and feathers under a broader animal-covering theme;
the Vanilla variants mix bird-related regions with fingers and other surfaces.
GGF has three confirmed specific pairs, versus one for each comparator.
Its visual-pair score is also higher. SLiCS still has a clear whole-group
majority, while both Vanilla variants are partial.
For each method, the group is its frozen category-mask-AP calibration
readout; each sheet shows that group's six strongest audit images.
Pink boxes and zoom strips retain all original marked patches. Specific
pairs and theme matches follow Figure~\ref{fig:response-geometry}.
Displayed themes quote reviewer 1; counts use each reviewer's own theme.
These seed-21, 75-group cases were selected after review for GGF's
specific-pair advantage over every comparator; the all-group comparison
is in Table~\ref{tab:matched75-coherence}.}
\label{fig:blinded-birds}
\endgroup

\clearpage
\begin{center}
\noindent{\small\textbf{(a) GGF; group 51; G0716}\hfill
\footnotesize Specific pairs: 3/3; visual pairs: 2/3.}\par
{\footnotesize Doughnut surfaces with sugar, glaze, or toppings. Theme matches: 6/6, 6/6, 6/6 (clear).}\par
\noindent\makebox[\textwidth][c]{\includegraphics[width=.82\textwidth]{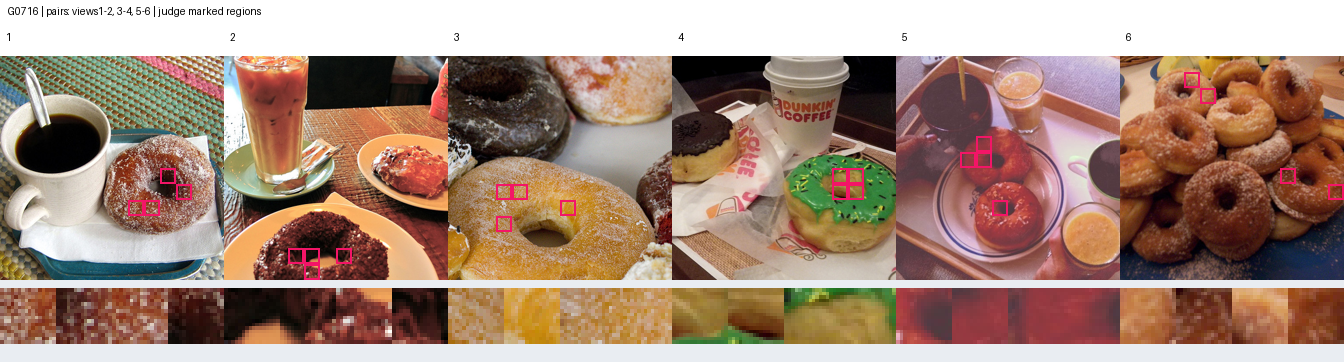}}\par\smallskip
\noindent{\small\textbf{(b) SLiCS; group 51; G0697}\hfill
\footnotesize Specific pairs: 1/3; visual pairs: 1/3.}\par
{\footnotesize Browned, bread-based savory foods, with topping-covered or filled surfaces. Theme matches: 6/6, 6/6, 6/6 (clear).}\par
\noindent\makebox[\textwidth][c]{\includegraphics[width=.82\textwidth]{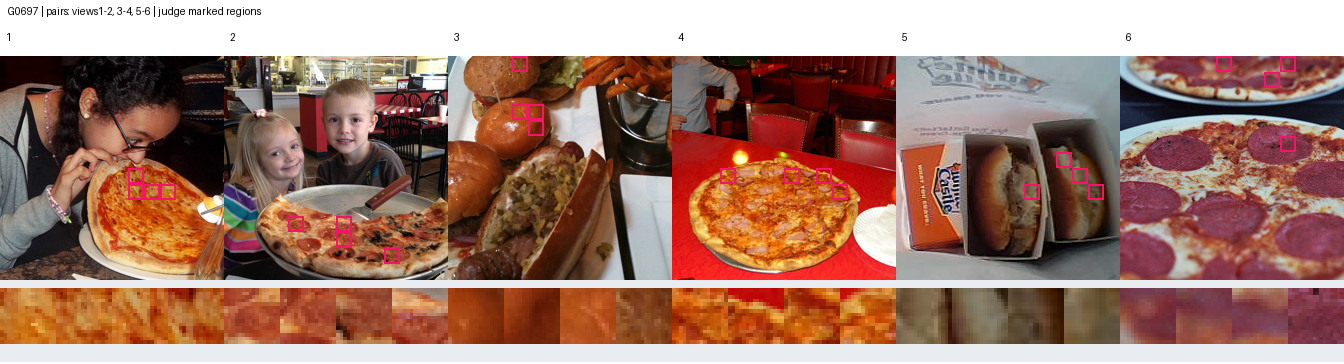}}\par\smallskip
\noindent{\small\textbf{(c) Supervised Vanilla BSF; group 7; G0468}\hfill
\footnotesize Specific pairs: 1/3; visual pairs: 1/3.}\par
{\footnotesize Open sky, including blue and cloudy portions. Theme matches: 4/6, 4/6, 4/6 (partial).}\par
\noindent\makebox[\textwidth][c]{\includegraphics[width=.82\textwidth]{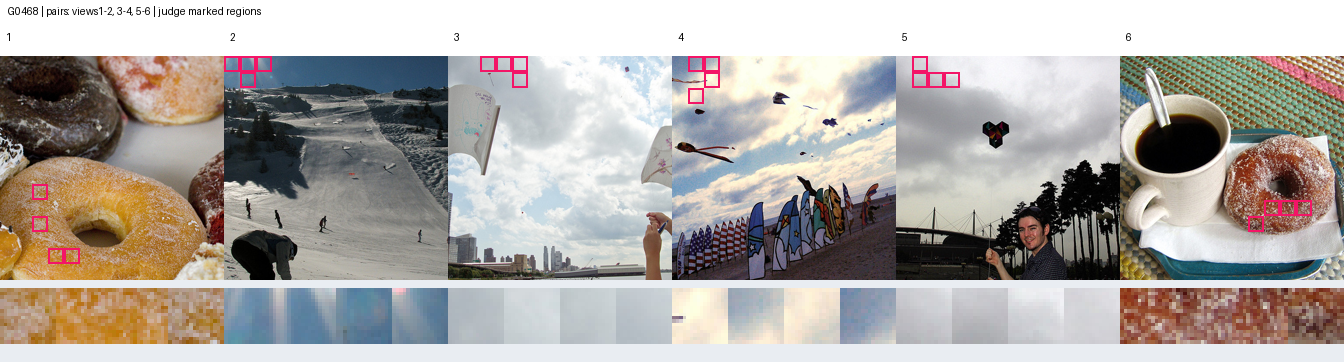}}\par\smallskip
\noindent{\small\textbf{(d) Vanilla BSF; group 6; G0364}\hfill
\footnotesize Specific pairs: 1/3; visual pairs: 1/3.}\par
{\footnotesize Dark chocolate-coated doughnut surfaces with an irregular granular texture. Theme matches: 3/6, 3/6, 3/6 (partial).}\par
\noindent\makebox[\textwidth][c]{\includegraphics[width=.82\textwidth]{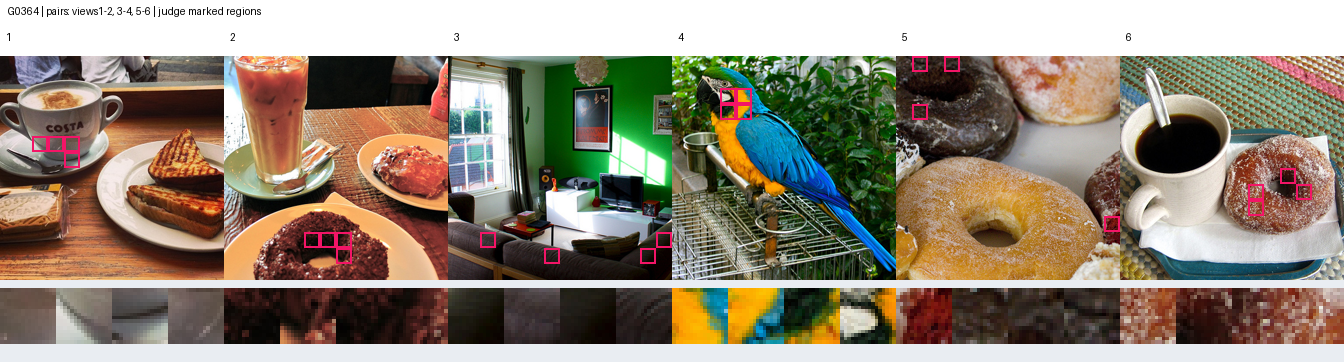}}\par\smallskip
\captionof{figure}{\textbf{Matched doughnut readouts.} GGF repeatedly marks doughnut surfaces,
whereas SLiCS groups other browned or topped foods, supervised Vanilla BSF
mixes doughnuts with sky, and Vanilla BSF mixes them with other regions.
GGF has three confirmed specific pairs, versus one for each comparator,
and a higher visual-pair score. SLiCS's broader food theme is clear;
both Vanilla variants have partial whole-group outcomes.
For each method, the group is its frozen category-mask-AP calibration
readout; each sheet shows that group's six strongest audit images.
Pink boxes and zoom strips retain all original marked patches. Specific
pairs and theme matches follow Figure~\ref{fig:response-geometry}.
Displayed themes quote reviewer 1; counts use each reviewer's own theme.
These seed-21, 75-group cases were selected after review for GGF's
specific-pair advantage over every comparator; the all-group comparison
is in Table~\ref{tab:matched75-coherence}.}
\label{fig:blinded-doughnuts}
\end{center}


\end{document}